\documentclass[10pt,twocolumn,letterpaper]{article}

\usepackage{subcaption}
\usepackage{stfloats}

\usepackage{arxiv}
\usepackage{times}
\usepackage{epsfig}
\usepackage{graphicx}
\usepackage{amsmath}
\usepackage{amssymb}
\usepackage{multicol}

\usepackage{grffile}
\usepackage[group-separator={,}]{siunitx}

\usepackage[breaklinks=true,bookmarks=false]{hyperref}

\begin{document}

\title{Deep Energies for Estimating Three-Dimensional Facial Pose and Expression}

\author{
Michael Bao\textsuperscript{1,2} \qquad Jane Wu\textsuperscript{1} \qquad Xinwei Yao\textsuperscript{1} \qquad Ronald Fedkiw\textsuperscript{1,2} \\
\textsuperscript{1}Stanford University \qquad \textsuperscript{2}Industrial Light \& Magic \\
{\tt\small \textsuperscript{1}\{mikebao,janehwu,yaodavid,rfedkiw\}@stanford.edu}
}

\maketitle

\begin{abstract}
    While much progress has been made in capturing high-quality facial performances using motion capture markers and shape-from-shading, high-end systems typically also rely on rotoscope curves hand-drawn on the image.
    These curves are subjective and difficult to draw consistently; moreover, ad-hoc procedural methods are required for generating matching rotoscope curves on synthetic renders embedded in the optimization used to determine three-dimensional facial pose and expression.
    We propose an alternative approach whereby these curves and other keypoints are detected automatically on both the image and the synthetic renders using trained neural networks, eliminating artist subjectivity and the ad-hoc procedures meant to mimic it.
    More generally, we propose using machine learning networks to implicitly define deep energies which when minimized using classical optimization techniques lead to three-dimensional facial pose and expression estimation.
\end{abstract}

\section{Introduction} \label{sec:intro}

For high-end face performance capture, either motion capture markers \cite{bhat2013high} or markerless techniques such as shape-from-shading \cite{aldrian2012inverse} or optical flow \cite{wu2016anatomically} are typically used; however, these methods are generally unable to capture the intricacies of the performance, especially around the lips.
To obtain high-end results, artists hand-draw rotoscope curves on the captured image; then, a variety of techniques are used to construct similar curves on the synthetic render of the estimated pose and to determine correspondences between the hand-drawn and synthetically generated curves.
The simplest such approach would be to use a pre-defined contour on the three-dimensional face model, clip it for occlusions, and create correspondences in a length proportional way; although this provides some consistency to the curves generated on the synthetic render, it is quite difficult for an artist to emulate these curves.
Thus, practical systems implement a number of ad-hoc methods in order to match the artist's more subjective interpretation.
The inability to embed the artist's subjectivity into the optimization loop and onto the synthetic render coupled with the artist's inability to faithfully reproduce procedurally generated curves leaves a gaping chasm in the uncanny valley.

Although one might debate what works best in order to align a three-dimensional virtual model with an image, it is clearly the case that a consistent metric should be applied to evaluate whether the synthetic render and image are aligned.
This motivatives the employment of a machine learning algorithm to draw the rotoscope curves on both the captured image and the synthetic render, hoping that accurately representing the real world pose would lead to a negligible difference between the two curves.
Although one might reasonably expect that the differences between real and synthetic camera behavior, albedo, lighting, etc.~may lead to different rotoscope curves being generated by the deep learning algorithm, GAN-like approaches \cite{goodfellow2014generative,li2017generative} could be used to rectify such issues by training a network to draw the curves such that a discriminator cannot tell which curves were generated on real images versus synthetic renders.

Recent advancements in using deep neural networks to detect face landmarks (see \eg \cite{bulat2017far}) make using machine learning to detect not only the lip curves but also other facial landmarks on both the captured image and the synthetic render an attractive option.
A traditional optimization approach can then be used to minimize the difference between the outputs from the captured image and the synthetic render.
This is feasible as long as one can backpropagate through the network to obtain the Jacobian of the output with respect to the synthetic render.
Of course, this assumes that the synthetic render is fully differentiable with respect to the face pose parameters, and we use OpenDR \cite{loper2014opendr} to satisfy this latter requirement; however, we note that any differentiable renderer can be used (\eg \cite{li2018differentiable}).

This approach is attractive as it replaces human subjectivity with a consistent, albeit sometimes consistently wrong, network to evaluate the semantic difference between two images.
Furthermore, as networks improve their ability to detect facial features/descriptors on a wider variety of poses and lighting conditions, our approach will also benefit.
More generally, we propose to use machine learning networks to embed various subjective, but consistent, ``evaluations'' into classical optimization approaches that estimate facial pose and expression.

\section{Related Work} \label{sec:related_work}

\begin{figure*}[t]
\centering
\begin{subfigure}[b]{0.16\linewidth}
    \includegraphics[width=\linewidth]{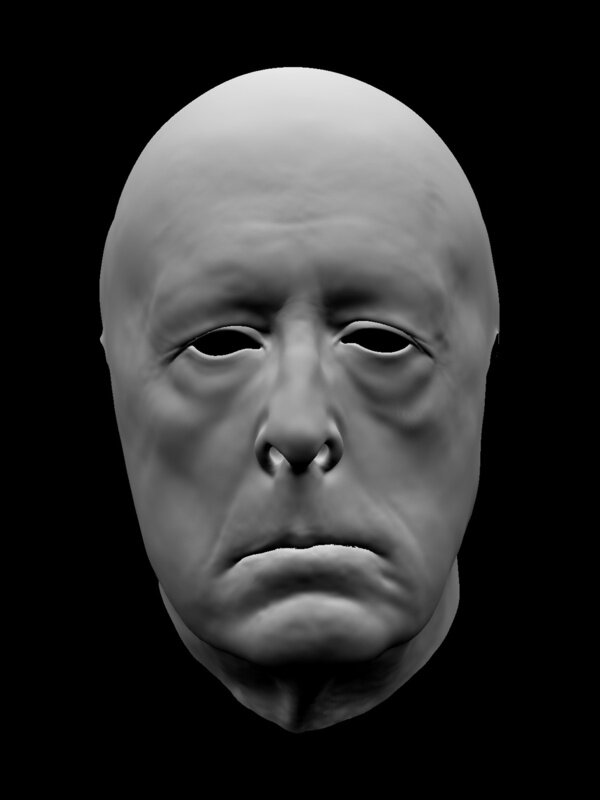}
    \caption{}
    \label{fig:overview_1}
\end{subfigure}
\begin{subfigure}[b]{0.16\linewidth}
    \includegraphics[width=\linewidth]{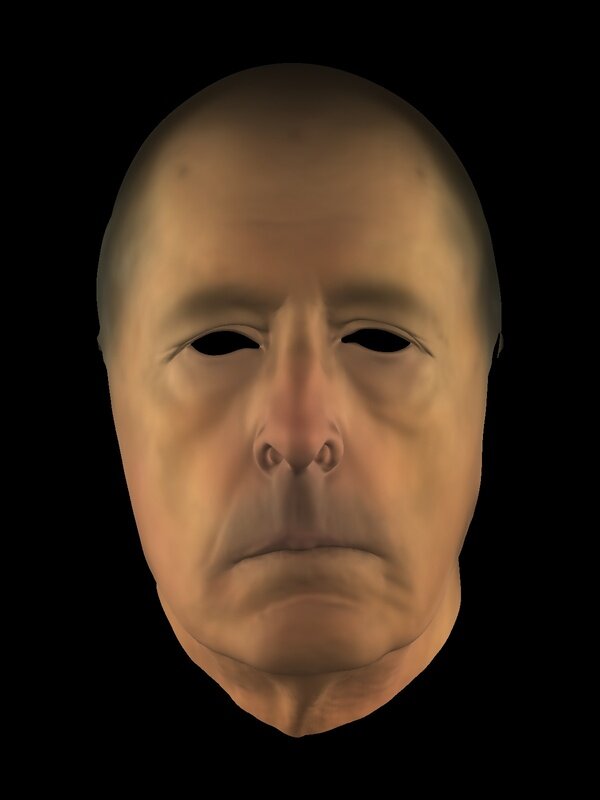}
    \caption{}
    \label{fig:overview_2}
\end{subfigure}
\begin{subfigure}[b]{0.16\linewidth}
    \includegraphics[width=\linewidth]{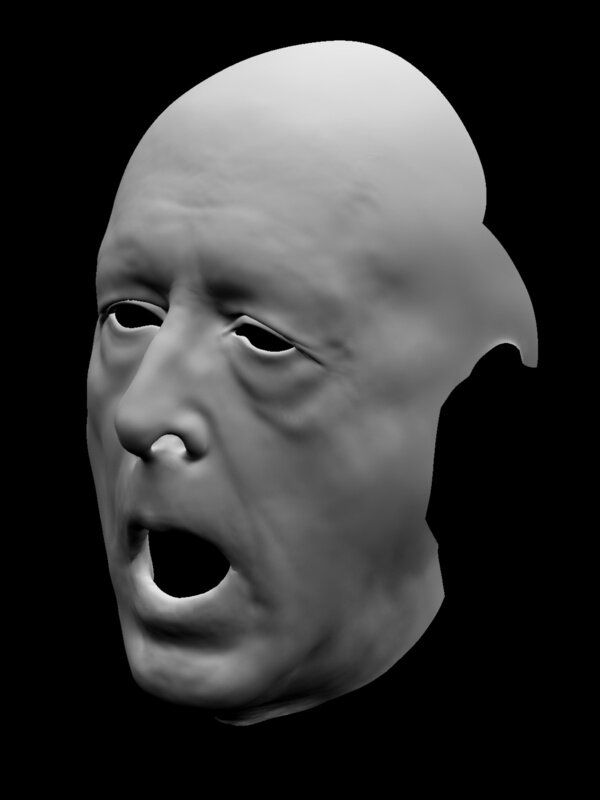}
    \caption{}
    \label{fig:overview_3}
\end{subfigure}
\begin{subfigure}[b]{0.16\linewidth}
    \includegraphics[width=\linewidth]{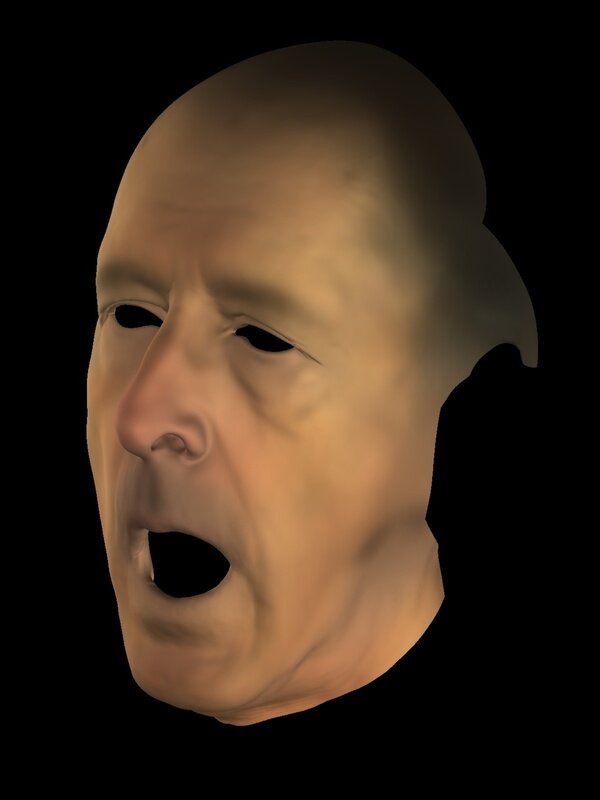}
    \caption{}
    \label{fig:overview_4}
\end{subfigure}
\begin{subfigure}[b]{0.16\linewidth}
    \includegraphics[width=\linewidth]{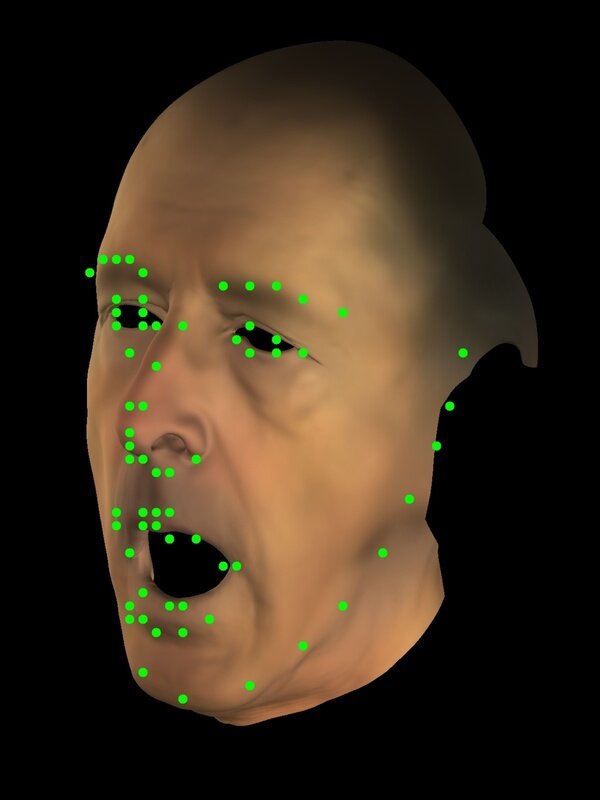}
    \caption{}
    \label{fig:overview_5}
\end{subfigure}
\begin{subfigure}[b]{0.16\linewidth}
    \includegraphics[width=\linewidth]{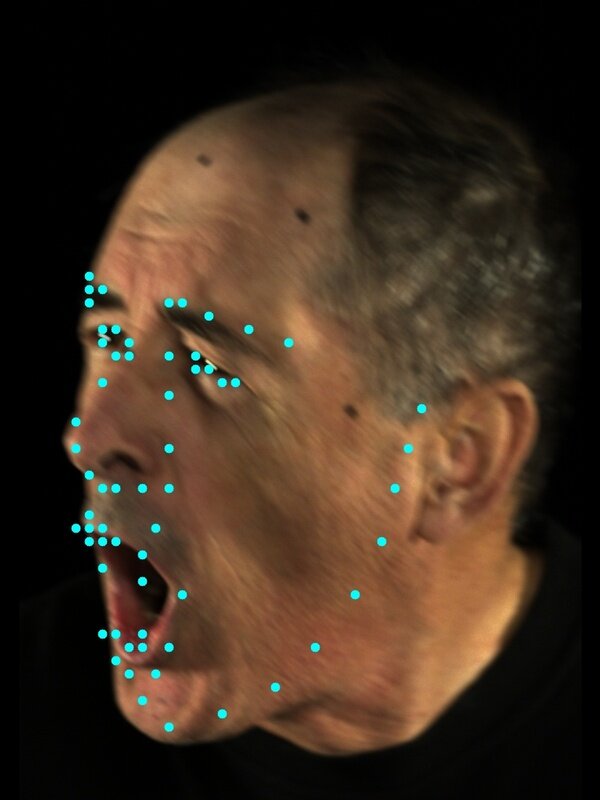}
    \caption{}
    \label{fig:overview_6}
\end{subfigure}
\hfill
\caption{An overview of our approach: we use our three-dimensional face model of the actor (a) and estimate the albedo and spherical harmonics lighting in a neutral pose (b).
Then, we can deform the face model into a variety of poses by changing the rigid parameters $\theta$ and $t$ and blendshape parameters $w$ (c), and generate synthetic images by rendering the face model in those poses (d).
We feed that synthetic render through the network to produce a set of outputs (e), which are then compared to the outputs produced by the same network when feeding it the captured image (f).
}
\label{fig:overview}
\end{figure*}

\textbf{Rotoscope Lip/Mouth Curves:}
Motion capture markers are commonly used to estimate facial pose and expression for a facial performance \cite{deng2006animating,ma2008facial,sifakis2005automatic}.
The corresponding points on the triangulated facial mesh are either selected by hand or found by projecting two-dimensional markers from the image to the three-dimensional mesh when the face is in a neutral pose \cite{bhat2013high}.
Rotoscope curves are typically used to refine the results around the lips to produce higher fidelity \cite{bhat2013high,dinev2018user}.
For view-independent curves such as the outer lip, the corresponding contour on the mesh can be pre-defined \cite{bhat2013high}; however, for view-dependent/silhouette curves such as the inner lip, the corresponding contour must be defined based on the current camera viewpoint and updated iteratively \cite{bhat2013high,dinev2018user}.

\textbf{Face Alignment:}
The earliest approaches to regression-based face alignment trained a cascade of regressors to detect face landmarks \cite{cao2014face,chen2014joint,feng2016dynamic,kazemi2014one,zhu2016unconstrained}.
More recently, deep convolutional neural networks (CNNs) have been used for both 2D and 3D facial landmark detection from 2D images \cite{jourabloo2017pose,wu2018look}.
These methods are generally classified into coordinate regression models \cite{jeni2017dense,jourabloo2017pose,toshev2014deeppose,xing2018towards}, where a direct mapping is learned between the image and the landmark coordinates, and heatmap regression models \cite{bulat2017far,deng2018cascade,zadeh2017convolutional}, where prediction heatmaps are learned for each landmark.
Heatmap-based architectures are generally derived from stacked hourglass \cite{bulat2017far,deng2018cascade,jackson2017large,newell2016stacked} or convolutional pose machine \cite{wei2016convolutional} architectures used for human body pose estimation.
Pixel coordinates can be obtained from the heatmaps by applying the argmax operation; however, \cite{dong2018supervision,sun2017integral} use soft-argmax to achieve end-to-end differentiability. 
A more comprehensive overview of face alignment methods can be found in \cite{jin2017face}.

\textbf{Optical Flow:}
Optical flow has had a long successful history started in part by \cite{black1996robust}.
Variational methods such as the Lucas-Kanade method \cite{lucas1981iterative} and the Brox method \cite{brox2004high} are commonly used.
Other correspondence finding strategies such as EpicFlow \cite{revaud2015epicflow} and FlowFields \cite{bailer2015flow} have also been used successfully.
We focus instead on trained deep networks.
End-to-end methods for learning optical flow using deep networks were first proposed by \cite{dosovitskiy2015flownet} and later refined in \cite{IMKDB17}.
Other methods include DeepFlow \cite{weinzaepfel2013deepflow}, etc. use deep networks to detect correspondences.
These methods are generally evaluated on the Middlebury dataset \cite{baker2011database}, the KITTI dataset \cite{geiger2013vision}, and the MPI Sintel dataset \cite{butler2012naturalistic}.

\textbf{Faces and Networks:}
Neural networks have been used for various other face image analysis tasks such as gender determination \cite{golomb1990sexnet} and face detection \cite{rowley1998face}. 
More recently, deep CNNs have been used to improve face detection results especially in uncontrolled environments and with more extreme poses \cite{haoxiangli2015cascade,kaipengzhang2016joint}. 
Additionally, CNNs have been employed for face segmentation \cite{sifeiliu2015labeling,nirkin2018swapping,saito2016segmentation}, facial pose and reflectance acquisition \cite{laine2017production,sengupta2018sfsnet}, and face recognition \cite{schroff2015facenet,taigman2014deepface}. 

Using deep networks such as VGG-16 \cite{simonyan2014very} for losses has been shown to be effective for training other deep networks for tasks such as style transfer and super-resolution \cite{johnson2016perceptual}.
Such techniques have also been used for image generation \cite{dosovitskiy2016generating} and face swapping \cite{korshunova2017fast}.
Furthermore, deep networks have been used in energies for traditional optimization problems for style transfer \cite{gatys2016image}, texture synthesis \cite{sendik2017deep}, and image generation \cite{mahendran2015understanding,ulyanov2018deep}.
While \cite{gatys2016image,sendik2017deep} use the L-BFGS \cite{zhu1997algorithm} method to minimize the optimization problem, \cite{mahendran2015understanding,ulyanov2018deep} use gradient descent methods \cite{ruder2016overview}.

\section{Overview} \label{sec:overview}

\begin{figure*}[t]
\centering
\includegraphics[width=\linewidth]{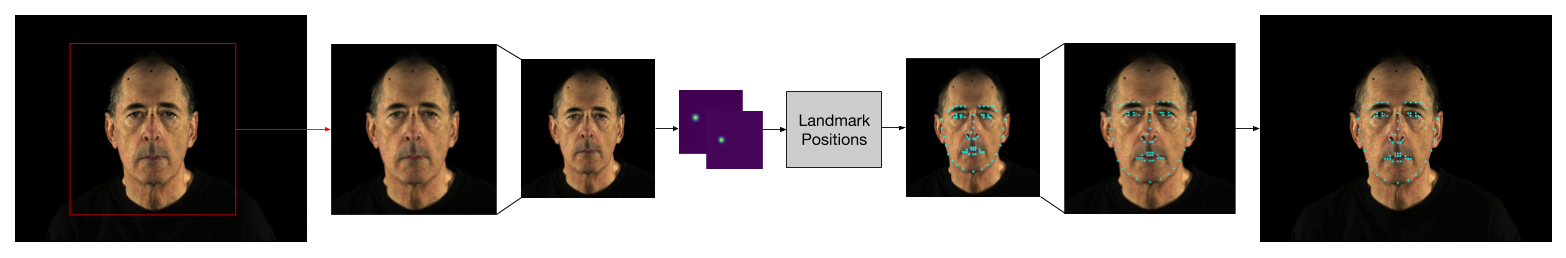}
\caption{A visual overview of our approach applied to facial landmark detection.
We pass the full resolution image through a facial detector and crop the face out of the image.
This crop is then resized to pass through the neural network which outputs, in this case, heatmaps for every landmark.
These heatmaps are processed using a soft-argmax operation to get facial landmark coordinates on the cropped and resized image.
These positions are then transformed back onto the full resolution image before being used as part of the objective function.
An identical process is performed for the synthetic render.}
\label{fig:landmarks_method}
\end{figure*}

In this paper, we advocate for a general strategy that uses classical optimization where the energy to be minimized is based on metrics ascertained from deep learning neural networks.
In particular, this removes the rotoscope artist and the ad-hoc contour drawing procedures meant to match the artist's work, or vice versa, from the pipeline.
This pins advancements in three-dimensional facial pose and expression estimation to those being made in machine learning, which are advancing at a fast pace.
Generally, we take the following approach: First, we estimate an initial rigid alignment of the three-dimensional face model to the two-dimensional image using a facial alignment network.
Then, we estimate an initial guess for the jaw and mouth expression using the same network.
Finally, we temporally refine the results and insert/repair failed frames (if/when necessary) using an optical flow network.

We use a blendshape model hybridized with linear blend skinning for a six degree of freedom jaw joint \cite{lewis2014practice}; let $w$ denote the parameters that drive the face triangulated surface $x(w)$.
The resulting surface has a rigid frame given by Euler angles $\theta$, rotation matrix $R(\theta)$, and a translation $t$ such that the final vertex positions are
\begin{align}
    x_R(\theta, t, w) = R(\theta)x(w) + t.
\label{eq:blendshapes}
\end{align}
We note that other geometry such as the teeth can be trivially handled by Equation \ref{eq:blendshapes} as well.
The geometry $x_R$ is rendered using OpenDR \cite{loper2014opendr} obtaining a rendered image $F(x_R)$.
As a precomputation, we estimate the face's albedo and nine coefficients for a spherical harmonics light \cite{ramamoorthi2001efficient} on a frame where the face is close to neutral; however, we note that a texture captured using a light stage \cite{debevec2000acquiring} (for example) would potentially work just as well if not better.
Then, our goal is to determine the parameters $\theta$, $t$, and $w$ that best match a given captured image $F^*$.

Both the pixels of captured image $F^*$ and the pixels of the rendered image $F(x_R)$ are fed through the same deep network to get two sets of landmark positions $N(F^*)$ and $N(F)$.
See Figure \ref{fig:overview}.
We use the L2 norm of the difference between them
\begin{align}
\| N(F^*) - N(F(x_R(\theta, t, w))) \|_2
\label{eq:network_diff_l2}
\end{align}
as the objective function to minimize via nonlinear least squares, which is solved using the Dogleg method \cite{lourakis2005levenberg} as implemented by Chumpy \cite{loperchumpy}.
This requires computing the Jacobian via the chain rule of the energy function; $\partial N / \partial F$, $\partial F / \partial x_R$, and $\partial x_R / \partial p$ where $p$ is one of $\theta$, $t$, and $w$ all need to be evaluated.
We use OpenDR to compute $\partial F / \partial x_R$, and Equation \ref{eq:blendshapes} yields $\partial x_R / \partial p$.
$\partial N / \partial F$ is computed by back-propagating through the trained network using one's deep learning library of choice; in this paper, we use PyTorch \cite{paszke2017automatic}.
Note that for computational efficiency, we do not compute $\partial N / \partial F$ explicitly, but instead compute the Jacobian of Equation \ref{eq:network_diff_l2} with respect to the rendered image pixels output by $F$ instead.

\section{Rigid Alignment} \label{sec:rigid}

We first solve for the initial estimate of the rigid alignment of the face, \ie $\theta$ and $t$ using the pre-trained 3D-FAN network \cite{bulat2017far}.
Note that 3D-FAN, and most other facial alignment networks, requires taking in a cropped and resized image of the face as input; we denote these two operations as $C$ and $S$ respectively.
The cropping function requires the bounding box output of a face detector $D$; we use the CNN-based face detector implemented by Dlib \cite{dlib09}.
The resize function resizes the crop to a resolution of $256 \times 256$ to feed into the network.
The final image passed to the network is thus $S(C(F(x_R), D(F(x_R))), D(F(x_R)))$ where we note that both the crop and resize functions depend on the output of the face detector; however, aggressively assuming that $\partial D / \partial F = 0$ did not impede our ability to estimate a reasonable facial pose.
Given $D$, $C$ is merely a subset of the pixels of $F$ so $\partial C / \partial p = \partial F / \partial p$ for all the pixels within the detected bounding box and $\partial C / \partial p = 0$ for all pixels outside.
$S$ resizes the crop using bilinear interpolation so $\partial S / \partial C$ can be computed using the size of the detected bounding box.

3D-FAN outputs a tensor of size $68 \times 64 \times 64$, \ie each of the \num{68} landmarks has a $64 \times 64$ heatmap specifying the likelihood of a particular pixel containing that landmark.
While one might difference the heatmaps directly, it is unlikely that this would sufficiently capture correspondences.
Instead, we follow the approach of \cite{dong2018supervision,sun2017integral} and apply a differentiable soft-argmax function to the heatmaps obtaining pixel coordinates for each of the \num{68} landmarks.
That is, given the marker position $m_i$ computed using the argmax function on heatmap $H_i$, we use a $3 \times 3$ patch of pixels $M_i$ around $m_i$ to compute the soft-argmax position as
\begin{align}
    \hat{m}_i & = \frac{\sum_{m \in M_i} m e^{\beta H_i(m)} }{\sum_{m \in M_i} e^{\beta H_i(m)}}
\end{align}
where $\beta = 50$ is set experimentally and $H_i(m)$ returns the heatmap value at a pixel coordinate $m$.
We found that using a small patch around the argmax landmark positions gives better results than running the soft-argmax operation on the entire heatmap.

The soft-argmax function returns an image coordinate on the $64 \times 64$ image, and these image coordinates need to be remapped to the full resolution image to capture translation between the synthetic face render and the captured image.
Thus, we apply inverse rescale $S^{-1}_m$ and crop operations $C^{-1}_m$, \ie $\tilde{m}_i = C^{-1}_m( S^{-1}_m(4\hat{m}_i, D), D)$.
The multiplication by $4$ rescales from the $64 \times 64$ heatmap to the original $256 \times 256$.
To summarize, we treat the \num{68} $\tilde{m}_i$ as the output of $N$, and Equation \ref{eq:network_diff_l2} measures the L2 distance between the \num{68} $\tilde{m}_i$ on the captured image and the corresponding $\tilde{m}_i$ on the synthetic render.
We once again assume $\partial D / \partial F = 0$.
$\partial C^{-1}_m / \partial S^{-1}_m$ is the identity matrix, and $\partial S^{-1}_m / \partial \hat{m}_i$ contains the scalar multipliers to resize the image from $256 \times 256$ to the original cropped size.
We stress that this entire process is end-to-end differentiable.
See Figure \ref{fig:landmarks_method}.

\section{Expression Estimation} \label{sec:expression}

After the rigid alignment determines $\theta$ and $t$, we solve for an initial estimate of the mouth and jaw blendshape parameters (a subset of $w$).
Generally, one would use hand-drawn rotoscope curves around the lips to accomplish this as discussed in Section \ref{sec:related_work}; however, given the multitude of problems associated with this method as discussed in Section \ref{sec:intro}, we instead turn to deep networks to accomplish the same goal.
We use 3D-FAN in the same manner as discussed in Section \ref{sec:rigid} to solve for a subset of the blendshape weights $w$ keeping the rigid parameters $\theta$ and $t$ fixed.
It is sometimes beneficial or even preferred to also allow $\theta$ and $t$ to be modified somewhat at this stage, although a prior energy term that penalizes these deviations from the values computed during the rigid alignment stage is often useful.

We note that the ideal solution would be to instead create new network architectures and train new models that are designed specifically for the purpose of detecting lip/mouth contours, especially since the $64 \times 64$ heatmaps generated by 3D-FAN are generally too low-resolution to detect fine mouth movements such as when the lips pucker.
However, since our goal in this paper is to show how to leverage existing architectures and pre-trained networks especially so one can benefit from the plethora of existing literature, for now, we bootstrap the mouth and jaw estimation using the existing facial landmark detection in 3D-FAN.

\section{Optical Flow for Missing Frames} \label{sec:optical_flow}

The face detector used in Section \ref{sec:rigid} can sometimes fail, \eg on our test sequence, the Dlib's HOG-based detector failed on \num{20} frames while Dlib's CNN-based detector succeeded on all frames.
We thus propose using optical flow networks to infer the rigid and blendshape parameters for failed frames by ``flowing'' these parameters from surrounding frames where the face detector succeeded.
This is accomplished by assuming that the optical flow of the synthetic render from one frame to the next should be identical to the corresponding optical flow of the captured image.
That is, given two synthetic renders $F_1$ and $F_2$ and two captured images $F^*_1$ and $F^*_2$, we can compute two optical flow fields $N(F_1, F_2)$ and $N(F^*_1, F^*_2)$ using FlowNet2 \cite{IMKDB17}.
We resize the synthetic renders and captured images to a resolution of $\num{512} \times \num{512}$ before feeding them through the optical flow network.
Assuming that $F^*_2$ is the image the face detector failed on, we solve for the parameters $p_2$ of $F_2$ starting with an initial guess $p_1$, the parameters of $F_1$, by minimizing the L2 difference between the flow field vectors $\|N(F^*_1, F^*_2) - N(F_1, F_2) \|_2$.
$\partial N / \partial F_2$ can be computed by back-propagating through the network.

\section{Temporal Refinement} \label{sec:temporal}

Since we solve for the rigid alignment and expression for all captured images in parallel, adjacent frames may produce visually disjointed results either because of noisy facial landmarks detected by 3D-FAN or due to the nonlinear optimization converging to different local minima.
Thus, we also use optical flow to refine temporal inconsistencies between adjacent frames.
We adopt a method that can be run in parallel.
Given three sequentially captured images $F^*_1$, $F^*_2$, and $F^*_3$, we compute two optical flow fields $N(F^*_1, F^*_2)$ and $N(F^*_2, F^*_3)$.
Similarly, we can compute $N(F_1, F_2)$ and $N(F_2, F_3)$.
Then, we solve for the parameters $p_2$ of $F_2$ by minimizing the sum of two L2 norms $\|N(F^*_1, F^*_2) - N(F_1, F_2) \|_2$ and $\|N(F^*_2, F^*_3) - N(F_2, F_3) \|_2$.
The details for computing the Jacobian follow that in Section \ref{sec:optical_flow}.
Optionally, one may also wish to add a prior penalizing the parameters $p_2$ from deviating too far from their initial value.
Here, step $k$ of smoothing to obtain a new set of parameters $p_i^k$ uses the parameters from the last step $p_{i \pm 1}^{k-1}$; however, one could also use the updated parameter values $p_{i \pm 1}^{k}$ whenever available in a Gauss-Seidel style approach.

Alternatively, one could adopt a self-smoothing approach by ignoring the capture image's optical flow and solving for the parameters $p_2$ that minimize $\|N(F_1, F_2) - N(F_2, F_3) \|_2$.
Such an approach in effect minimizes the second derivative of the motion of the head in the image plane, causing any sudden motions to be smoothed out; however, since the energy function contains no knowledge of the data being targeted, it is possible for such a smoothing operation to cause the model to deviate from the captured image.

While we focus on exploring deep learning based techniques, more traditional smoothing/interpolation techniques can also be applied in place of in addition to the proposed optical flow approaches.
Such methods include: spline fitting the rigid parameters and blendshape weights, smoothing the detected landmarks/bounding boxes on the captured images as a preprocess, smoothing each frame's parameters using the adjacent frame's estimations, etc.

\section{Results} \label{sec:results}

\begin{figure}[t]
\centering
\begin{subfigure}[b]{0.42\linewidth}
    \includegraphics[width=\linewidth]{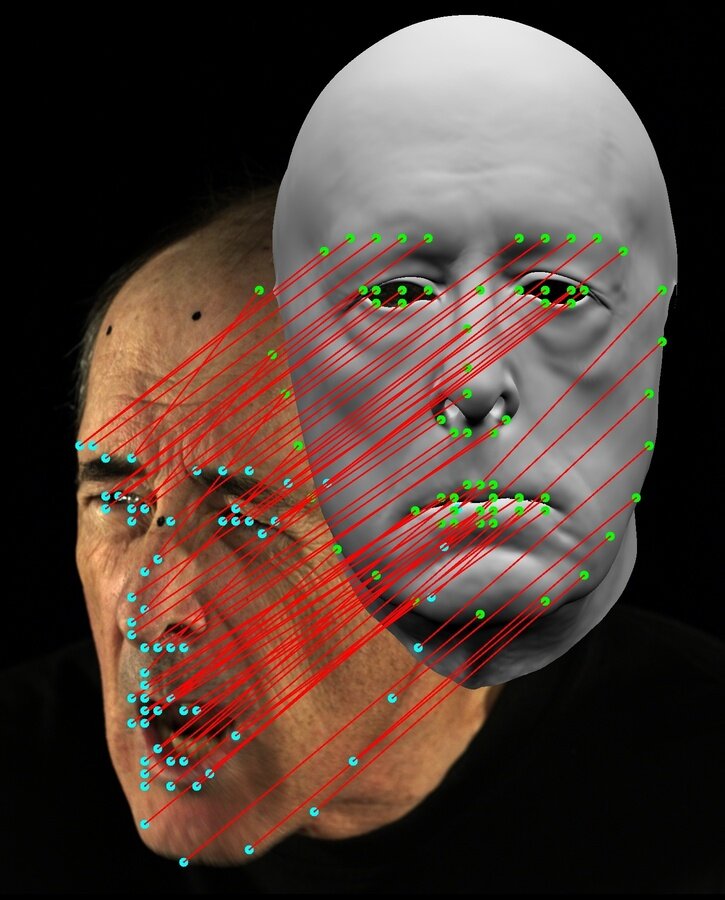}
\end{subfigure}
\begin{subfigure}[b]{0.42\linewidth}
    \includegraphics[width=\linewidth]{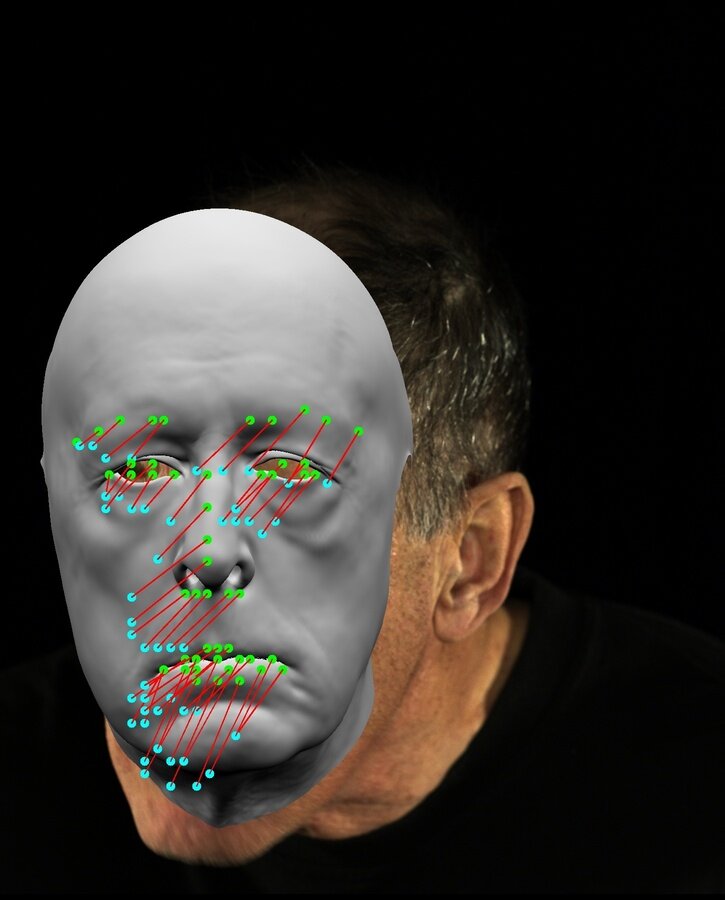}
\end{subfigure}
\hfill
\caption{
The green dots denote the landmarks detected on the synthetic render, the teal dots denote the landmarks detected on the captured image, and the red lines show correspondences.
Left: The initial state where the face is front facing and centered (note the figured is cropped) in the image plane. 
Right: The initial translation stage in the rigid alignment step roughly aligns the synthetic render of the face to the face in the captured image.
(Note that we display the synthetic render without the estimated albedo for clarity, but the network sees the version with the albedo as in Figures \ref{fig:overview_4} and \ref{fig:overview_6}, not \ref{fig:overview_3}).
}
\label{fig:initial_rigid}
\end{figure}

\begin{figure}[t]
\centering
\begin{subfigure}[b]{0.32\linewidth}
    \includegraphics[width=\linewidth]{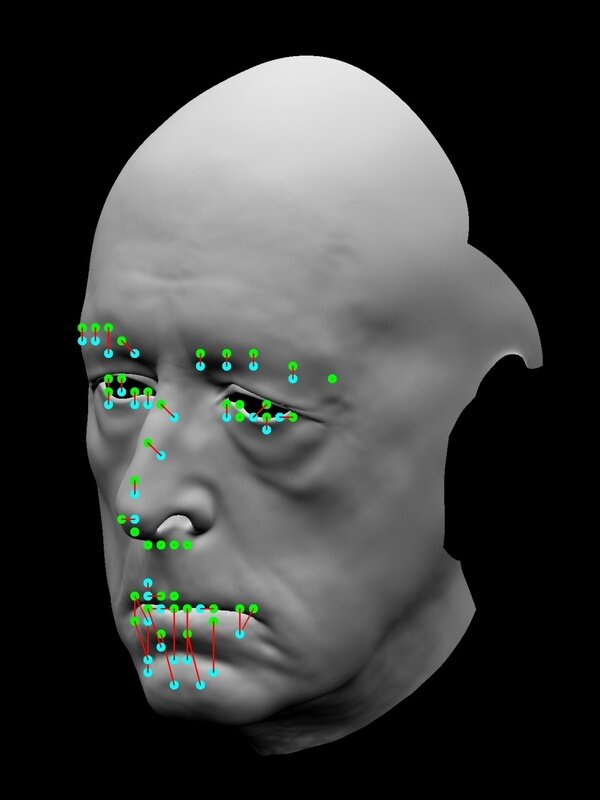}
\end{subfigure}
\begin{subfigure}[b]{0.32\linewidth}
    \includegraphics[width=\linewidth]{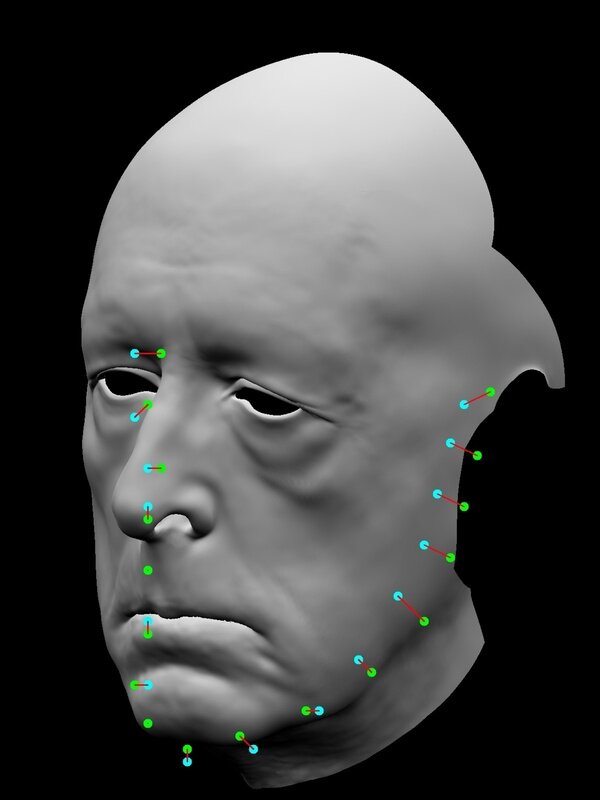}
\end{subfigure}
\begin{subfigure}[b]{0.32\linewidth}
    \includegraphics[width=\linewidth]{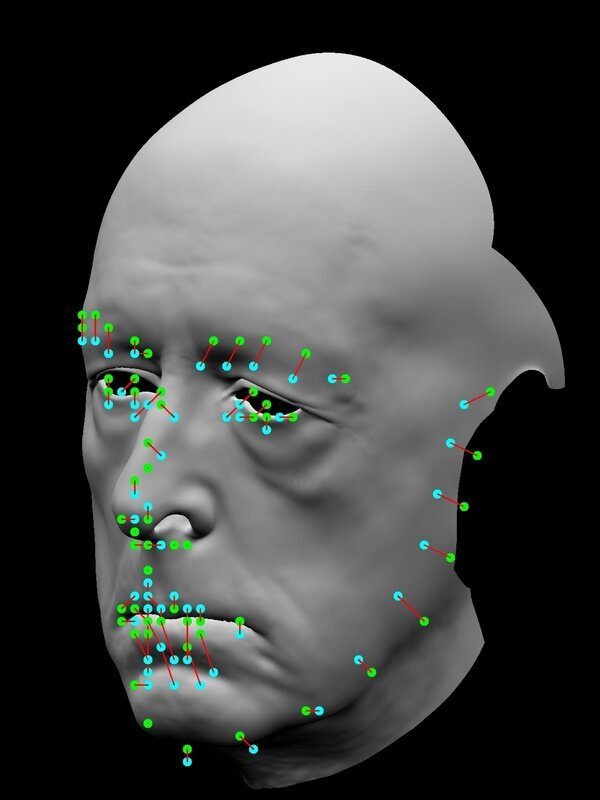}
\end{subfigure}
\hfill
\caption{From left to right: the result after solving for $\theta$ and $t$ using the non-jaw markers, using only the jaw markers, and using all the markers.
}
\label{fig:rigid_alignment}
\end{figure}

We estimate the facial pose and expression on a moderately challenging performance captured by a single ARRI Alexa XT Studio running at 24 frames-per-second with an \num{180} degree shutter angle at ISO \num{800} where numerous captured images exhibit motion blur.
These images are captured at a resolution of $2880 \times 2160$, but we downsample them to $720 \times 540$ before feeding them through our pipeline.
We assume that the camera intrinsics and extrinsics have been pre-calibrated, the captured images have been undistorted, and that the face model described in Equation \ref{eq:blendshapes} has already been created.
Furthermore, we assume that the face's rigid transform has been set such that the rendered face is initially visible and forward-facing in all the captured viewpoints.

\subsection{Rigid Alignment} \label{sec:rigid_results}

\begin{figure}[t]
\centering
\begin{subfigure}[b]{0.32\linewidth}
    \includegraphics[width=\linewidth]{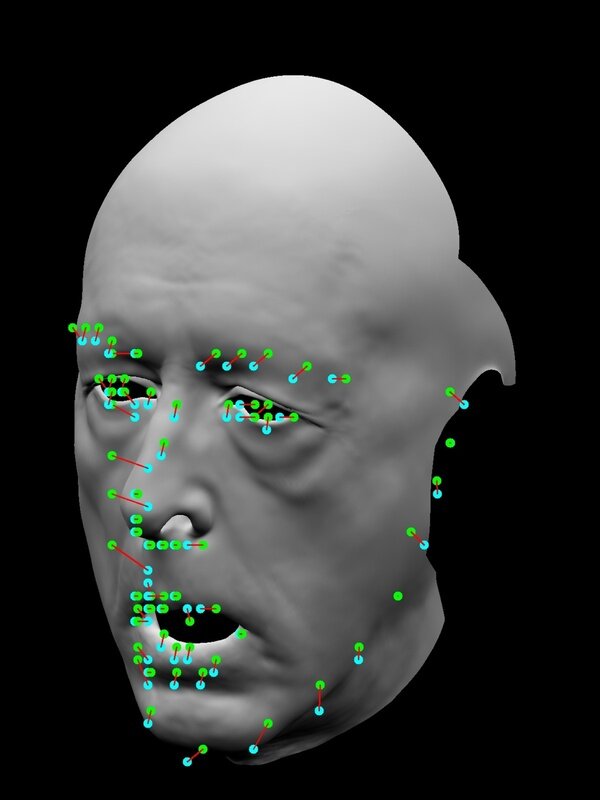}
\end{subfigure}
\begin{subfigure}[b]{0.32\linewidth}
    \includegraphics[width=\linewidth]{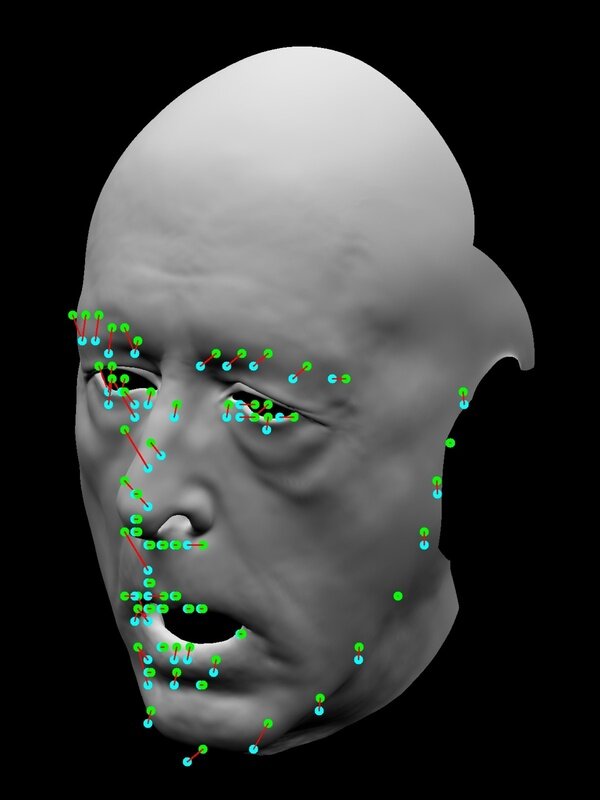}
\end{subfigure}
\begin{subfigure}[b]{0.32\linewidth}
    \includegraphics[width=\linewidth]{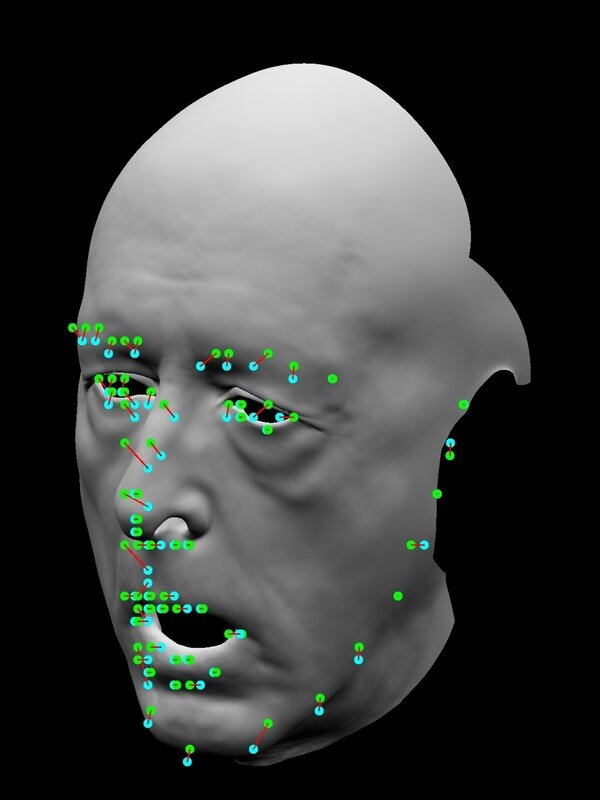}
\end{subfigure}
\hfill
\caption{From left to right: the results after solving for only the jaw open blendshape, all jaw-related blendshapes, and all jaw and mouth related blendshapes.
}
\label{fig:expression_estimation}
\end{figure}

\begin{figure}[t]
\begin{subfigure}[b]{0.32\linewidth}
    \includegraphics[width=\linewidth]{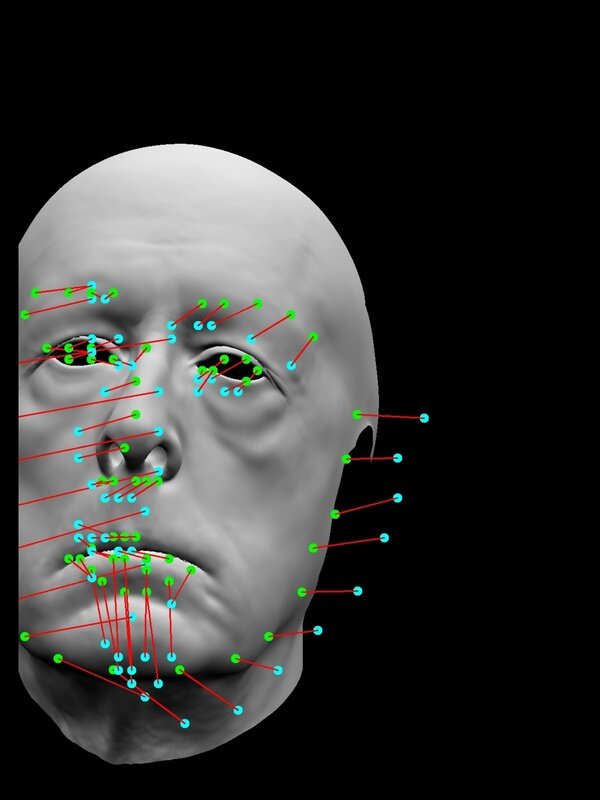}
\end{subfigure}
\begin{subfigure}[b]{0.32\linewidth}
    \includegraphics[width=\linewidth]{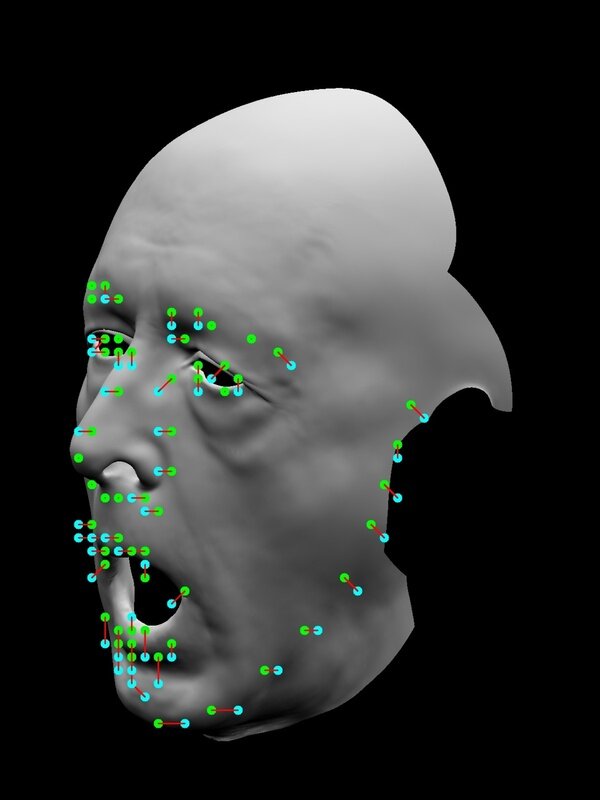}
\end{subfigure}
\begin{subfigure}[b]{0.32\linewidth}
    \includegraphics[width=\linewidth]{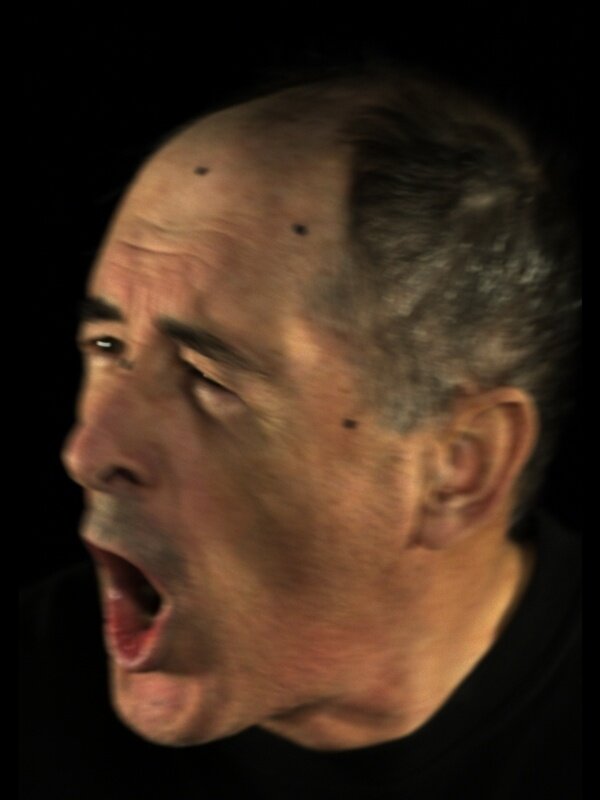}
\end{subfigure}
\hfill
\caption{
From left to right: the rigid alignment overfit to the mouth markers, the correction by expression estimation, and the target captured image.
The rigid alignment overfits to the non-jaw contour markers during the initial translation and rotation steps; however, this is corrected during expression estimation when the jaw is allowed to open.}
\label{fig:expression_fix_rigid}
\end{figure}

We estimate the rigid alignment (\ie $\theta$ and $t$) of the face using 3D-FAN.
We use an energy $E_1 = W(N(F) - N(F^*))$ where $N$ are the image space coordinates of the facial landmarks as described in Section \ref{sec:rigid} and $W$ is a per-landmark weighting matrix.
Furthermore, we use an edge-preserving energy $E_2 = \sum_i (\tilde{m}_i^{F^*} - \tilde{m}_{i-1}^{F^*}) - (\tilde{m}_i^F - \tilde{m}_{i-1}^F)$ where $\tilde{m}_i^{F^*}$ are the landmark positions on the captured image and $\tilde{m}_i^F$ are the landmark positions on the synthetic renders to ensure that the face does not erroneously grow/shrink in projected size as it moves towards the target landmarks, which may prevent the face detector from working.

First, we only solve for $t$ using all the landmarks except for those around the jaw to bring the initial state of the face into the general area of the face on the captured image.
See Figure \ref{fig:initial_rigid}.
We prevent the optimization from overfitting to the landmarks by limiting the maximum number of iterations.
Next, we solve for both $\theta$ and $t$ in three steps: using the non-jaw markers, using only the jaw markers, and using all markers.
We perform these steps in stages as we generally found the non-jaw markers to be more reliable and use them to guide the face model to the approximate location before trying to fit to all existing markers.
See Figure \ref{fig:rigid_alignment}.

\subsection{Expression Estimation} \label{sec:expression_results}

\begin{figure*}[t]
\centering
\begin{subfigure}[b]{\dimexpr0.10\linewidth+20pt\relax}
    \makebox[20pt]{\raisebox{30pt}{\rotatebox[origin=c]{90}{Target}}}%
    \includegraphics[width=\dimexpr\linewidth-20pt\relax]{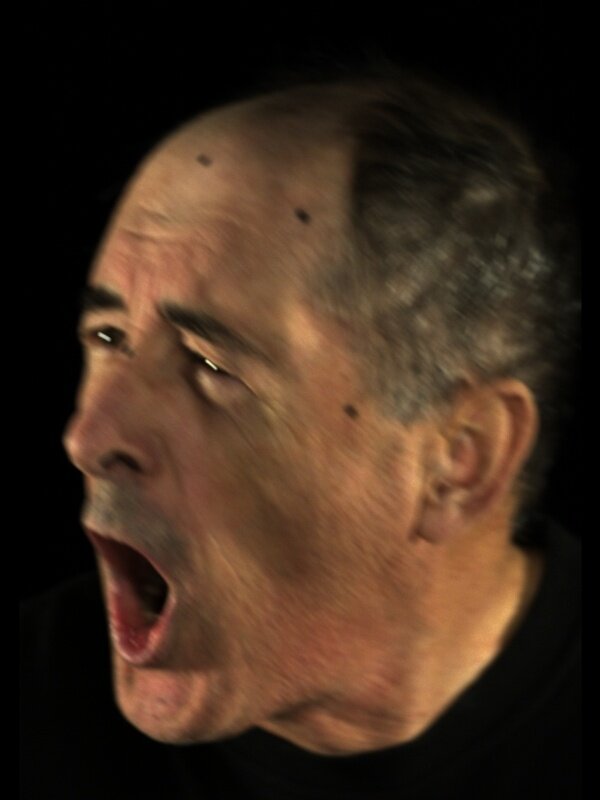}
    \makebox[20pt]{\raisebox{30pt}{\rotatebox[origin=c]{90}{Render}}}%
    \includegraphics[width=\dimexpr\linewidth-20pt\relax]{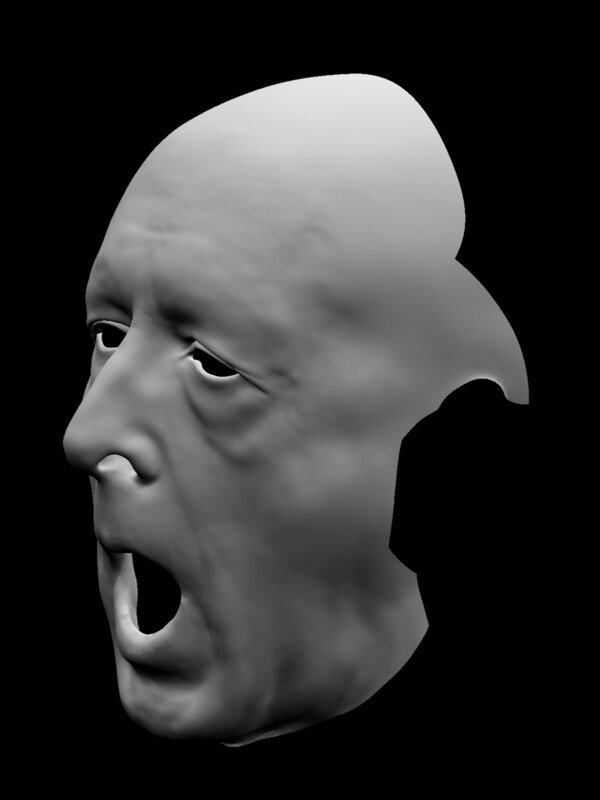}
    \caption*{1142}
\end{subfigure}
\begin{subfigure}[b]{0.10\linewidth}
    \includegraphics[width=\linewidth]{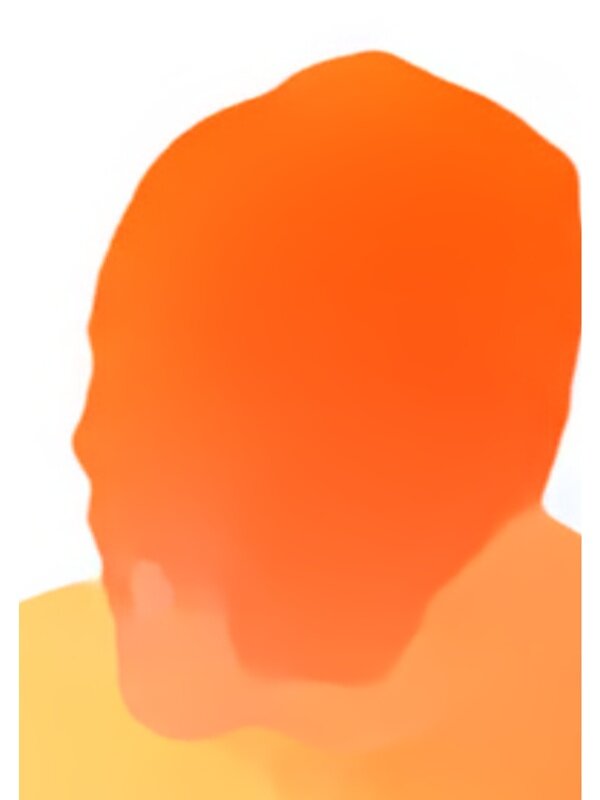}
    \includegraphics[width=\linewidth]{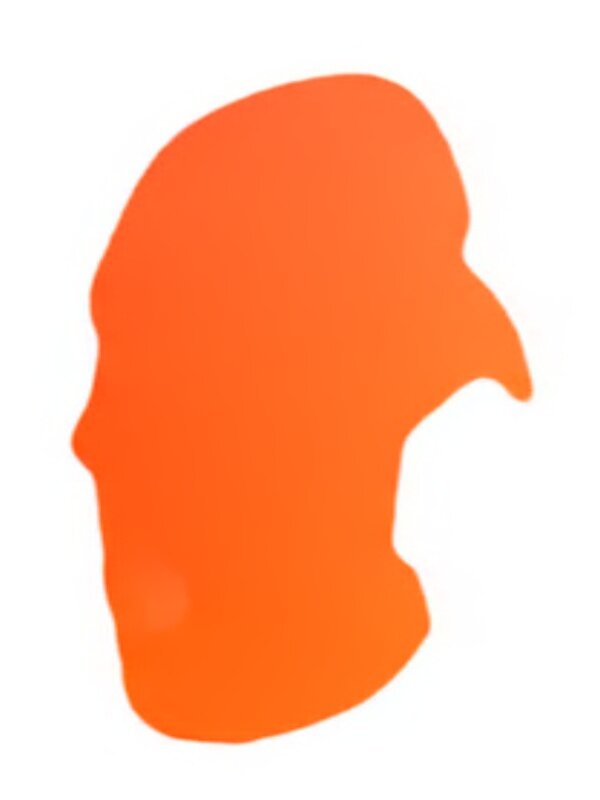}
    \caption*{}
\end{subfigure}
\begin{subfigure}[b]{0.10\linewidth}
    \includegraphics[width=\linewidth]{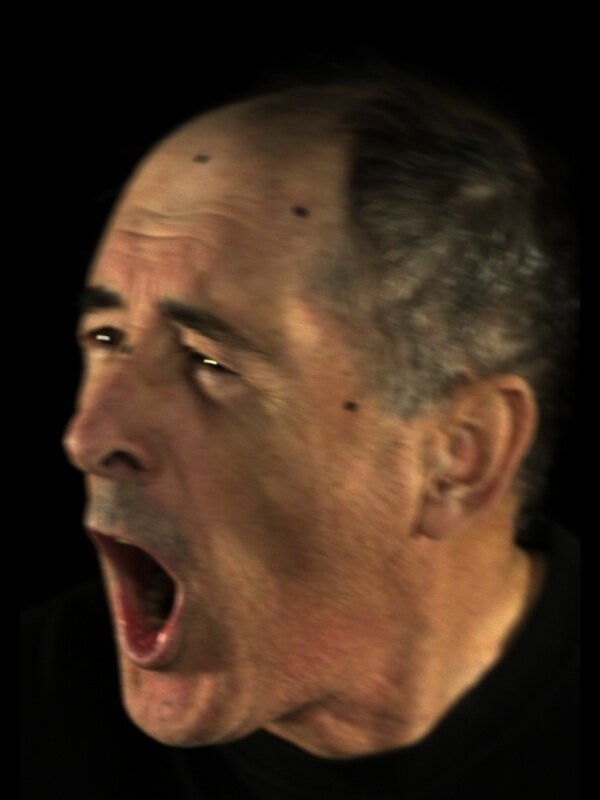}
    \includegraphics[width=\linewidth]{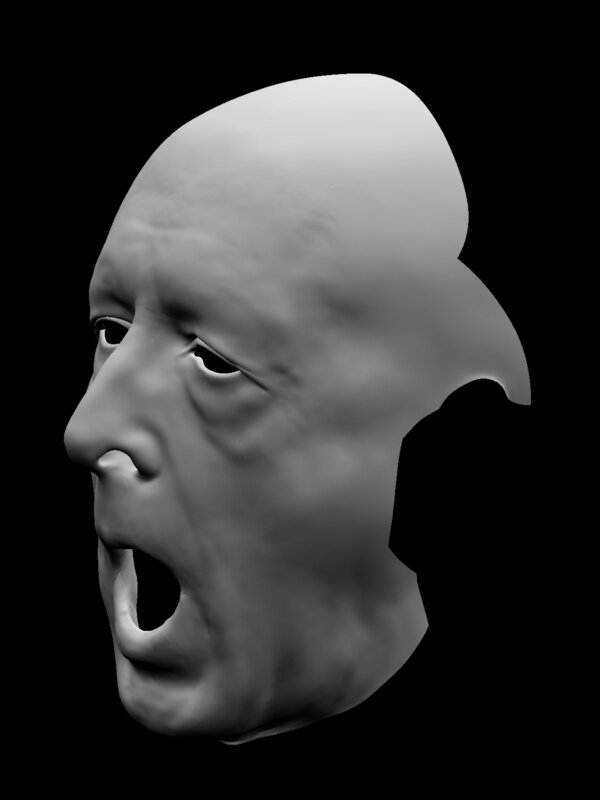}
    \caption*{1143}
\end{subfigure}
\begin{subfigure}[b]{0.10\linewidth}
    \includegraphics[width=\linewidth]{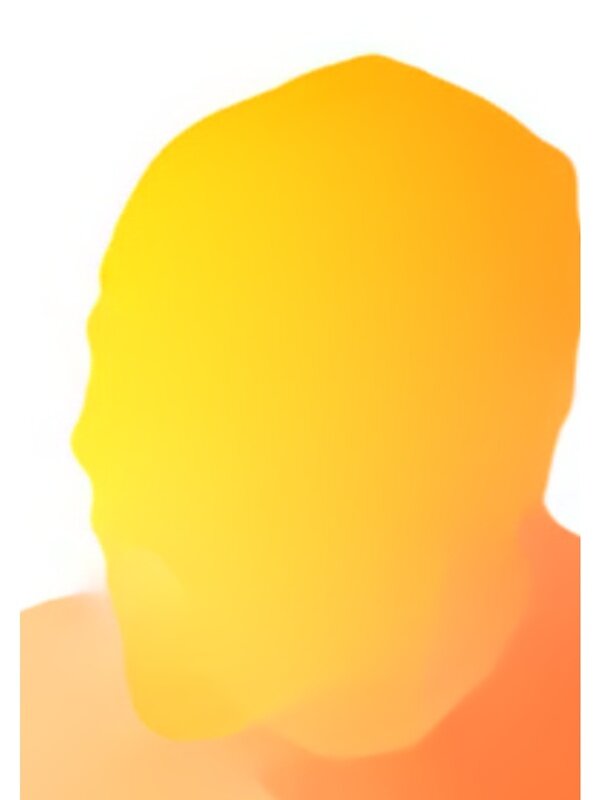}
    \includegraphics[width=\linewidth]{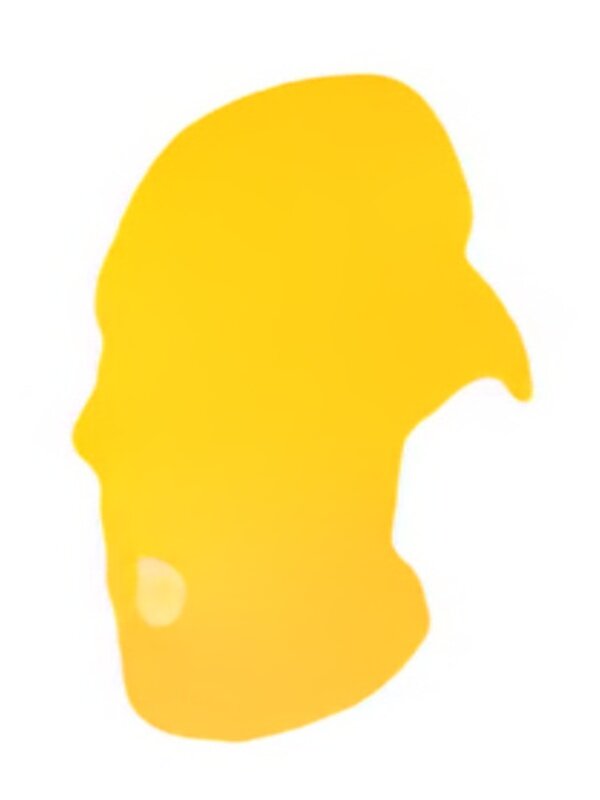}
    \caption*{}
\end{subfigure}
\begin{subfigure}[b]{0.10\linewidth}
    \includegraphics[width=\linewidth]{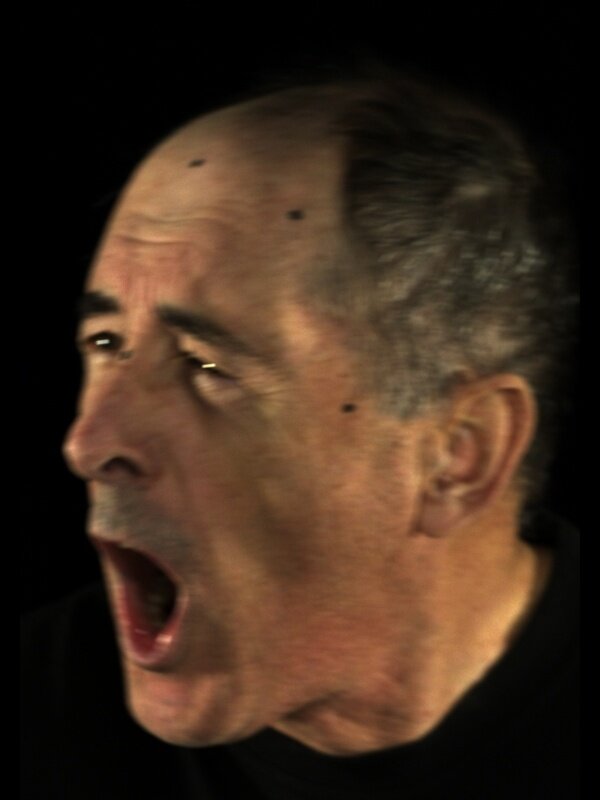}
    \includegraphics[width=\linewidth]{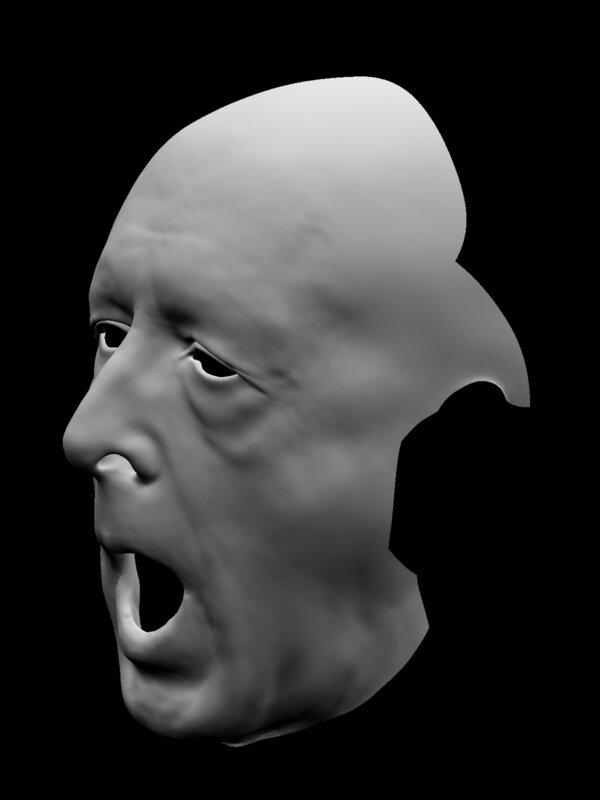}
    \caption*{1144}
\end{subfigure}
\begin{subfigure}[b]{0.10\linewidth}
    \includegraphics[width=\linewidth]{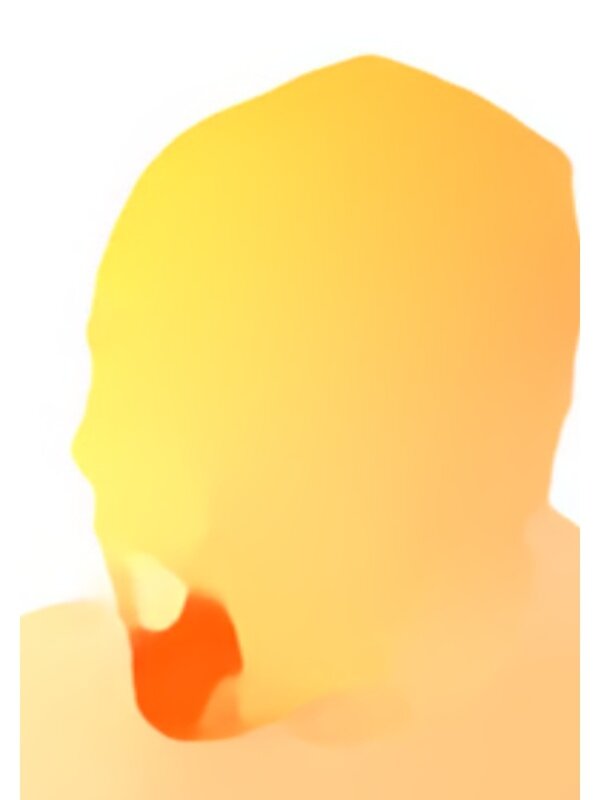}
    \includegraphics[width=\linewidth]{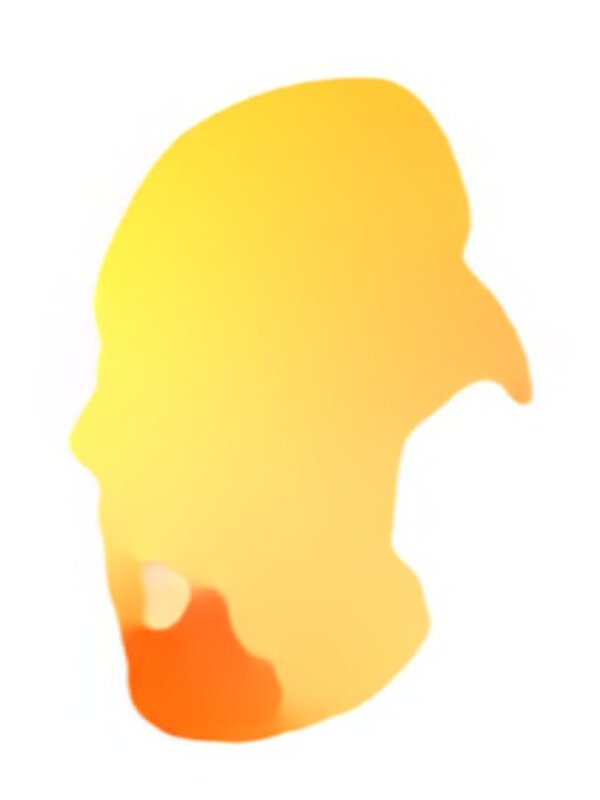}
    \caption*{}
\end{subfigure}
\begin{subfigure}[b]{0.10\linewidth}
    \includegraphics[width=\linewidth]{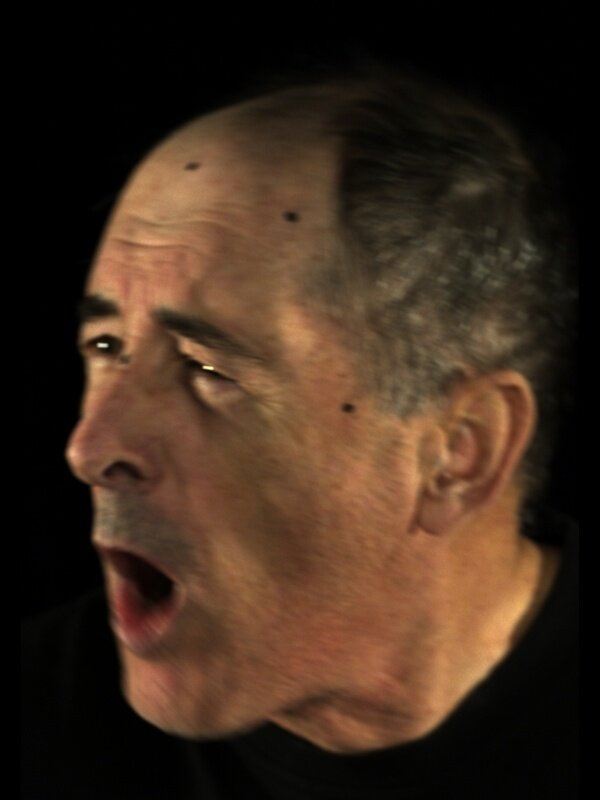}
    \includegraphics[width=\linewidth]{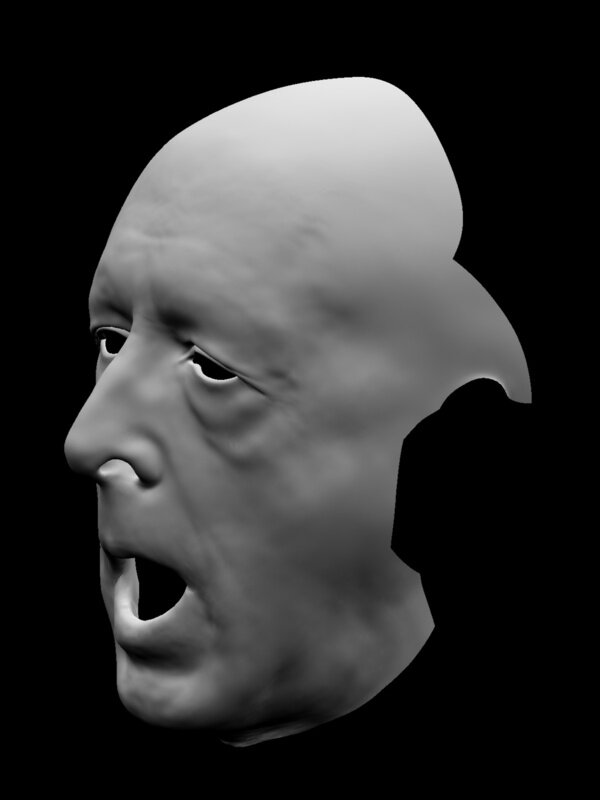}
    \caption*{1145}
\end{subfigure}
\begin{subfigure}[b]{0.10\linewidth}
    \includegraphics[width=\linewidth]{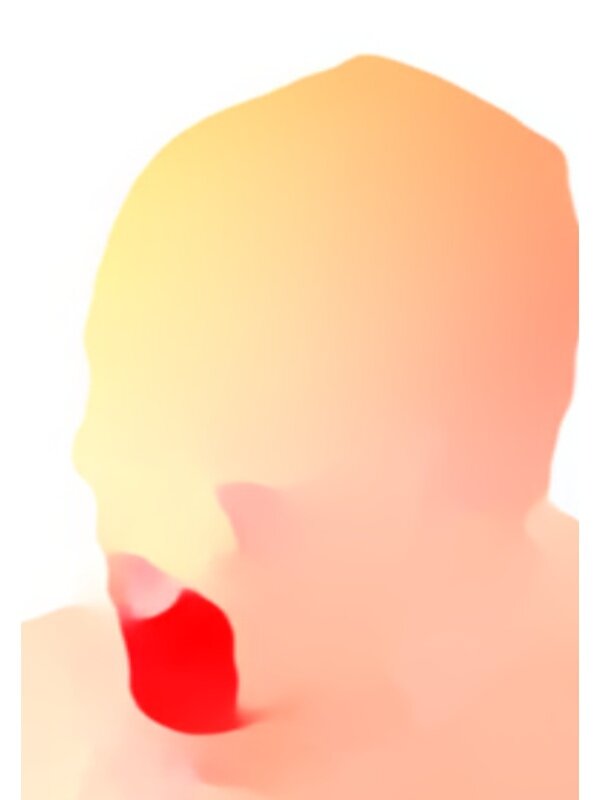}
    \includegraphics[width=\linewidth]{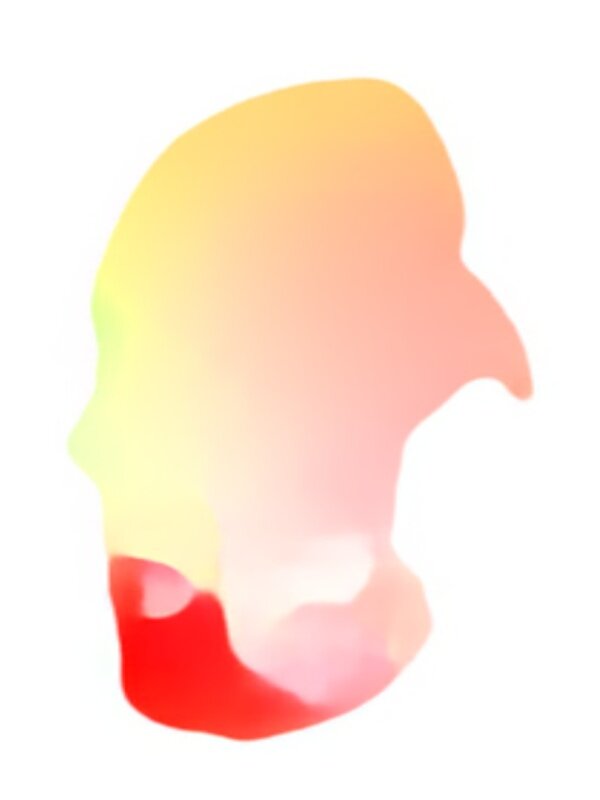}
    \caption*{}
\end{subfigure}
\begin{subfigure}[b]{0.10\linewidth}
    \includegraphics[width=\linewidth]{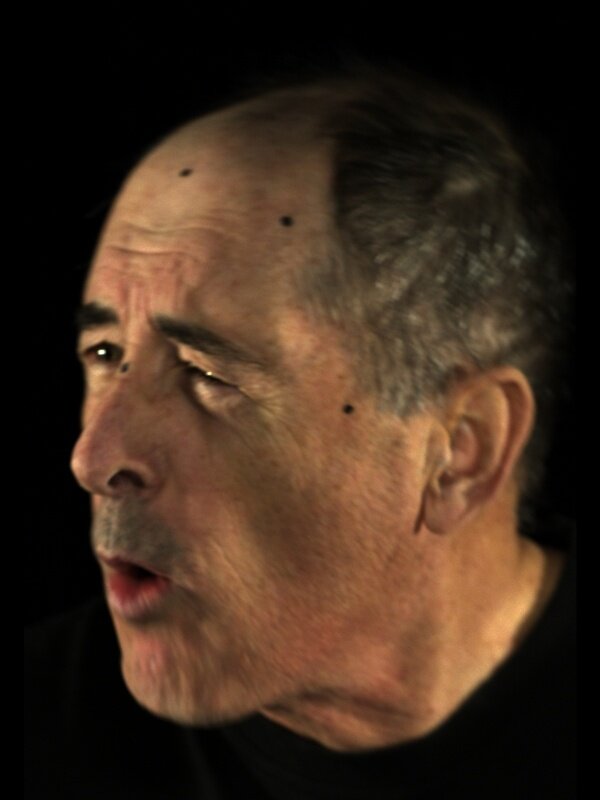}
    \includegraphics[width=\linewidth]{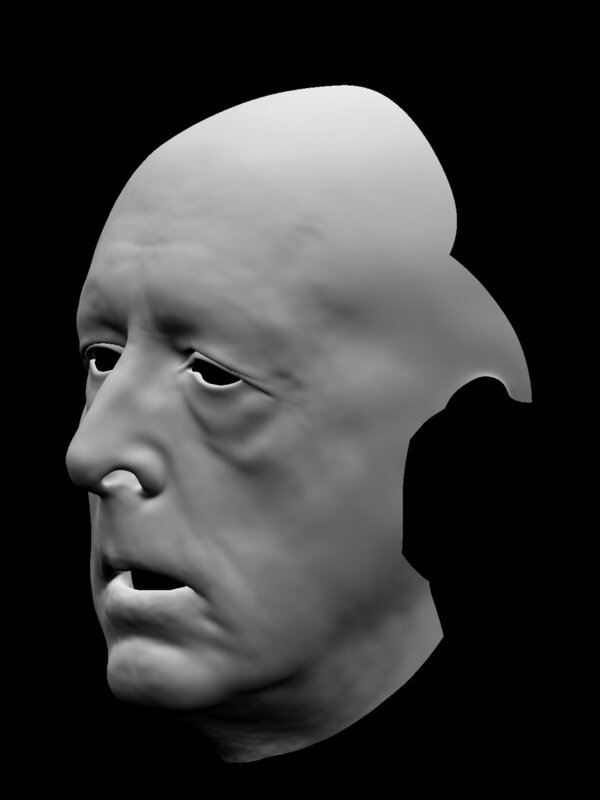}
    \caption*{1146}
\end{subfigure}
\hfill
\caption{We use optical flow to infill frames where the face detector fails.
Starting from frame \num{1142}, the optimization moves the head to match the optical flow of the synthetic render to the optical flow of the captured image.
After solving for frame \num{1145}, we perform another round of optimization except that we start from frame \num{1146} instead; this way, each frame will capture optical flow information from both anchor frames.
Using optical flow allows the mouth to stay open longer (\eg in frame \num{1144}) than what one would obtain using simple interpolation.
}
\label{fig:oflow_infill}
\vspace{-2.5mm}
\end{figure*}

\begin{figure}[b]
\begin{subfigure}[b]{0.32\linewidth}
    \includegraphics[width=\linewidth]{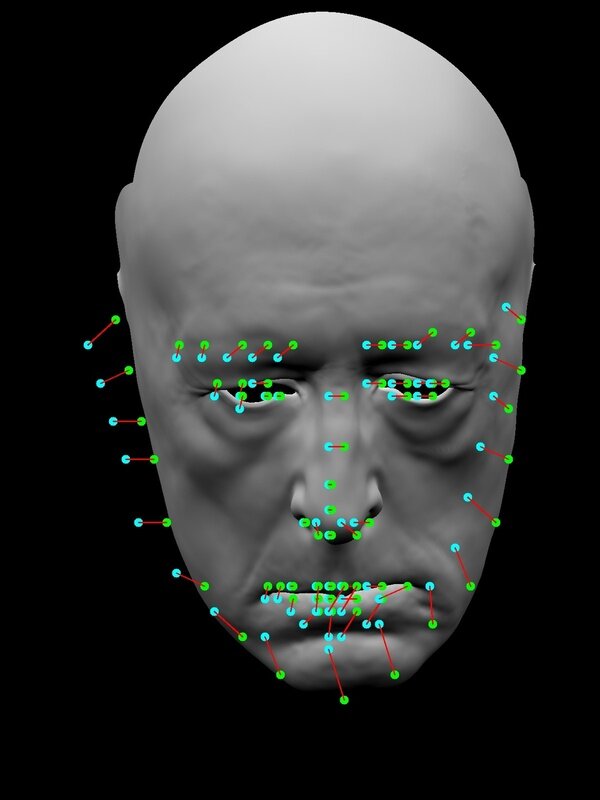}
\end{subfigure}
\begin{subfigure}[b]{0.32\linewidth}
    \includegraphics[width=\linewidth]{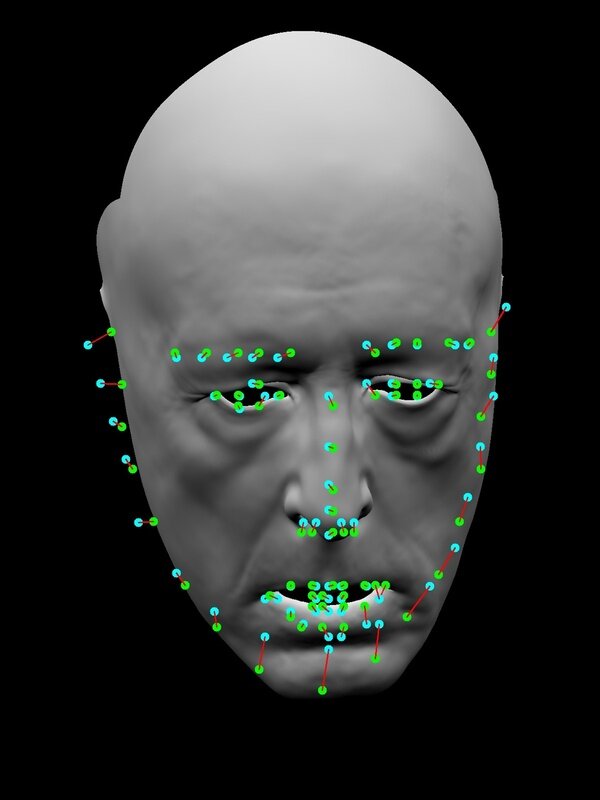}
\end{subfigure}
\begin{subfigure}[b]{0.32\linewidth}
    \includegraphics[width=\linewidth]{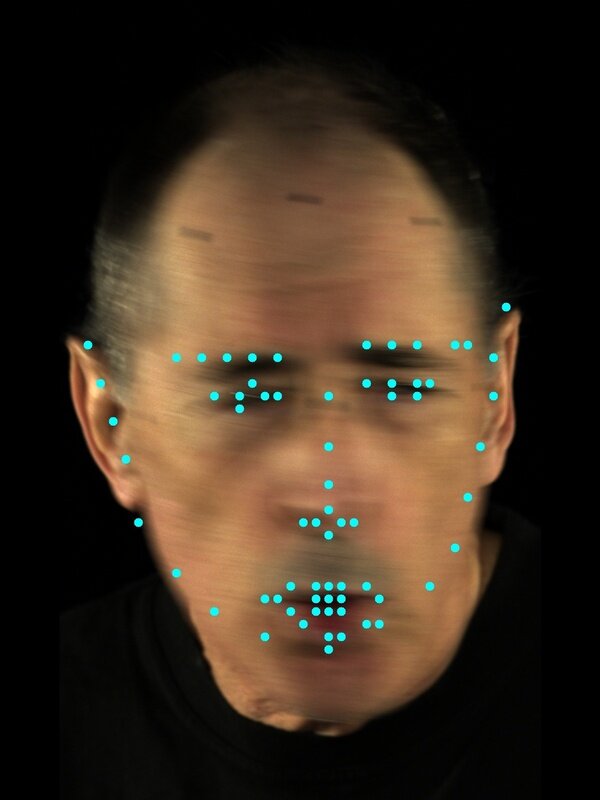}
\end{subfigure}
\hfill
\caption{
From left to right: the result after rigid alignment, after expression estimation, and the captured image.
Erroneous markers such as those around the jaw cause the optimization to land in an inaccurate local minima.
}
\label{fig:fan_failure_cases}
\end{figure}

We run a similar multi-stage process to estimate facial expression using the detected 3D-FAN landmarks.
We use the same energy term $E_1$ as Section \ref{sec:rigid_results}, but also introduce L2 regularization on the blendshape weights $E_2 = \lambda w$ with $\lambda = \num{1e2}$ set experimentally.
In the first stage, we weight the landmarks around the mouth and lips more heavily and estimate only the jaw open parameter along with the rigid alignment.
The next stage estimates all available jaw-related blendshape parameters using the same set of landmarks.
The final stage estimates all available jaw and mouth-related blendshapes as well as the rigid alignment using all available landmarks.
See Figure \ref{fig:expression_estimation}.
This process will also generally correct any overfitting introduced during the rigid alignment due to not being able to fully match the markers along the mouth.
See Figure \ref{fig:expression_fix_rigid}.

Our approach naturally depends on the robustness of 3D-FAN's landmark detection on both the captured images and synthetic renders.
As seen in Figure \ref{fig:fan_failure_cases}, the optimization will try to target the erroneous markers producing inaccurate $\theta$, $t$, and $w$ which overfit to the markers.
Such frames should be considered a failure case and thus require using the optical flow approach described in Section \ref{sec:optical_flow} for infill.
Alternatively, one could manually modify the multi-stage process for rigid alignment and expression estimation to remove the erroneous markers around the jaw; however, such an approach may then overfit to the potentially inaccurate mouth markers.
We note that such concerns will gradually become less prominent as these networks improve.

\subsection{Optical Flow Infill}

\begin{figure}[b]
\centering
\begin{subfigure}[b]{\dimexpr0.28\linewidth+20pt\relax}
    \makebox[20pt]{\raisebox{40pt}{\rotatebox[origin=c]{90}{Optical Flow}}}%
    \includegraphics[width=\dimexpr\linewidth-20pt\relax]{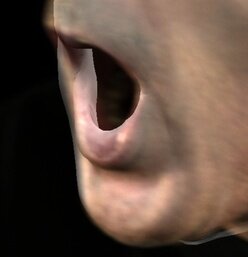}
    \makebox[20pt]{\raisebox{40pt}{\rotatebox[origin=c]{90}{Interpolation}}}%
    \includegraphics[width=\dimexpr\linewidth-20pt\relax]{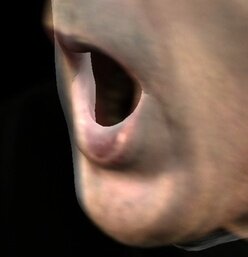}
    \caption*{1143}
\end{subfigure}
\begin{subfigure}[b]{0.28\linewidth}
    \includegraphics[width=\linewidth]{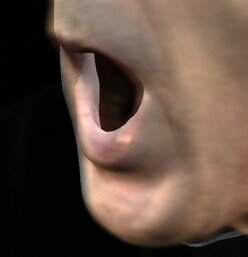}
    \includegraphics[width=\linewidth]{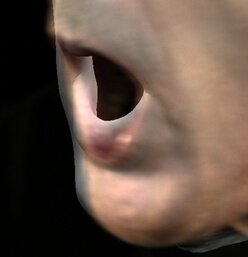}
    \caption*{1144}
\end{subfigure}
\begin{subfigure}[b]{0.28\linewidth}
    \includegraphics[width=\linewidth]{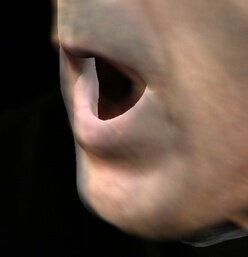}
    \includegraphics[width=\linewidth]{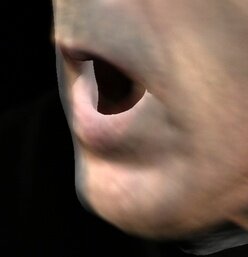}
    \caption*{1145}
\end{subfigure}
\hfill
\caption{Using optical flow to interpolate missing frames produces results that better match the plate than simple linear interpolation.
Notice how the lips and the face boundary match better in the optical flow results (particularly in frame \num{1144}) than in the simple interpolation results.}
\label{fig:oflow_vs_interp}
\end{figure}

Consider, for example, Figure \ref{fig:oflow_infill} where frames \num{1142} and \num{1146} were solved for successfully and we wish to fill frames \num{1143}, \num{1144}, and \num{1145}.
We visualize the optical flow fields using the coloring scheme of \cite{baker2011database}.
We adopt our proposed approach from Section \ref{sec:optical_flow} whereby the parameters of frames \num{1143}, \num{1144}, and \num{1145} are first solved for sequentially starting from frame \num{1142}.
Then, the frames are solved again in reverse order starting from frame \num{1146}.
This back-and-forth process which can be repeated multiple times ensures that the infilled frames at the end of the sequence have not accumulated so much error that they no longer match the other known frame.

Using optical flow information is preferable to using simple interpolation as it is able to more accurately capture any nonlinear motion in the captured images (\eg the mouth staying open and then suddenly closing).
We compare the results of our approach of using optical flow to using linear interpolation for $t$ and $w$ and spherical linear interpolation for $\theta$ in Figure \ref{fig:oflow_vs_interp}.

\begin{figure}[t!]
\centering
\begin{subfigure}[b]{\dimexpr0.32\linewidth+10pt\relax}
    \makebox[10pt]{\raisebox{50pt}{\rotatebox[origin=c]{90}{Camera A}}}%
    \includegraphics[width=\dimexpr\linewidth-10pt\relax]{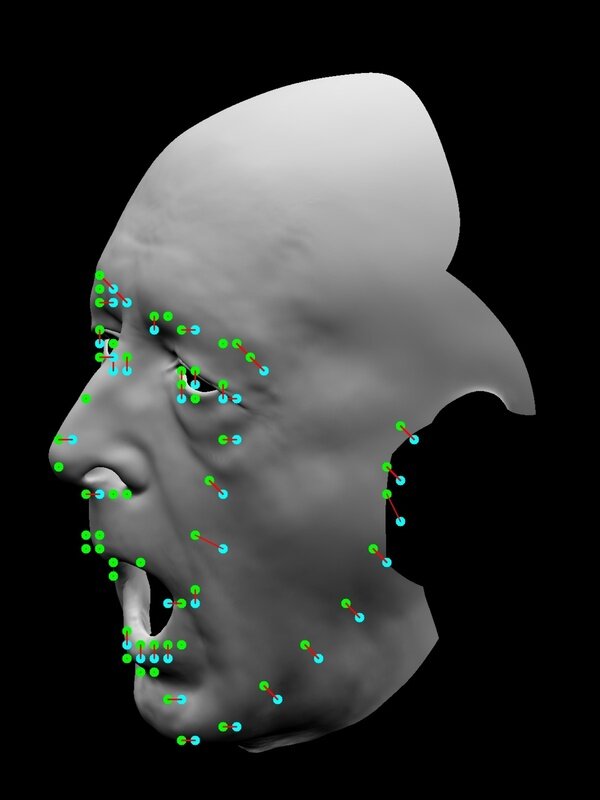}
    \makebox[10pt]{\raisebox{50pt}{\rotatebox[origin=c]{90}{Camera B}}}%
    \includegraphics[width=\dimexpr\linewidth-10pt\relax]{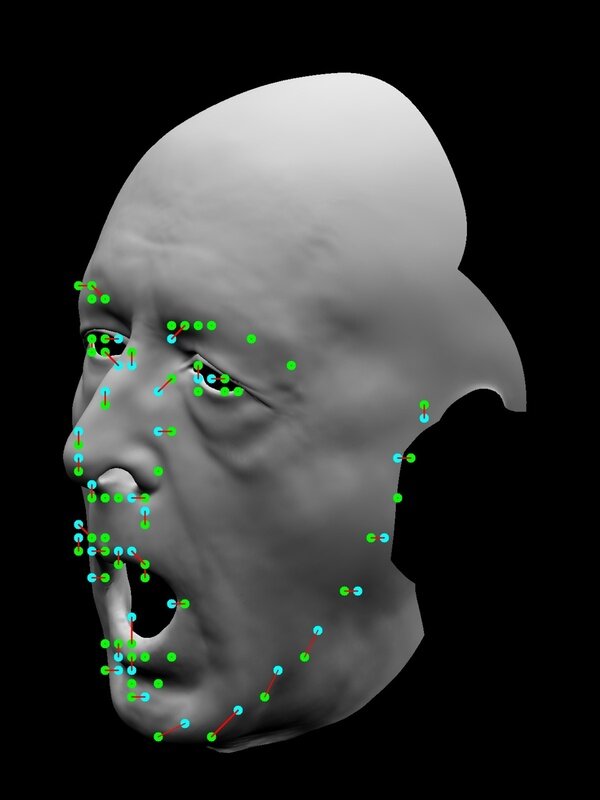}
    \caption{Render}
\end{subfigure}
\begin{subfigure}[b]{0.32\linewidth}
	\includegraphics[width=\linewidth]{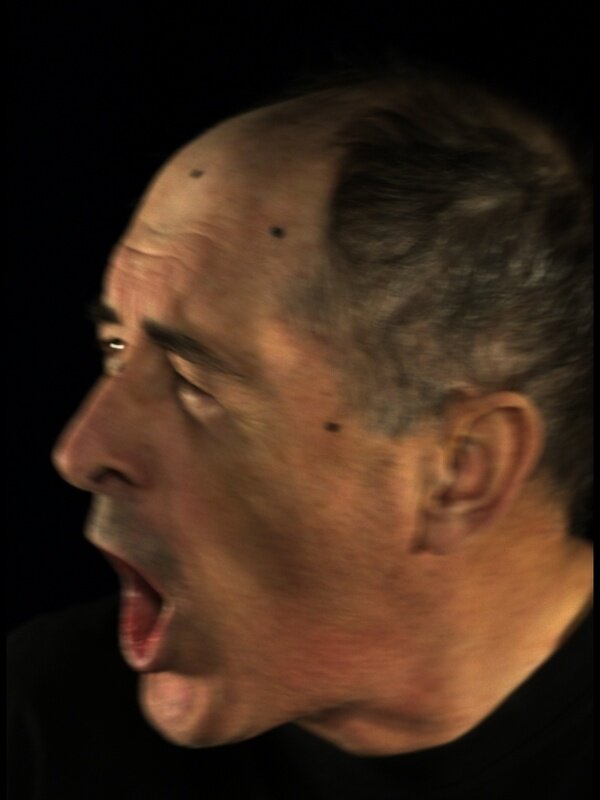}
	\includegraphics[width=\linewidth]{figures/plate/plate_1141.jpg}
    \caption{Image}
\end{subfigure}
\hfill
\caption{We trivially extend Sections \ref{sec:rigid_results} and \ref{sec:expression_results} to handle landmarks from two cameras by simply combining the objective functions.}
\label{fig:stereo}
\end{figure}

\begin{figure}[b!]
\centering
\begin{subfigure}[b]{\dimexpr0.22\linewidth+10pt\relax}
    \makebox[10pt]{\raisebox{35pt}{\rotatebox[origin=c]{90}{1132}}}%
    \includegraphics[width=\dimexpr\linewidth-10pt\relax]{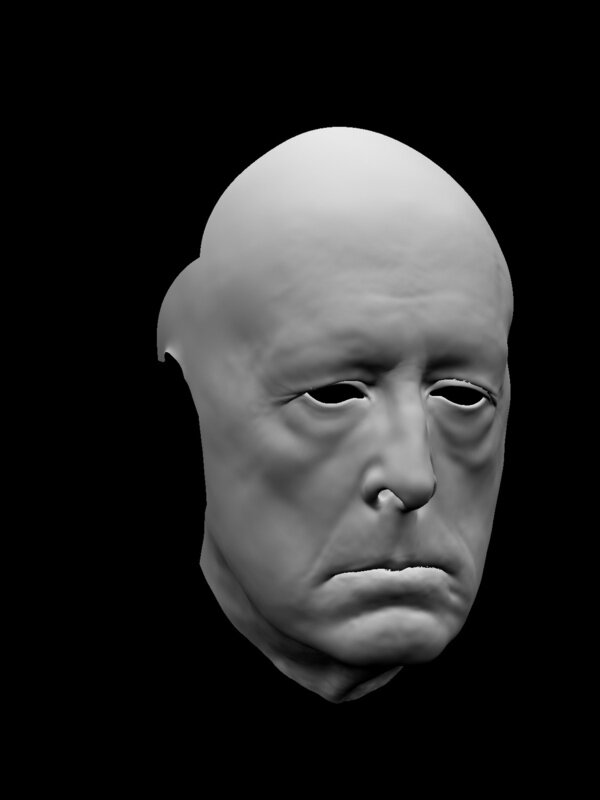}
    \makebox[10pt]{\raisebox{35pt}{\rotatebox[origin=c]{90}{1134}}}%
    \includegraphics[width=\dimexpr\linewidth-10pt\relax]{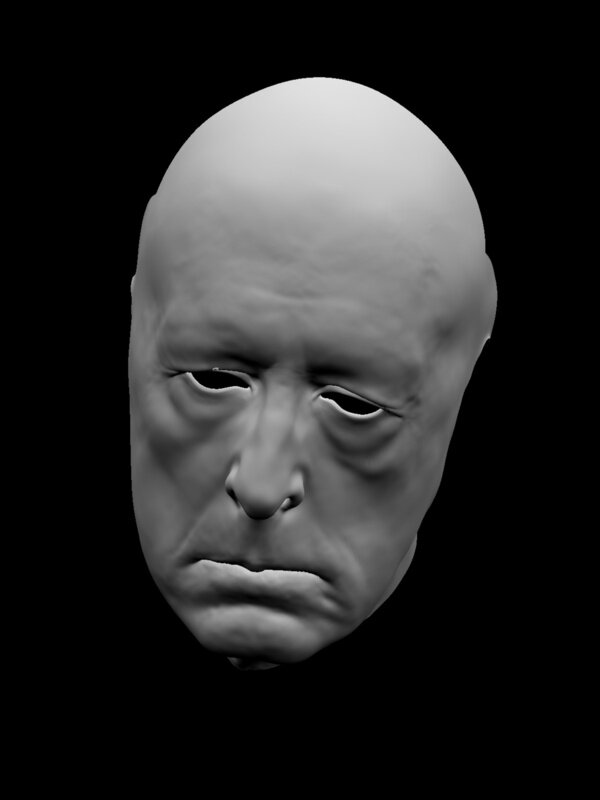}
    \makebox[10pt]{\raisebox{35pt}{\rotatebox[origin=c]{90}{1141}}}%
    \includegraphics[width=\dimexpr\linewidth-10pt\relax]{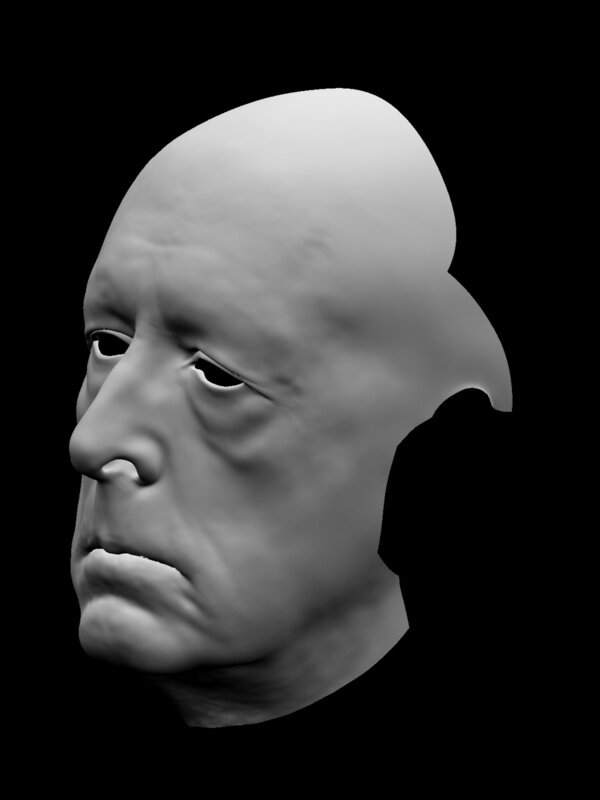}
    \makebox[10pt]{\raisebox{35pt}{\rotatebox[origin=c]{90}{1171}}}%
    \includegraphics[width=\dimexpr\linewidth-10pt\relax]{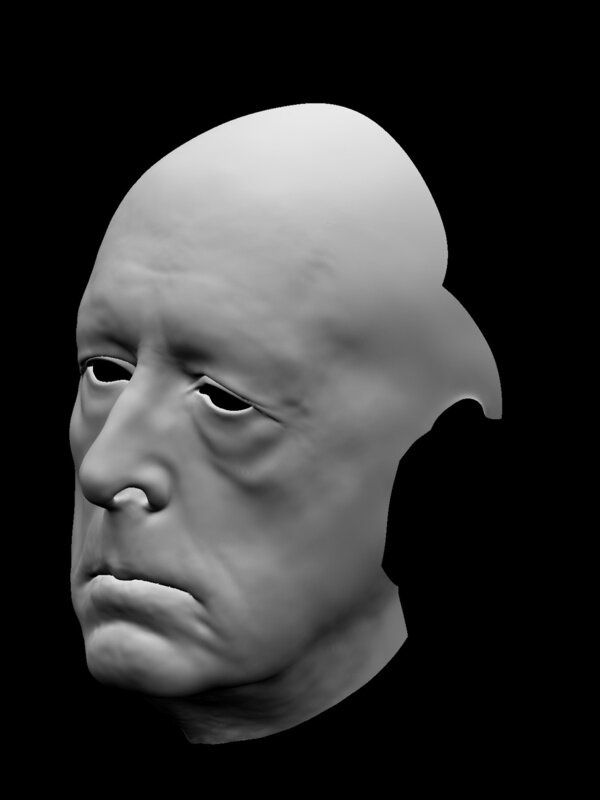}
    \caption{Monocular}
\end{subfigure}
\begin{subfigure}[b]{0.22\linewidth}
    \includegraphics[width=\linewidth]{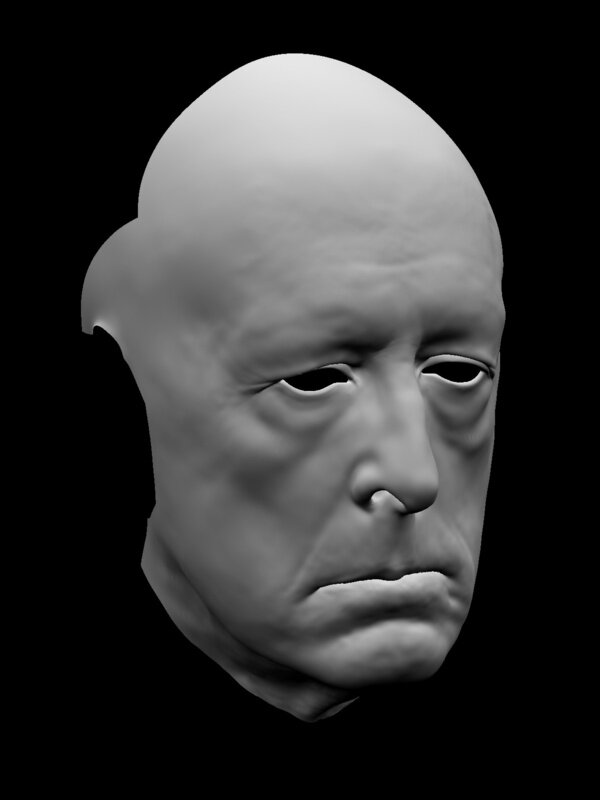}
    \includegraphics[width=\linewidth]{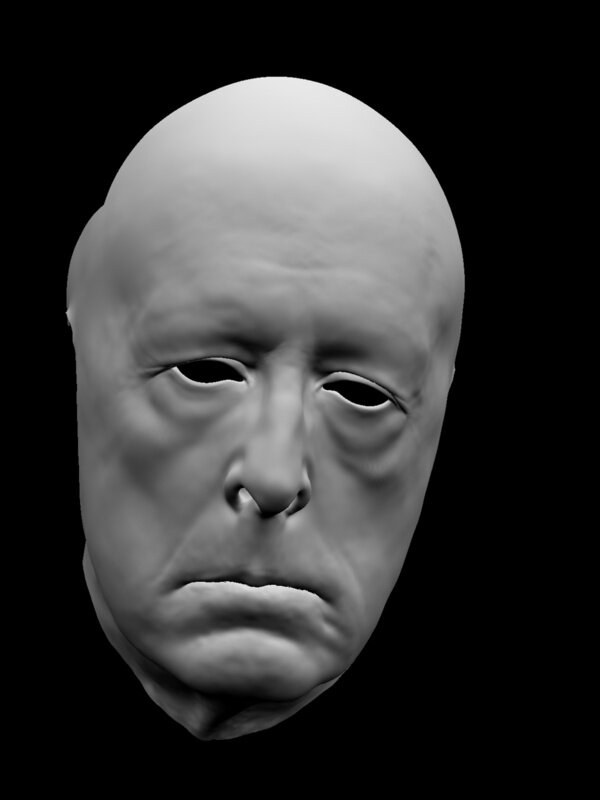}
    \includegraphics[width=\linewidth]{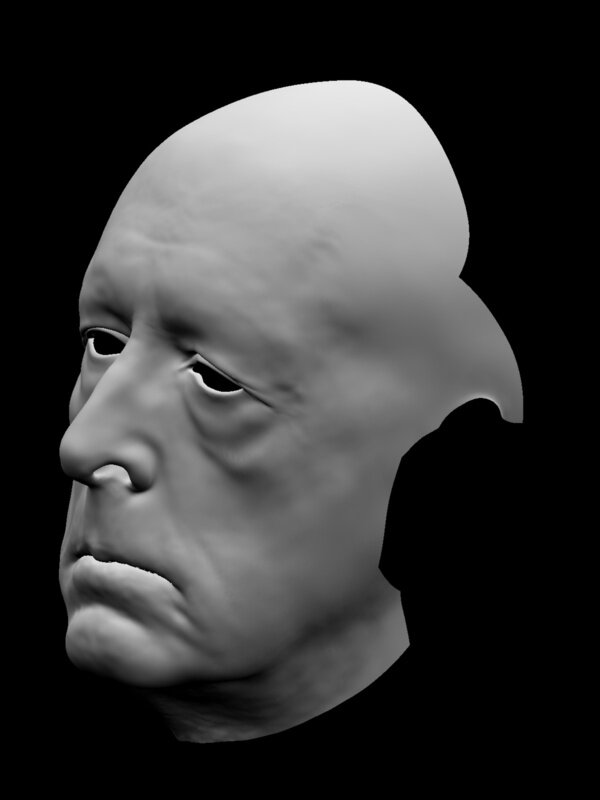}
    \includegraphics[width=\linewidth]{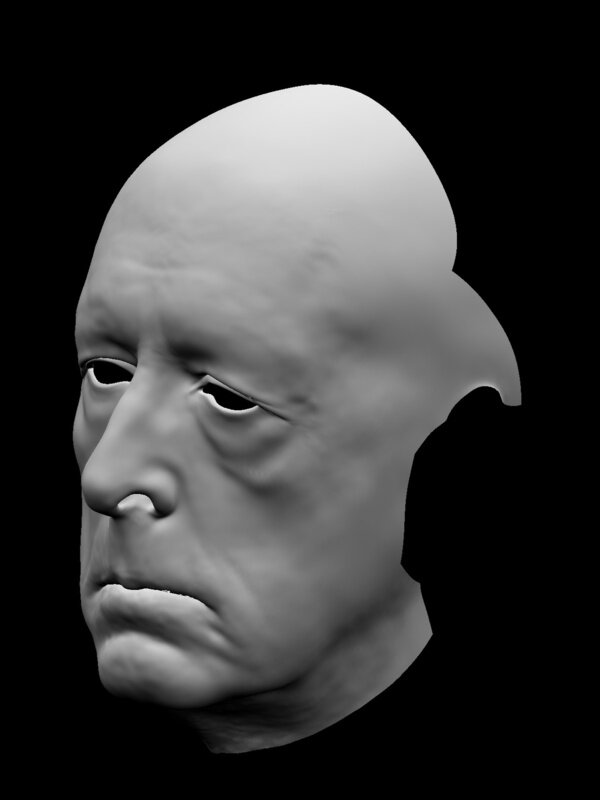}
    \caption{Stereo}
\end{subfigure}
\begin{subfigure}[b]{0.22\linewidth}
    \includegraphics[width=\linewidth]{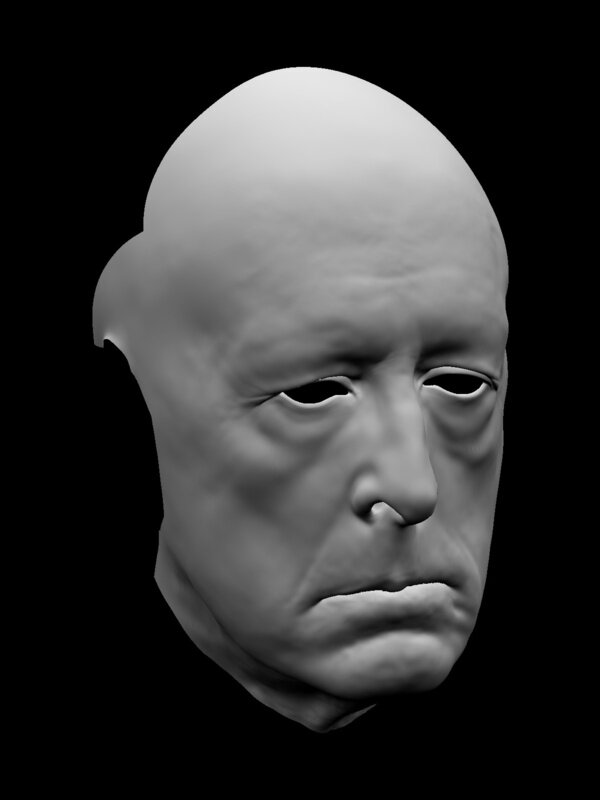}
    \includegraphics[width=\linewidth]{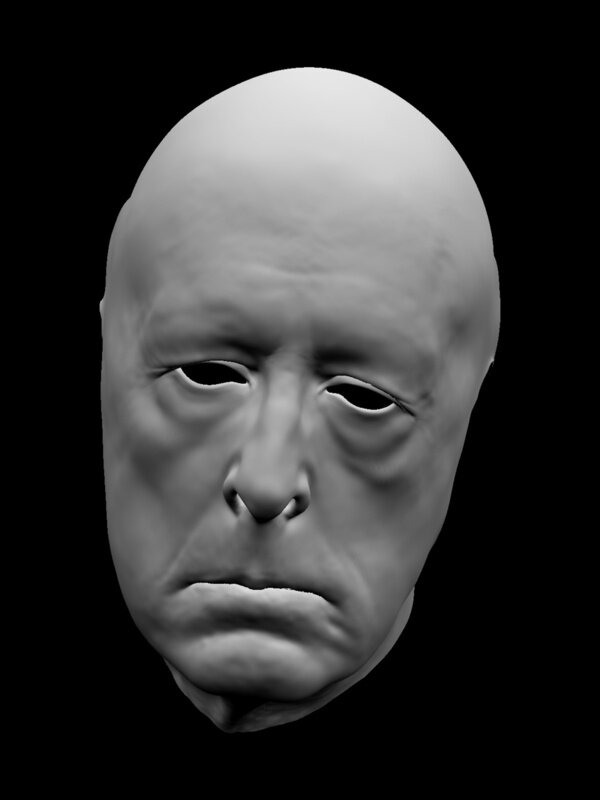}
    \includegraphics[width=\linewidth]{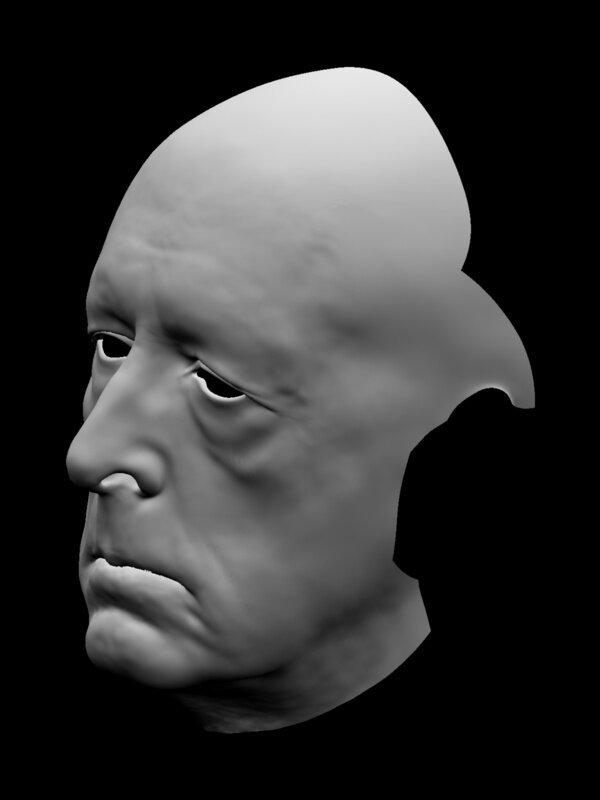}
    \includegraphics[width=\linewidth]{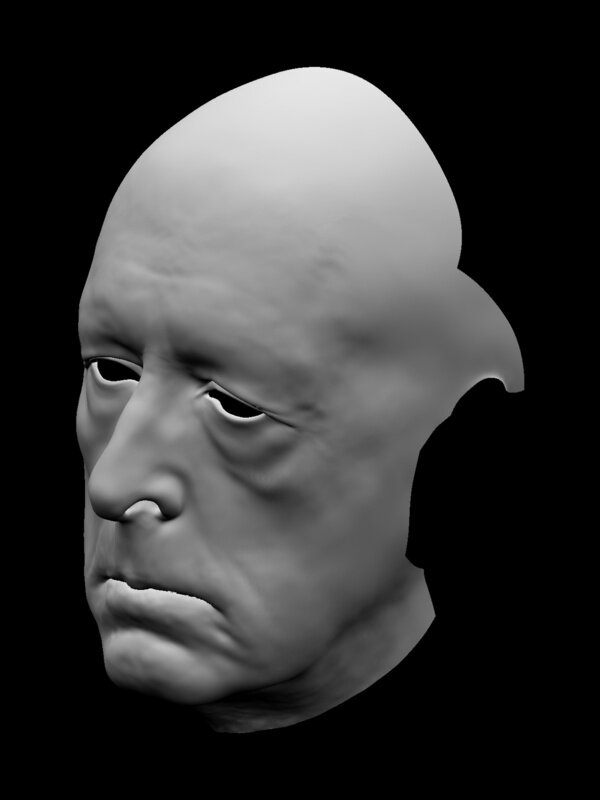}
    \caption{Manual}
\end{subfigure}
\begin{subfigure}[b]{0.22\linewidth}
    \includegraphics[width=\linewidth]{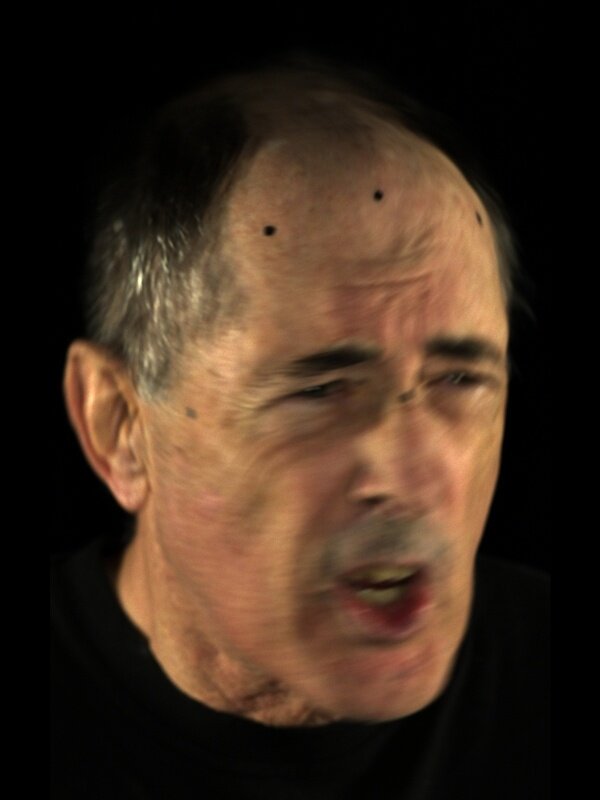}
    \includegraphics[width=\linewidth]{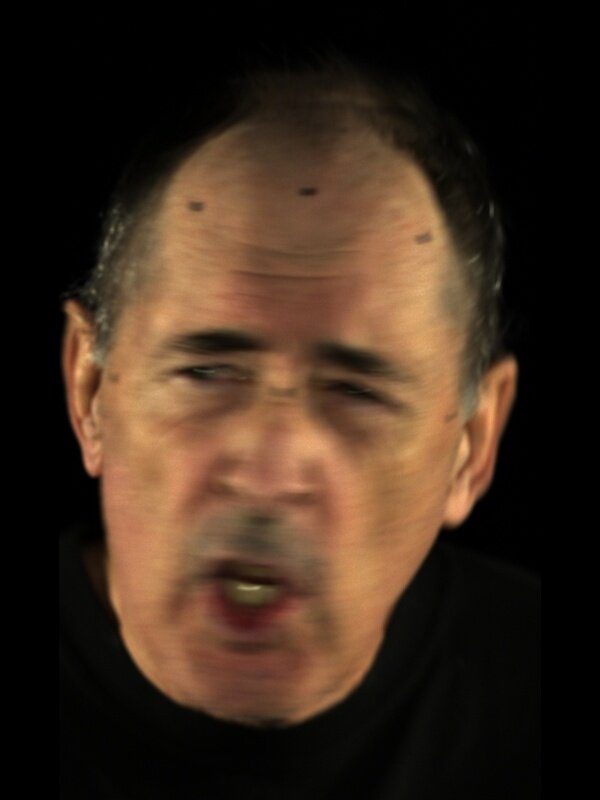}
    \includegraphics[width=\linewidth]{figures/plate/plate_1141.jpg}
    \includegraphics[width=\linewidth]{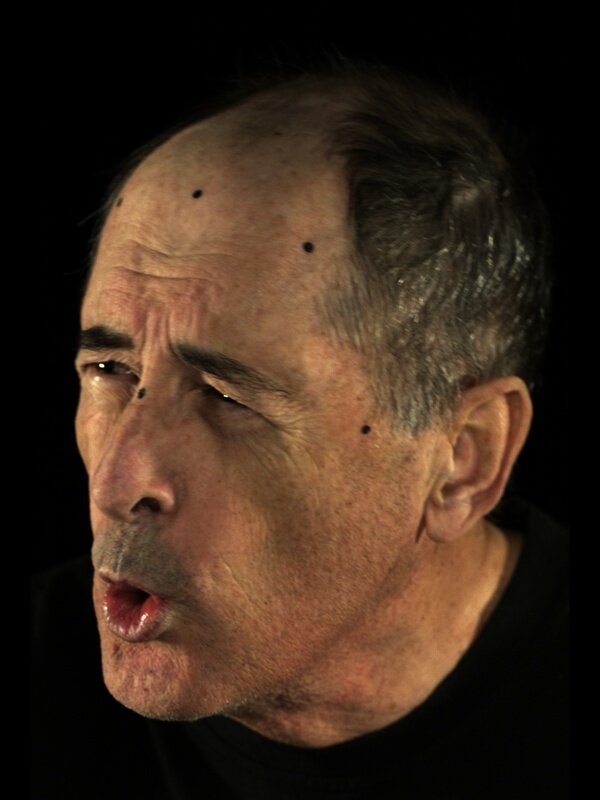}
    \caption{Target}
\end{subfigure}
\hfill
\caption{A comparison of the rigid alignment computed manually by a skilled artist versus the rigid aligment computed by our approach in both the monocular and stereo case.}
\label{fig:compare_artist}
\end{figure}

\subsection{Multi-Camera}

\begin{figure*}[t]
\centering
\begin{subfigure}[b]{\dimexpr0.10\linewidth+20pt\relax}
    \makebox[20pt]{\raisebox{30pt}{\rotatebox[origin=c]{90}{Result}}}%
    \includegraphics[width=\dimexpr\linewidth-20pt\relax]{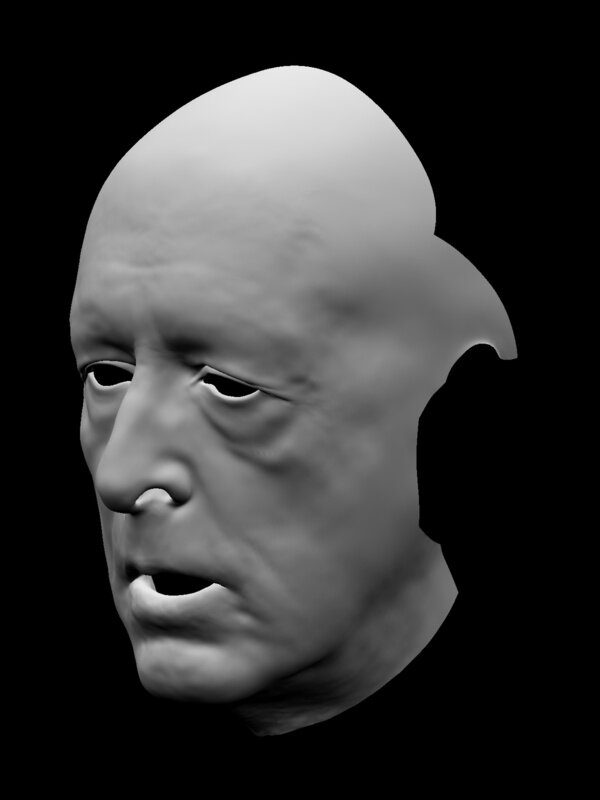}
    \makebox[20pt]{\raisebox{30pt}{\rotatebox[origin=c]{90}{Smoothed}}}%
    \includegraphics[width=\dimexpr\linewidth-20pt\relax]{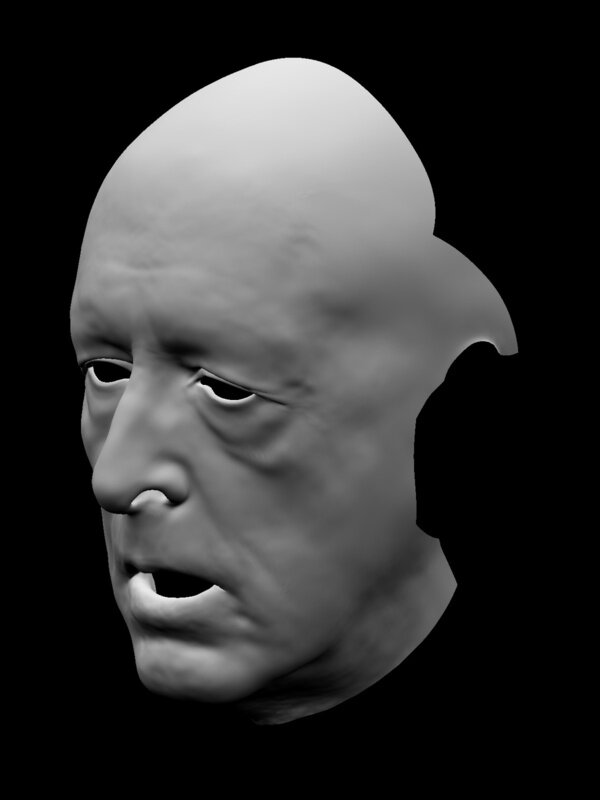}
    \makebox[20pt]{\raisebox{30pt}{\rotatebox[origin=c]{90}{Target}}}%
    \includegraphics[width=\dimexpr\linewidth-20pt\relax]{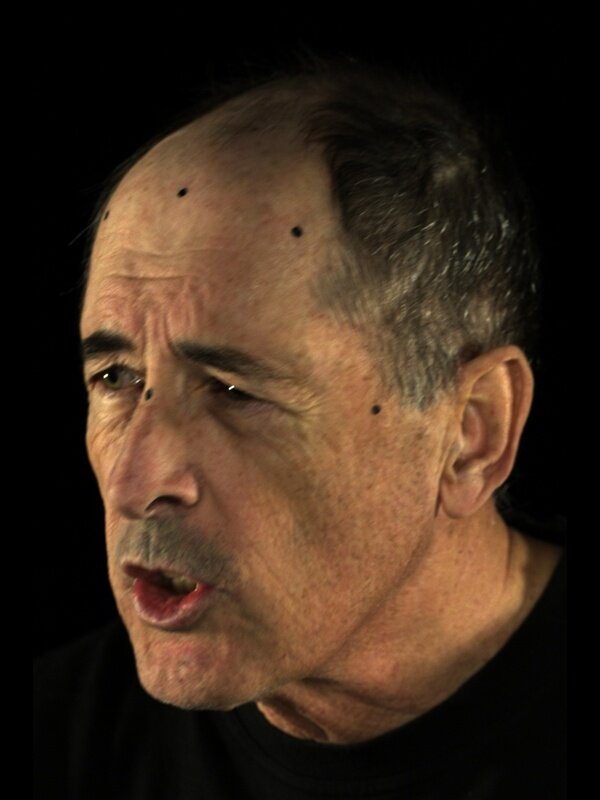}
    \caption*{1111}
\end{subfigure}
\begin{subfigure}[b]{0.10\linewidth}
    \includegraphics[width=\linewidth]{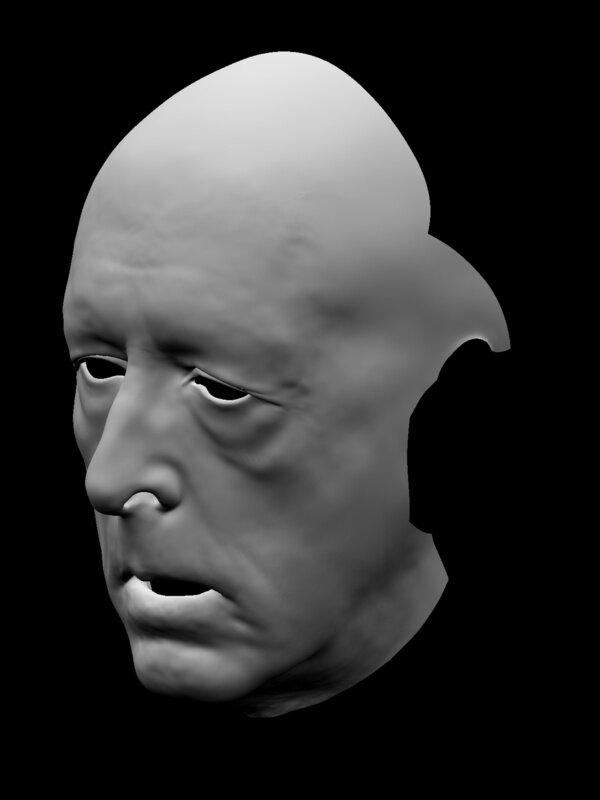}
    \includegraphics[width=\linewidth]{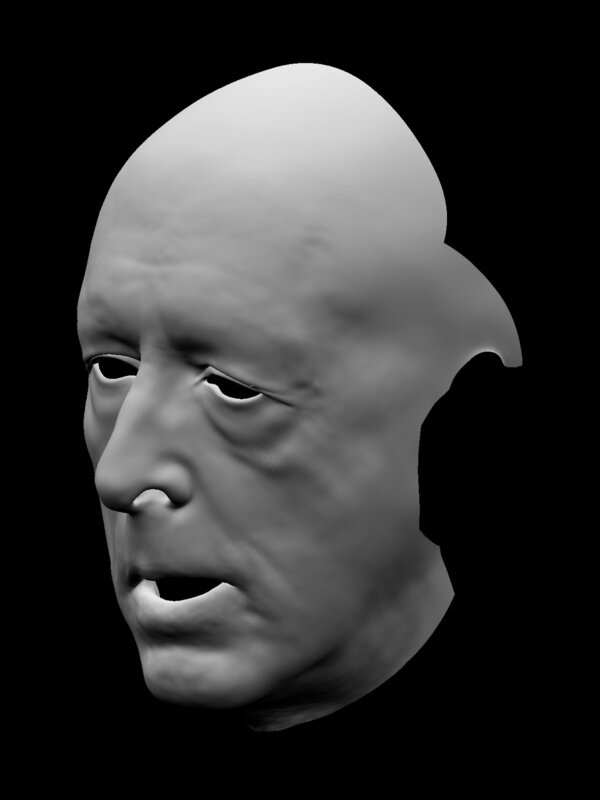}
    \includegraphics[width=\linewidth]{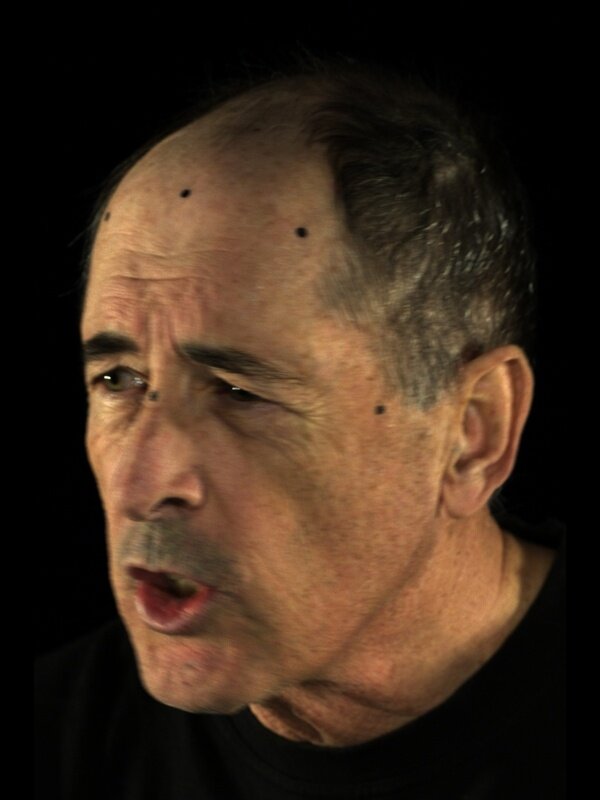}
    \caption*{1112}
\end{subfigure}
\begin{subfigure}[b]{0.10\linewidth}
    \includegraphics[width=\linewidth]{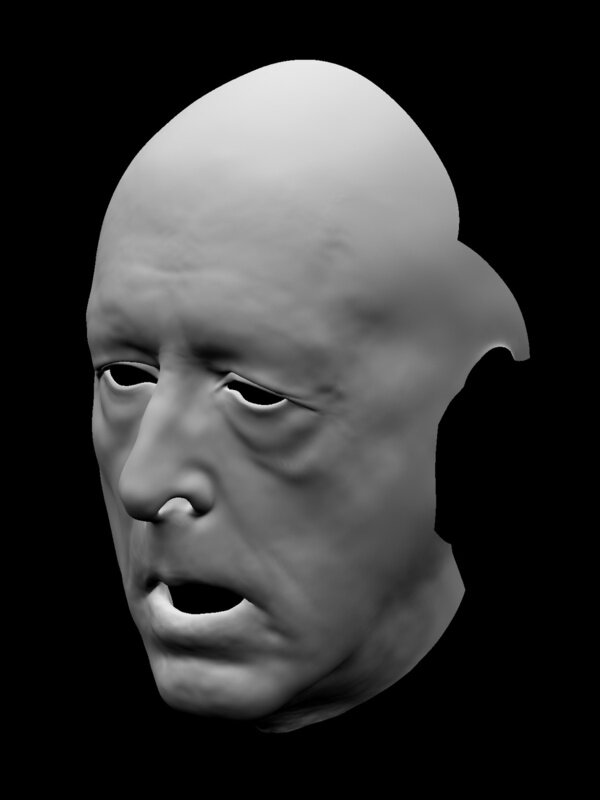}
    \includegraphics[width=\linewidth]{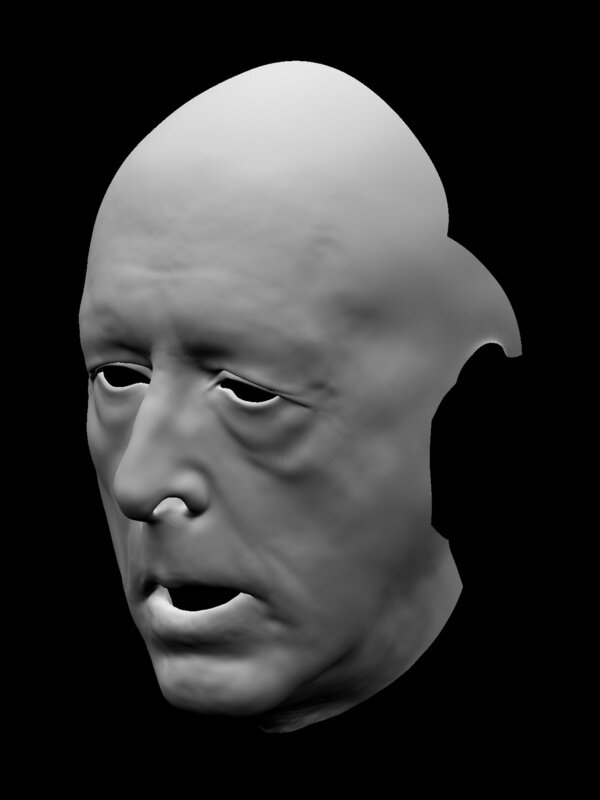}
    \includegraphics[width=\linewidth]{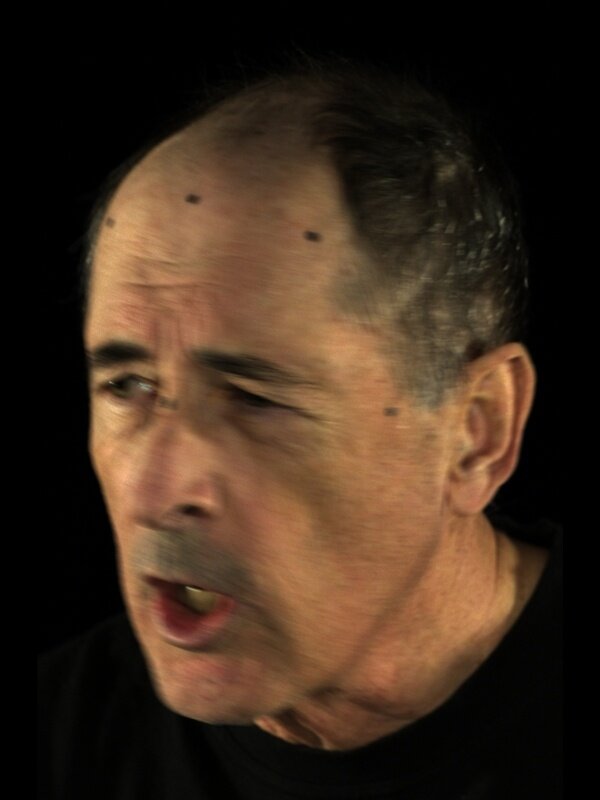}
    \caption*{1113}
\end{subfigure}
\begin{subfigure}[b]{0.10\linewidth}
    \includegraphics[width=\linewidth]{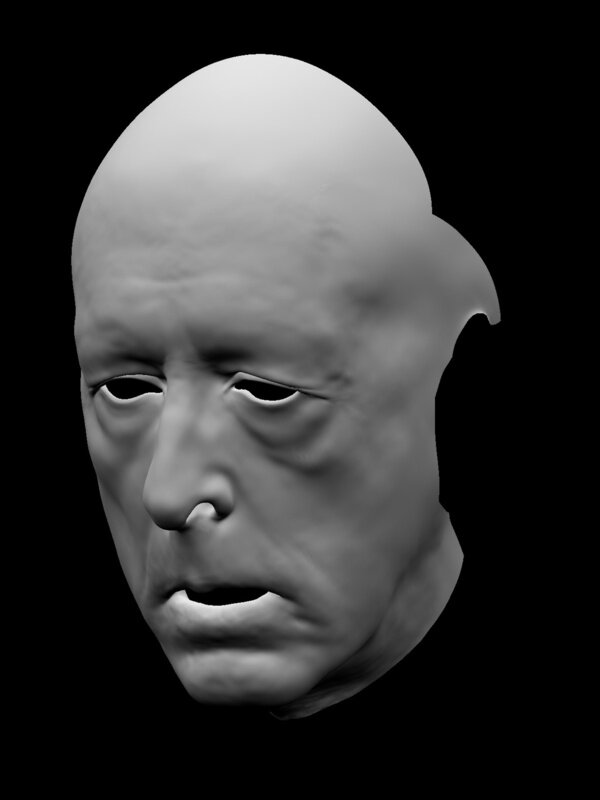}
    \includegraphics[width=\linewidth]{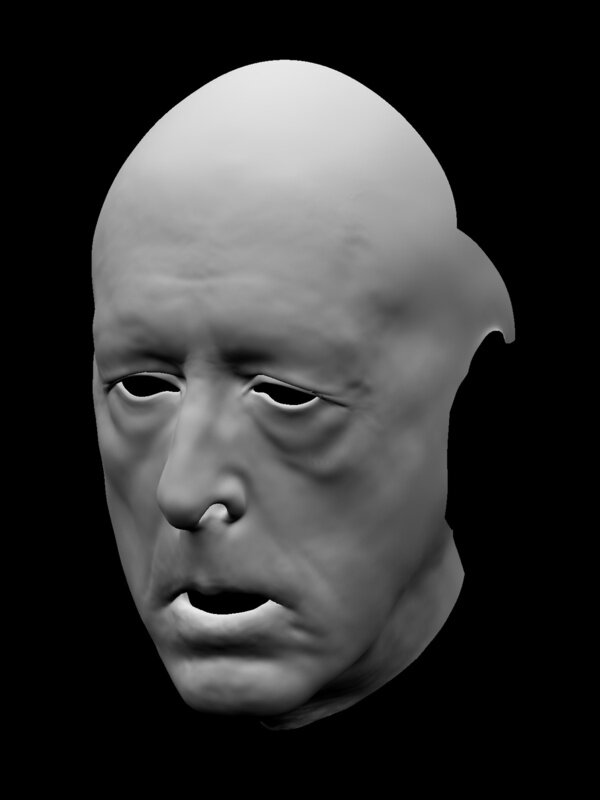}
    \includegraphics[width=\linewidth]{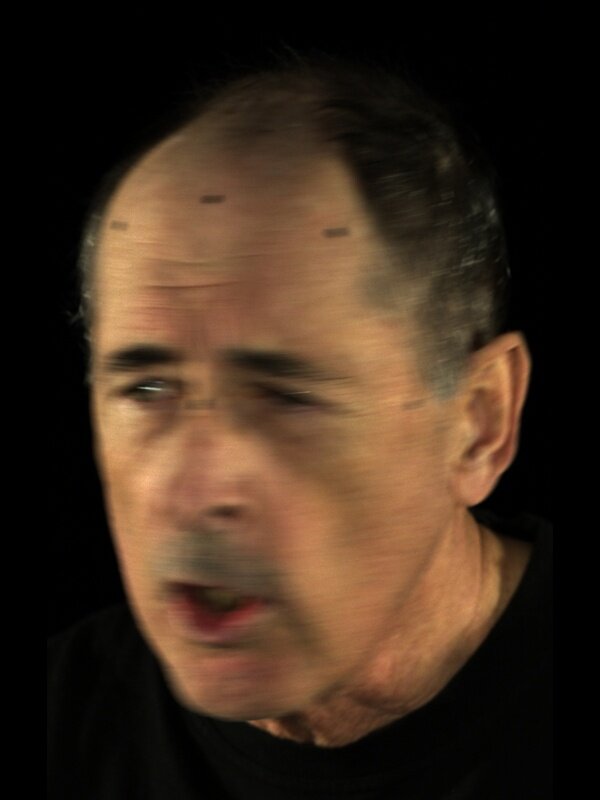}
    \caption*{1114}
\end{subfigure}
\begin{subfigure}[b]{0.10\linewidth}
    \includegraphics[width=\linewidth]{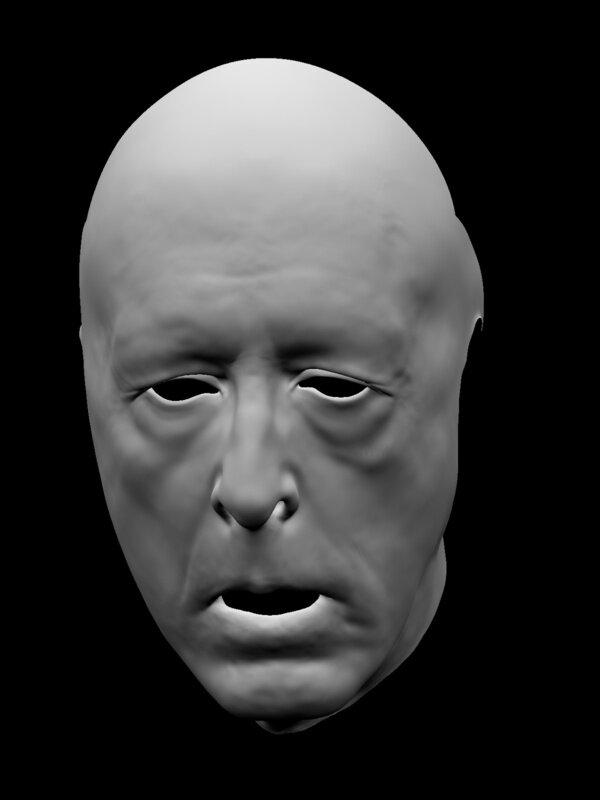}
    \includegraphics[width=\linewidth]{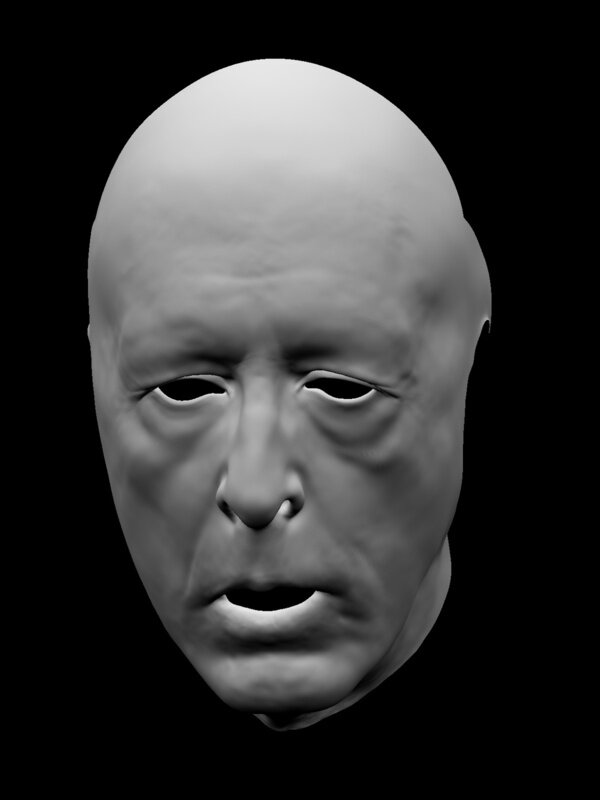}
    \includegraphics[width=\linewidth]{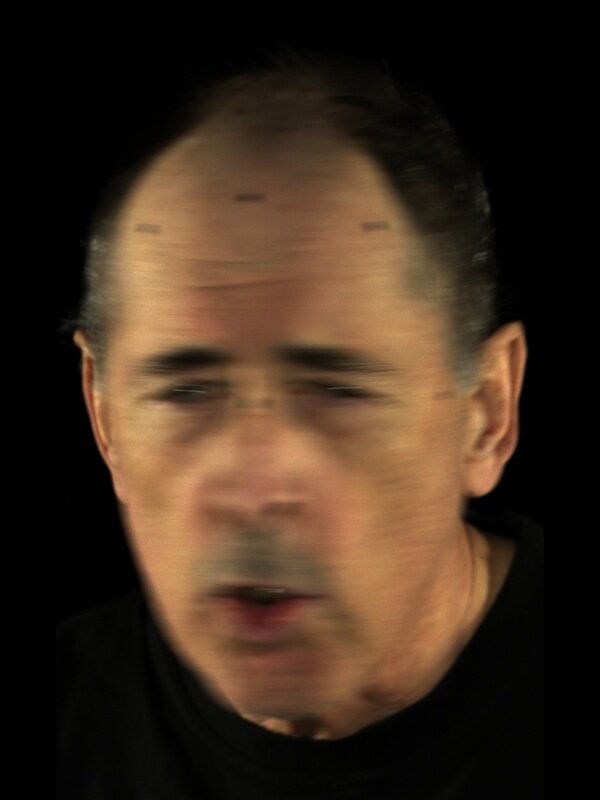}
    \caption*{1115}
\end{subfigure}
\begin{subfigure}[b]{0.10\linewidth}
    \includegraphics[width=\linewidth]{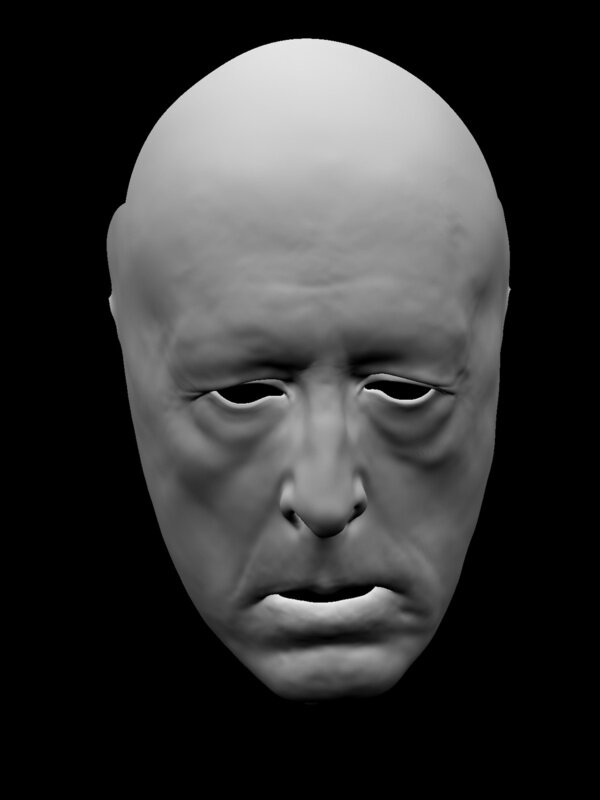}
    \includegraphics[width=\linewidth]{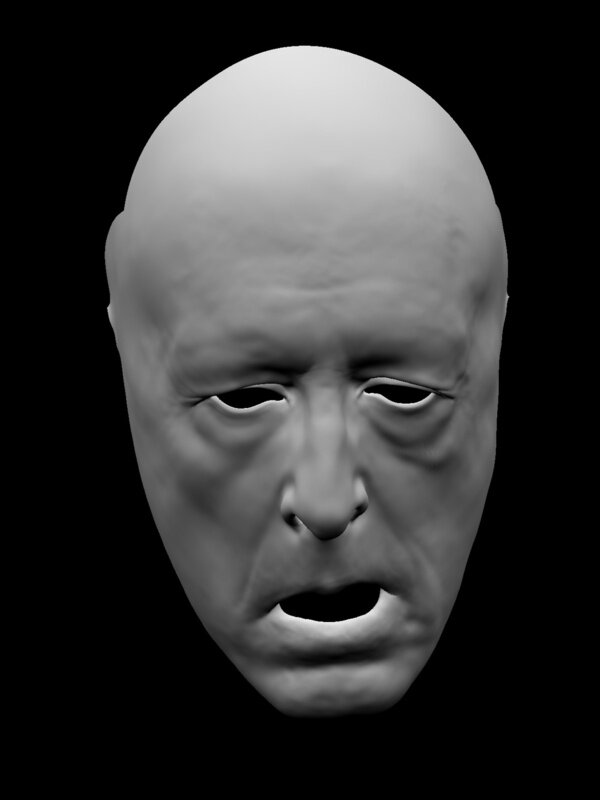}
    \includegraphics[width=\linewidth]{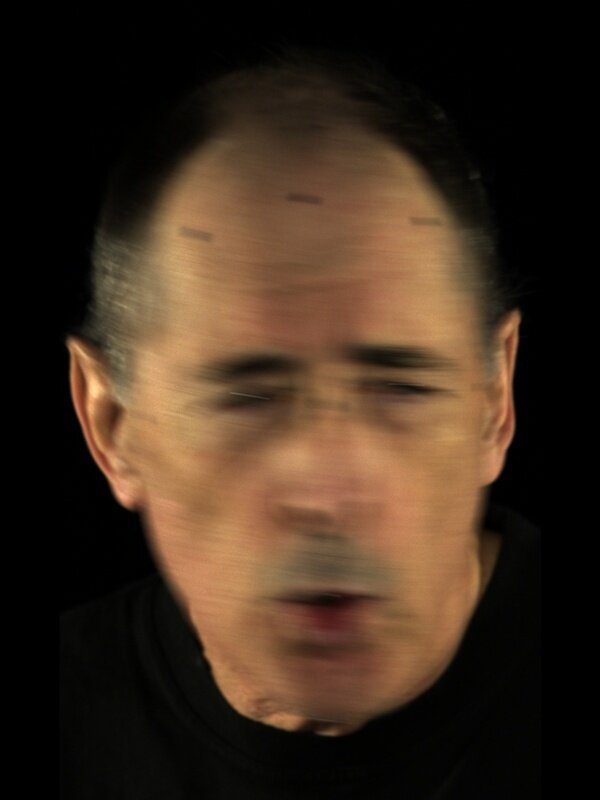}
    \caption*{1116}
\end{subfigure}
\begin{subfigure}[b]{0.10\linewidth}
    \includegraphics[width=\linewidth]{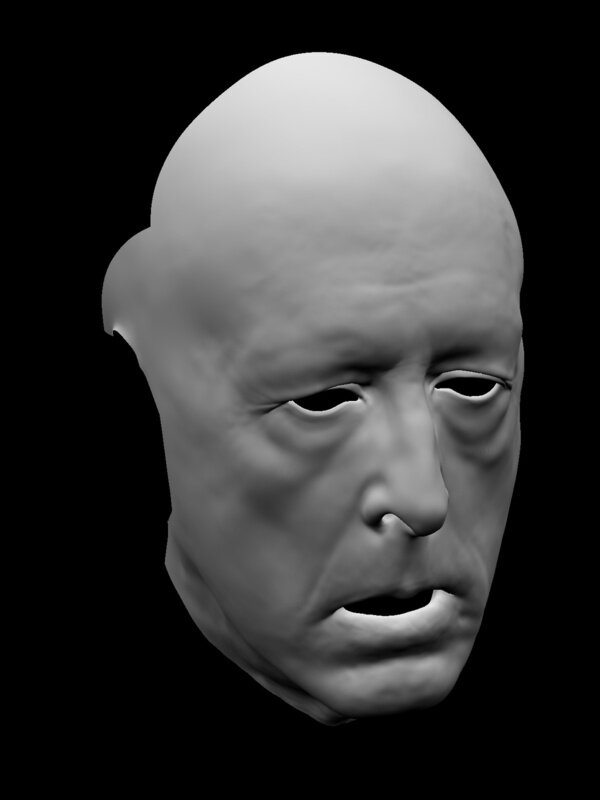}
    \includegraphics[width=\linewidth]{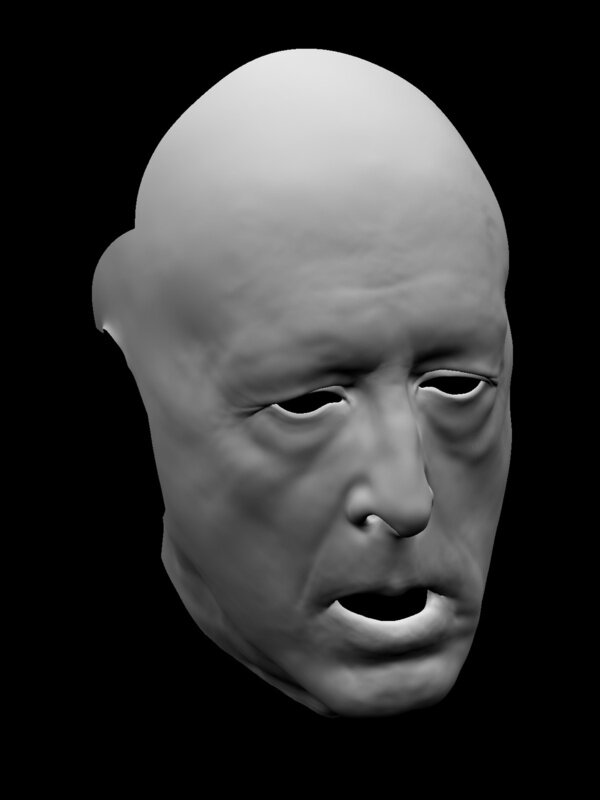}
    \includegraphics[width=\linewidth]{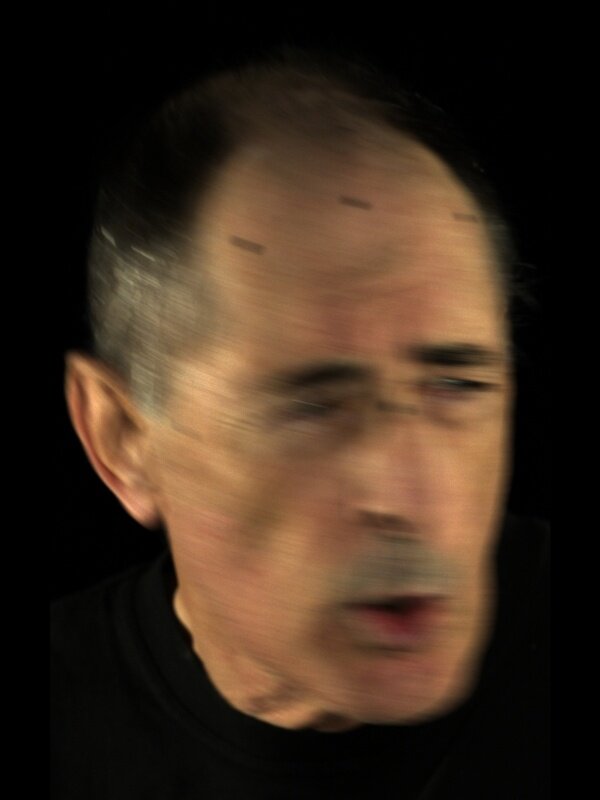}
    \caption*{1117}
\end{subfigure}
\begin{subfigure}[b]{0.10\linewidth}
    \includegraphics[width=\linewidth]{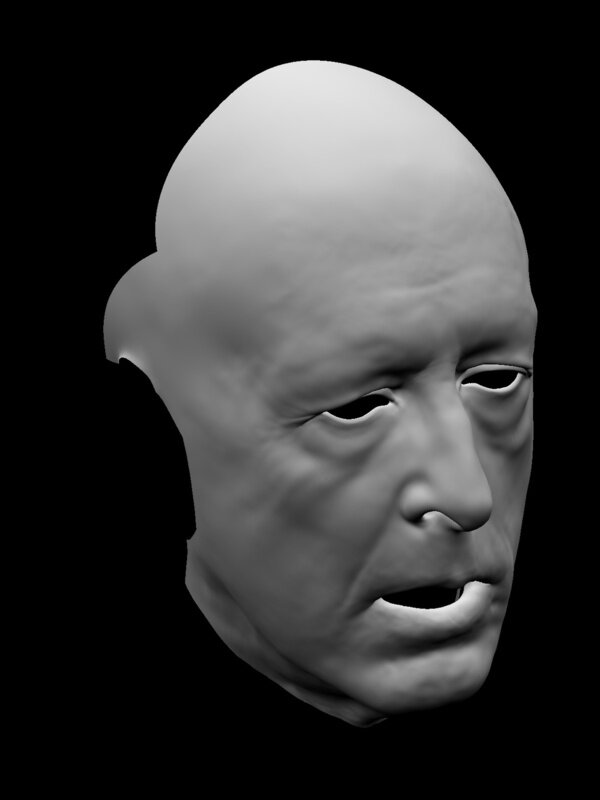}
    \includegraphics[width=\linewidth]{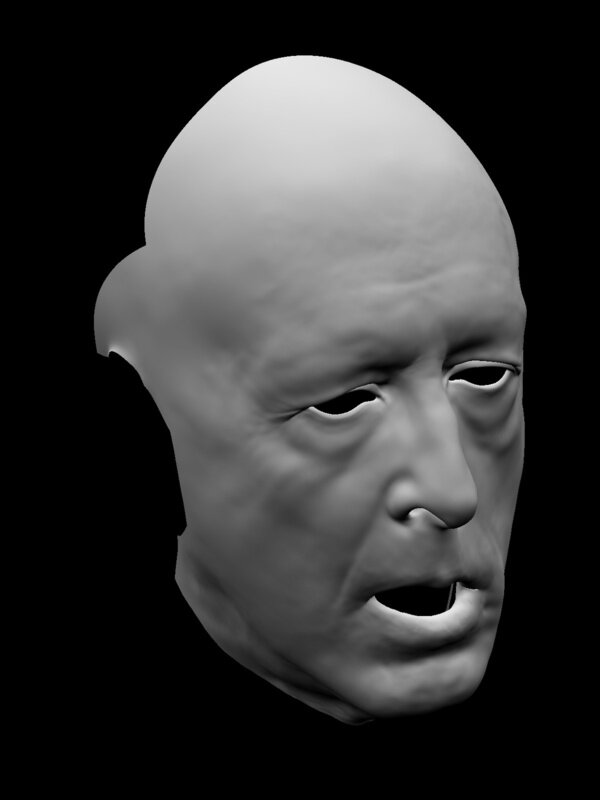}
    \includegraphics[width=\linewidth]{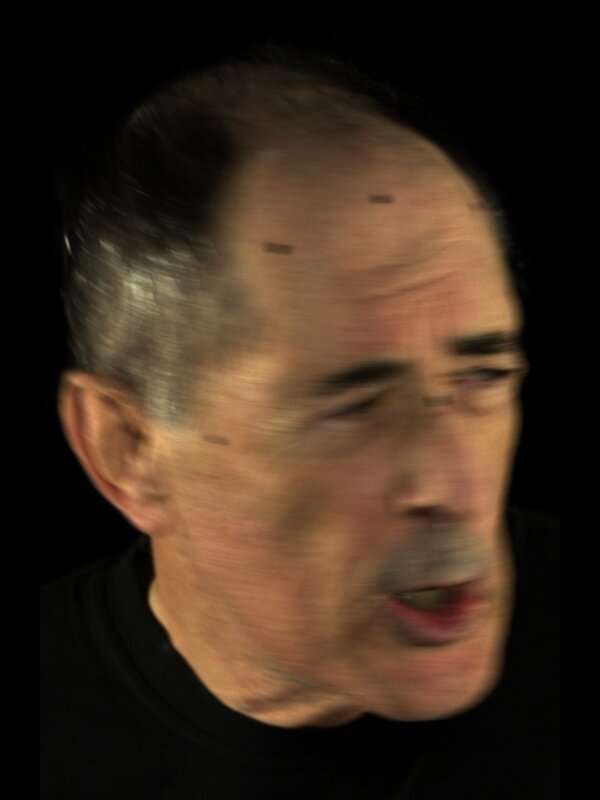}
    \caption*{1118}
\end{subfigure}
\begin{subfigure}[b]{0.10\linewidth}
    \includegraphics[width=\linewidth]{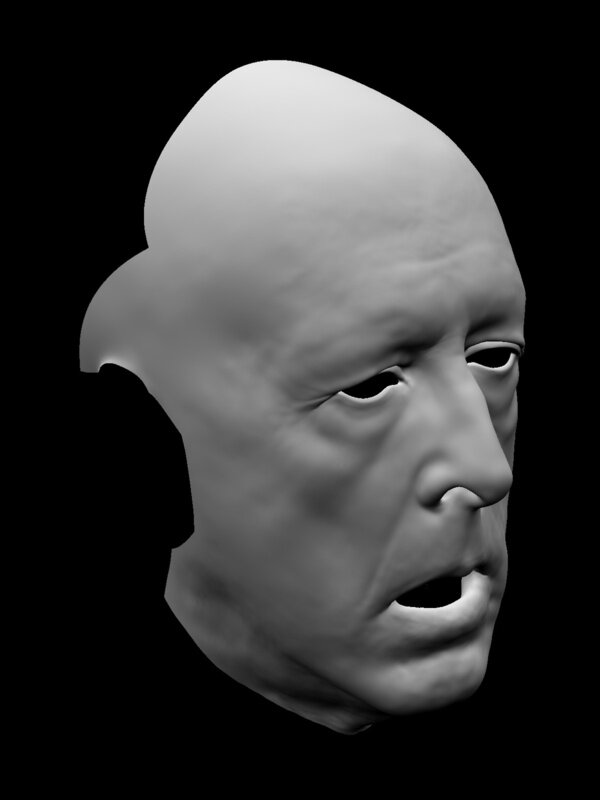}
    \includegraphics[width=\linewidth]{figures/temporal/temporal_crop_1118.jpg}
    \includegraphics[width=\linewidth]{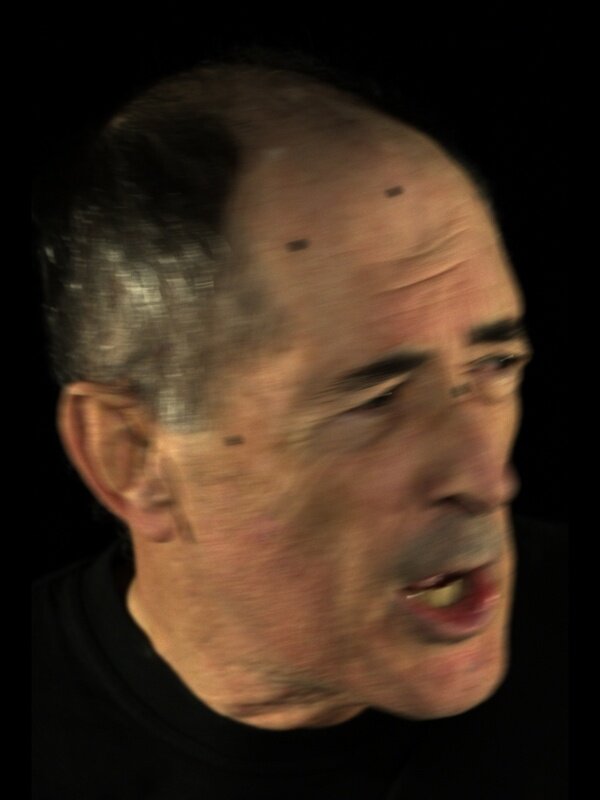}
    \caption*{1119}
\end{subfigure}
\hfill
\caption{Smoothing the performance using the captured images' optical flow fields between adjacent frames produces more temporally consistent results.
Similar to the optical flow fill-in stage, any errors in the initial landmark estimation will propagate into the optical flow solve, \eg frame 1116 where the jaw opens in an attempt to smooth the rigid of the face.}
\label{fig:temporal_refinement}
\end{figure*}

Our approach can trivially be extended to multiple calibrated camera viewpoints as it only entails adding another duplicate set of energy terms to the nonlinear least squares objective function.
We demonstrate the effectiveness of this approach by applying our approach from Sections \ref{sec:rigid_results} and \ref{sec:expression_results} to the same performance captured using an identical ARRI Alexa XT Studio from another viewpoint.
See Figure \ref{fig:stereo}.

\begin{figure}[b]
\centering
    \includegraphics[width=0.8\linewidth]{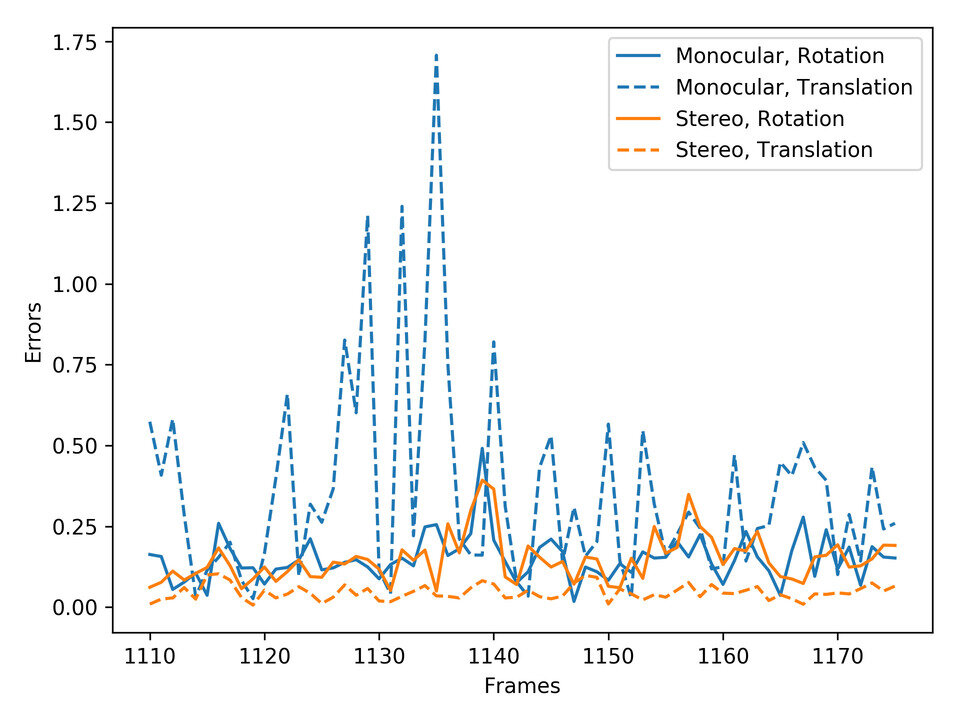}
\hfill
\caption{Assuming the manually done rigid alignment is the ``ground truth,'' we measure the errors for rigid parameters for the monocular and stereo case.}
\label{fig:compare_artist_quant}
\end{figure}

We also compare the rigid alignment estimated by our automatic method to the rigid alignment created by a skilled matchmove artist for the same performance.
The manual rigid alignment was performed by tracking the painted black dots on the face along with other manually tracked facial features.
In comparison, our rigid alignment was done using only the markers detected by 3D-FAN on both the captured images and the synthetic renders.
See Figure \ref{fig:compare_artist}.
Our approach using only features detected by 3D-FAN produces visually comparable results.
In Figure \ref{fig:compare_artist_quant}, we assume the manually done rigid alignment is the ``ground truth'' and quantitatively evaluate the rigid alignment computed by the monocular and stereo solves.
Both the monocular and stereo solves are able to recover similar rotation parameters, and the stereo solve is able to much more accurately determine the rigid translation.
We note, however, that it is unlikely that the manually done rigid alignment can be considered ``ground truth'' as it more than likely contains errors as well.

\subsection{Temporal Refinement}

As seen in the supplementary video, the facial pose and expression estimations are generally temporally inconsistent.
We adopt our proposed approach from Section \ref{sec:temporal}.
This attempts to mimic the captured temporal performance which not only helps to better match the synthetic render to the captured image but also introduces temporal consistency between renders.
While this is theoretically susceptible to noise in the optical flow field, we did not find this to be a problem.
See Figure \ref{fig:temporal_refinement}.
We explore additional methods of performing temporal refinement in the supplementary material.

\section{Conclusion and Future Work} \label{sec:conclusion}

We have proposed and demonstrated the efficacy of a fully automatic pipeline for estimating facial pose and expression using pre-trained deep networks as the objective functions in traditional nonlinear optimization.
Such an approach is advantageous as it removes the subjectivity and inconsistency of the artist.
Our approach heavily depends upon the robustness of the face detector and the facial alignment networks, and any failures in those cause the optimization to fail.
Currently, we use optical flow to fix such problematic frames, and we leave exploring methods to automatically avoid problematic areas of the search space for future work.
Furthermore, as the quality of these networks improve, our proposed approach would similarly benefit, leading to higher-fidelity results.
While we have only explored using pre-trained facial alignment and optical flow networks, using other types of networks (\eg face segmentation, face recognition, etc.) and using networks trained specifically on the vast repository of data from decades of visual effects work are exciting avenues for future work.

\section*{Acknowledgements}
Research supported in part by ONR N00014-13-1-0346, ONR N00014-17-1-2174, ARL AHPCRC W911NF-07-0027, and generous gifts from Amazon and Toyota.
In addition, we would like to thank both Reza and Behzad at ONR for supporting our efforts into computer vision and machine learning, as well as Cary Phillips and Industrial Light \& Magic for supporting our efforts into facial performance capture.
M.B. was supported in part by The VMWare Fellowship in Honor of Ole Agesen.
J.W. was supported in part by the Stanford School of Engineering Fellowship.
We would also like to thank Paul Huston for his acting.

\section*{Appendix}
\appendix
\section{Temporal Smoothing Alternatives}
\label{sec:smoothing}

\begin{figure*}[p]
\centering
\begin{subfigure}[b]{\dimexpr0.15\linewidth+20pt\relax}
    \makebox[20pt]{\raisebox{50pt}{\rotatebox[origin=c]{90}{Averaging}}}%
    \includegraphics[width=\dimexpr\linewidth-20pt\relax]{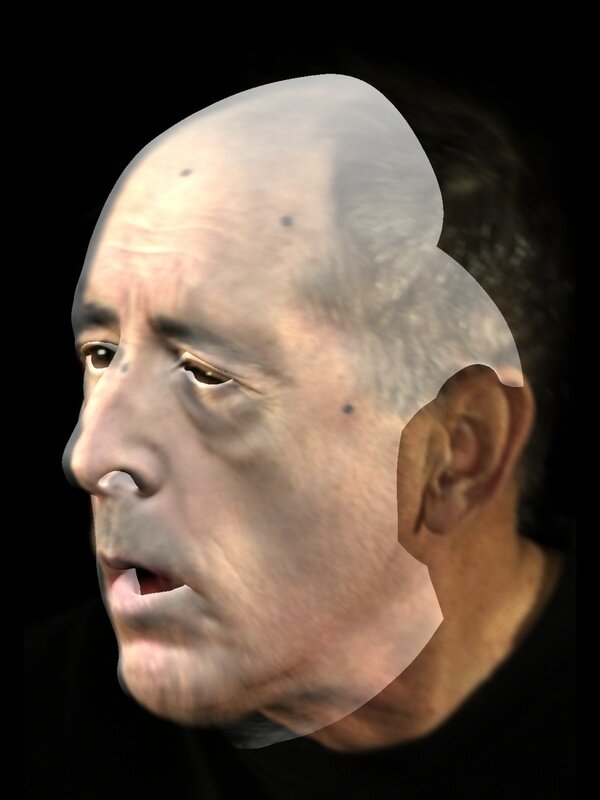}
    \makebox[20pt]{\raisebox{50pt}{\rotatebox[origin=c]{90}{Self Flow}}}%
    \includegraphics[width=\dimexpr\linewidth-20pt\relax]{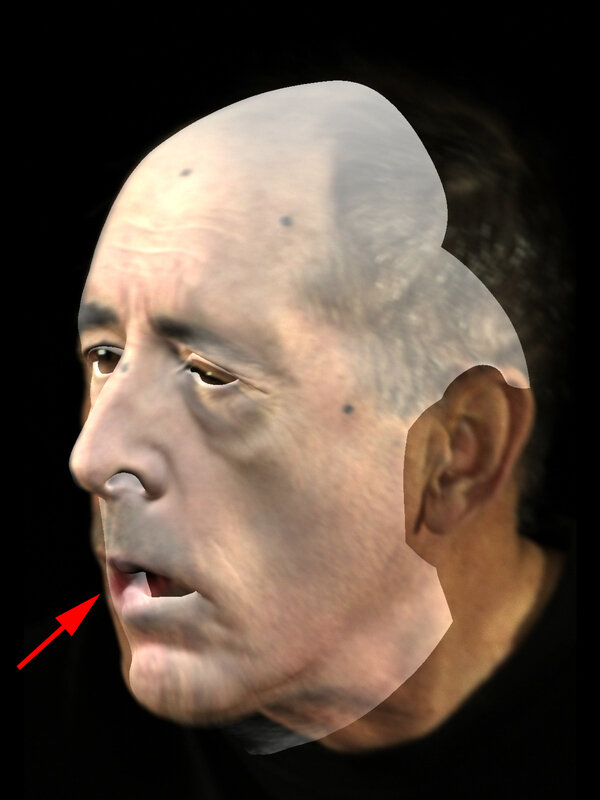}
    \makebox[20pt]{\raisebox{50pt}{\rotatebox[origin=c]{90}{Plate Flow}}}%
    \includegraphics[width=\dimexpr\linewidth-20pt\relax]{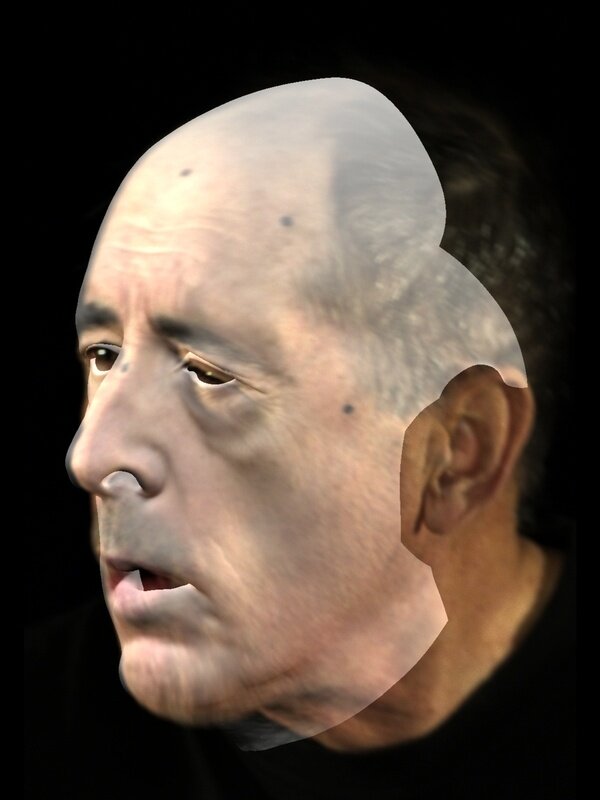}
    \makebox[20pt]{\raisebox{50pt}{\rotatebox[origin=c]{90}{Hybrid}}}%
    \includegraphics[width=\dimexpr\linewidth-20pt\relax]{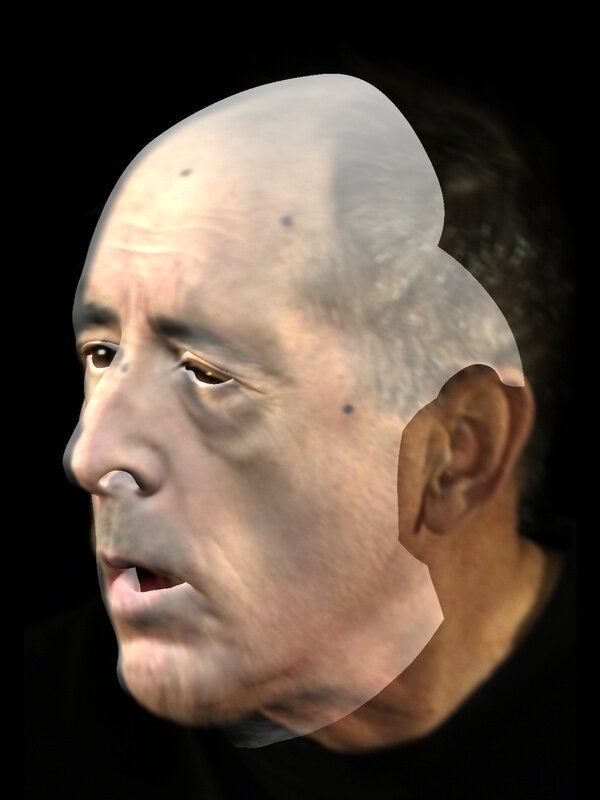}
    \caption*{1146}
\end{subfigure}
\begin{subfigure}[b]{0.15\linewidth}
    \includegraphics[width=\linewidth]{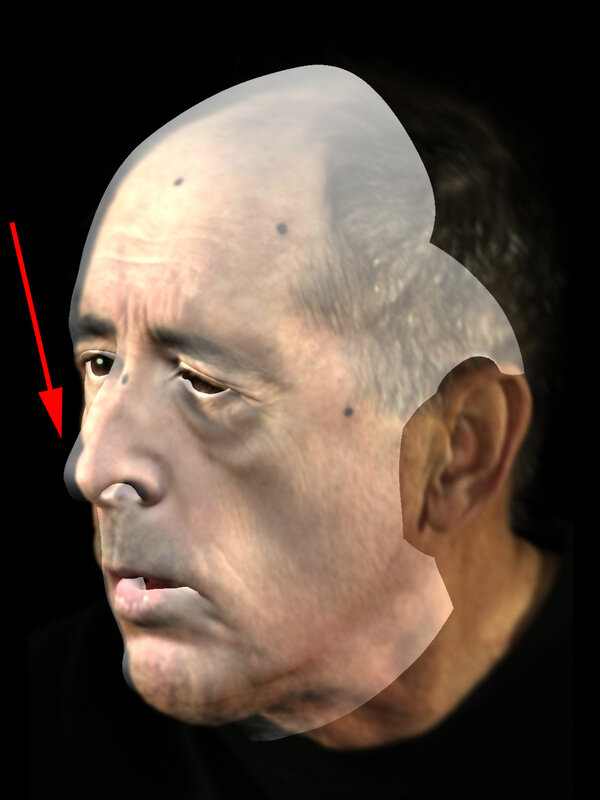}
    \includegraphics[width=\linewidth]{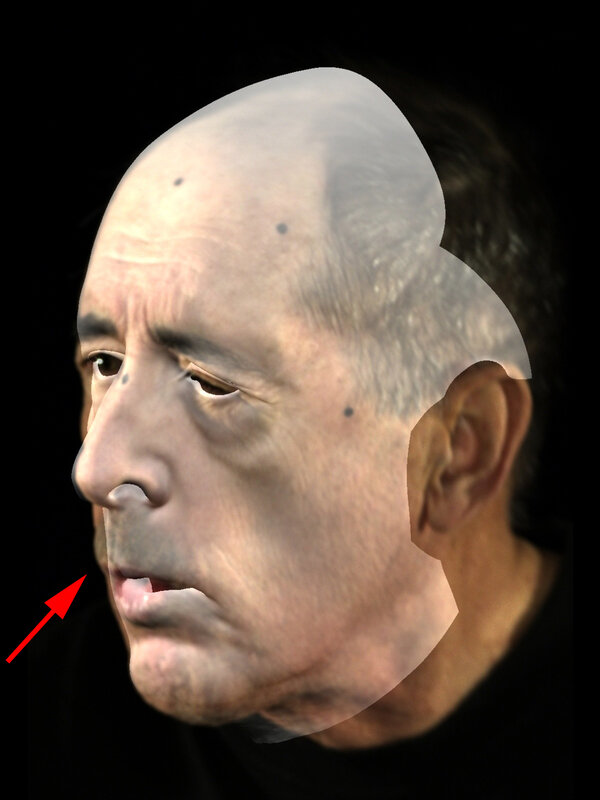}
    \includegraphics[width=\linewidth]{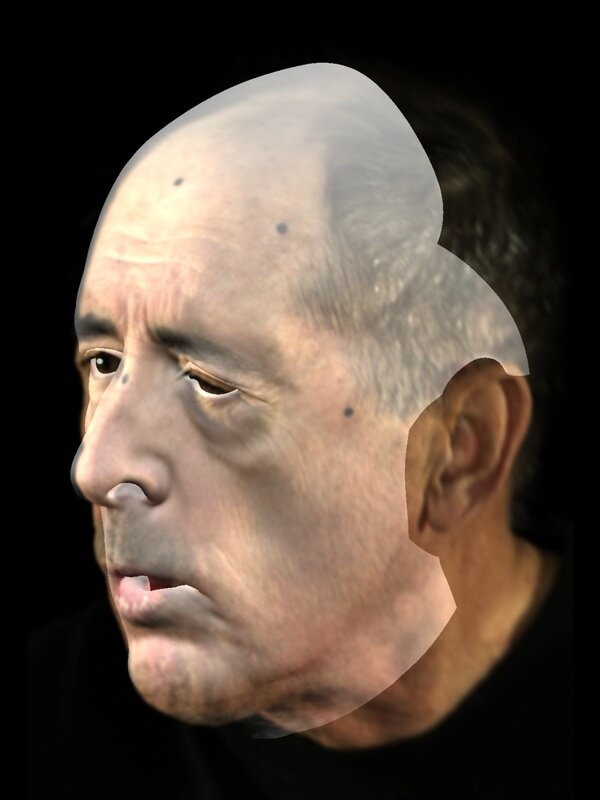}
    \includegraphics[width=\linewidth]{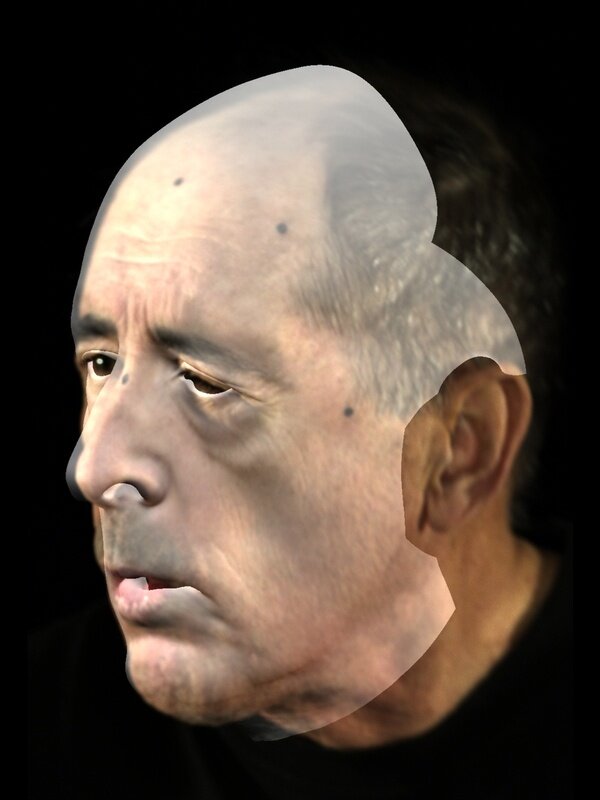}
    \caption*{1147}
\end{subfigure}
\begin{subfigure}[b]{0.15\linewidth}
    \includegraphics[width=\linewidth]{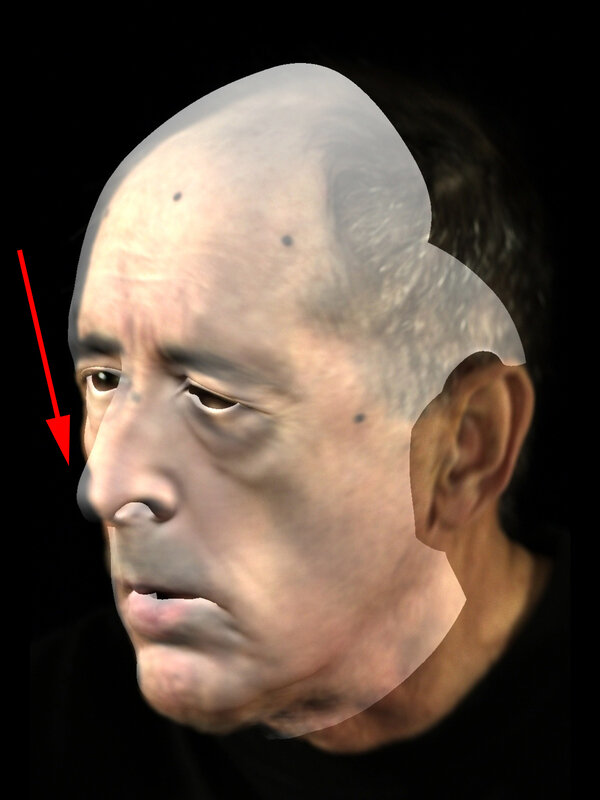}
    \includegraphics[width=\linewidth]{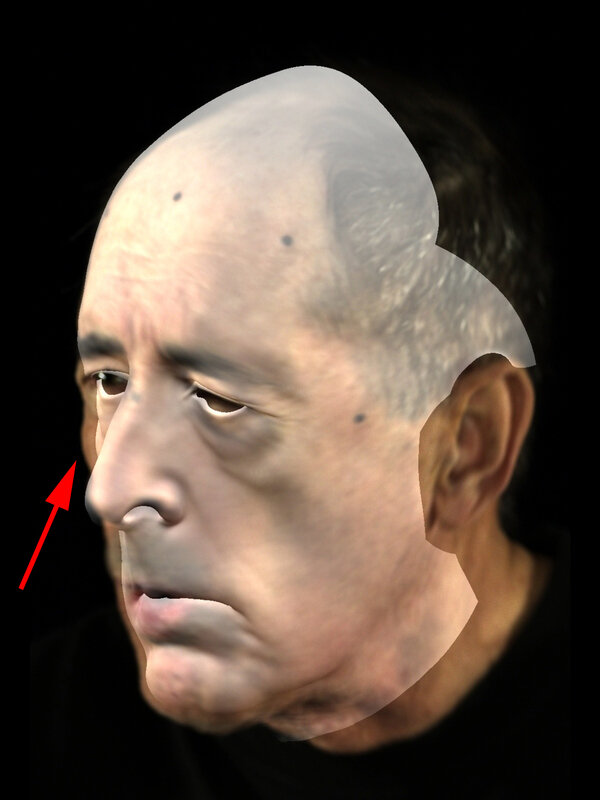}
    \includegraphics[width=\linewidth]{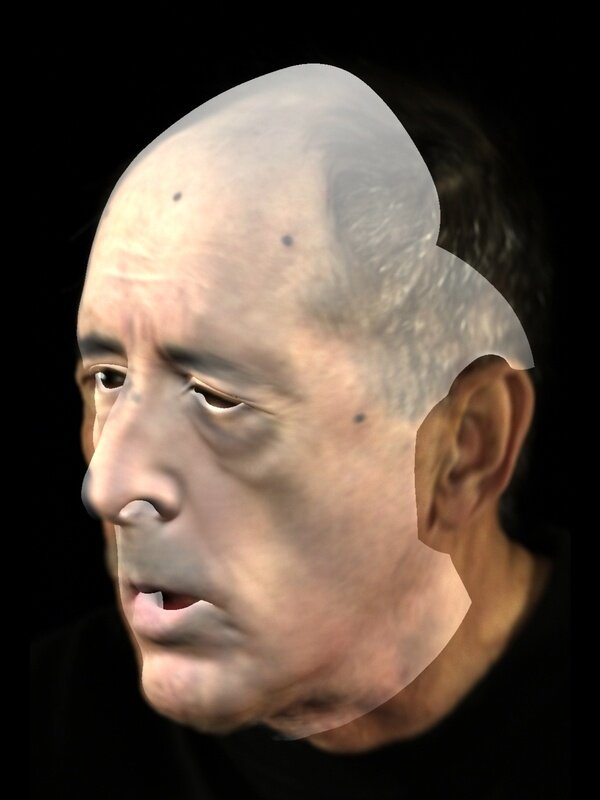}
    \includegraphics[width=\linewidth]{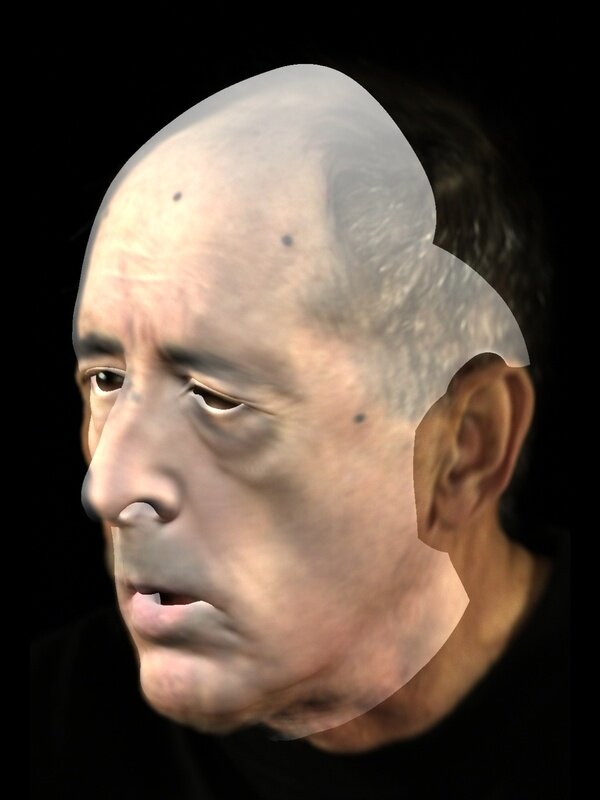}
    \caption*{1148}
\end{subfigure}
\begin{subfigure}[b]{0.15\linewidth}
    \includegraphics[width=\linewidth]{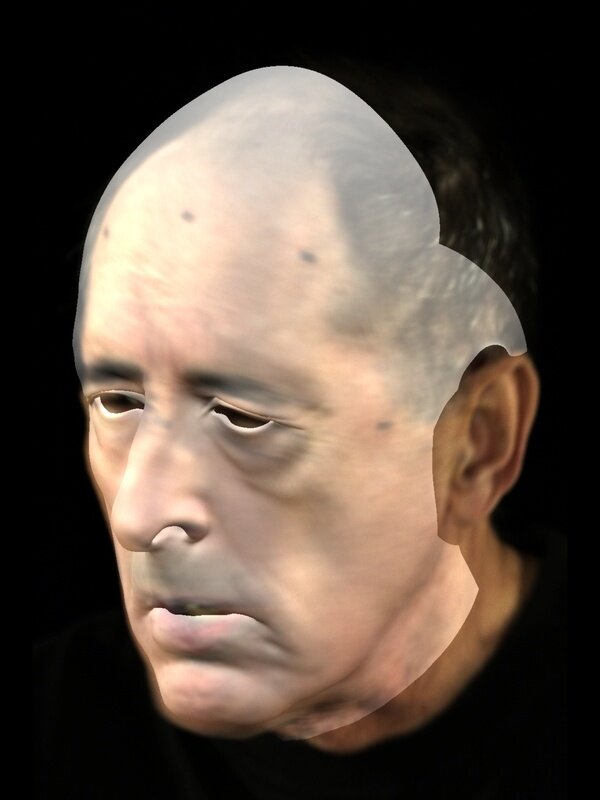}
    \includegraphics[width=\linewidth]{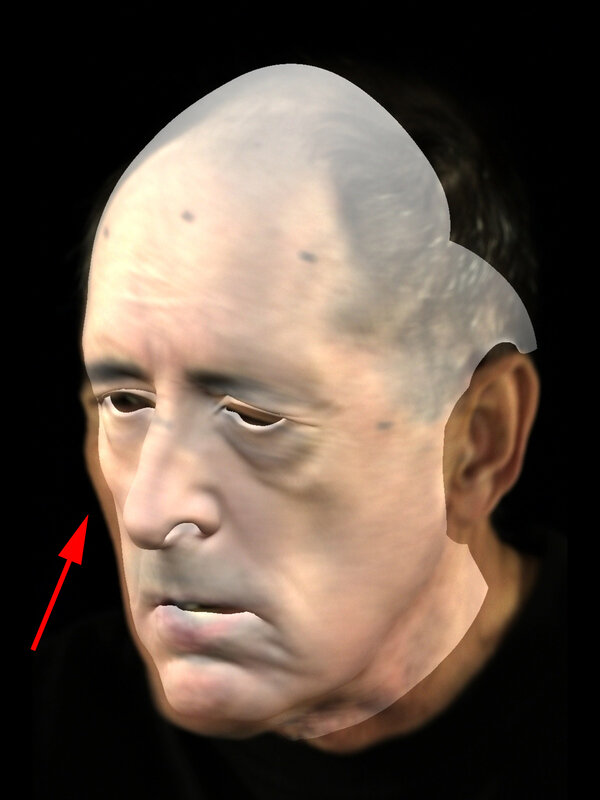}
    \includegraphics[width=\linewidth]{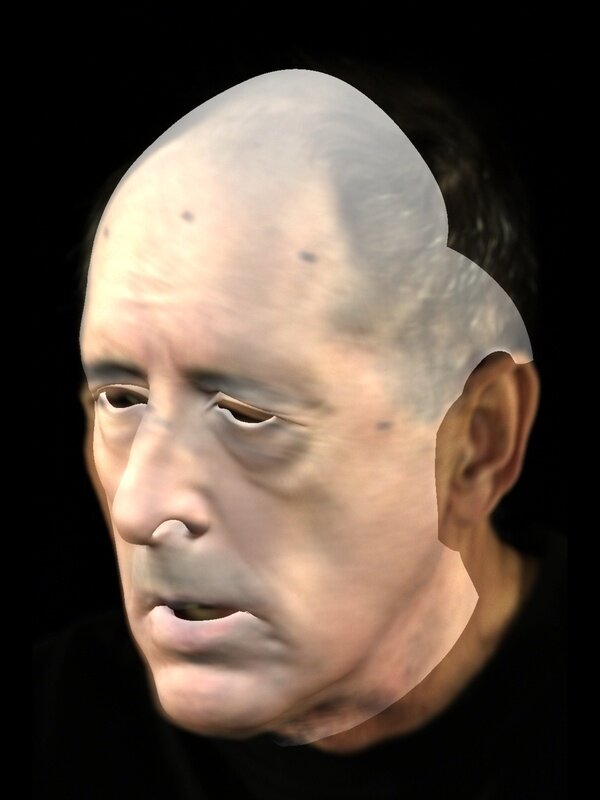}
    \includegraphics[width=\linewidth]{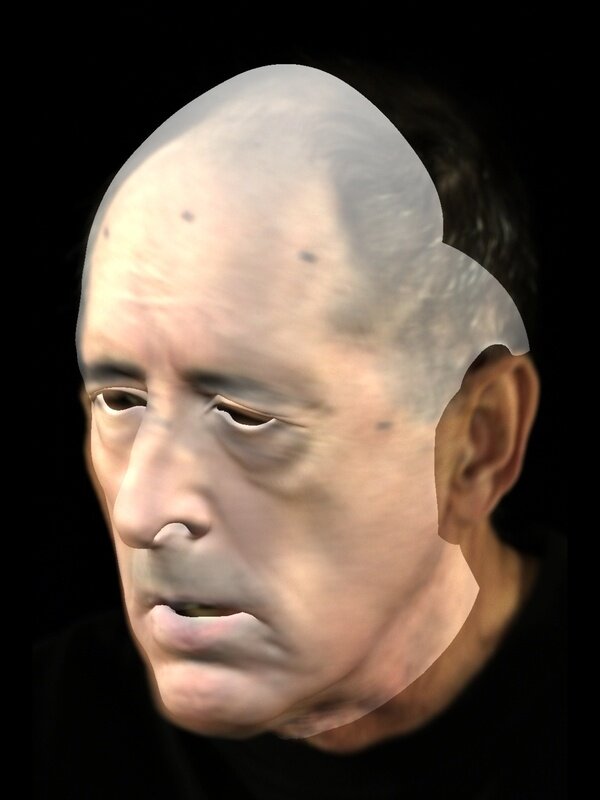}
    \caption*{1149}
\end{subfigure}
\begin{subfigure}[b]{0.15\linewidth}
    \includegraphics[width=\linewidth]{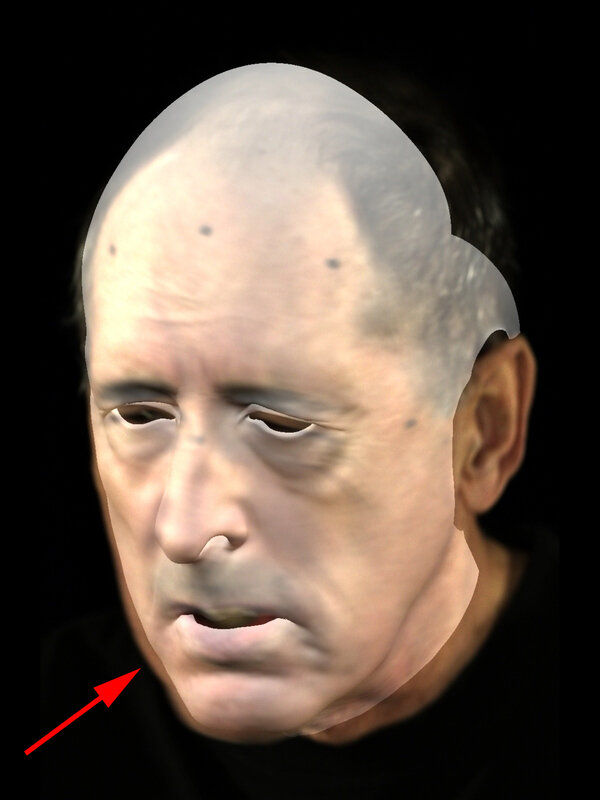}
    \includegraphics[width=\linewidth]{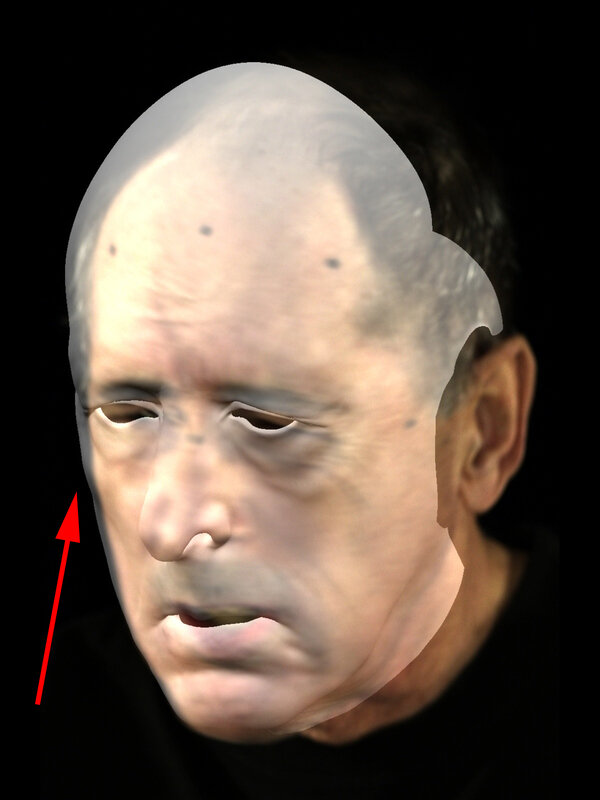}
    \includegraphics[width=\linewidth]{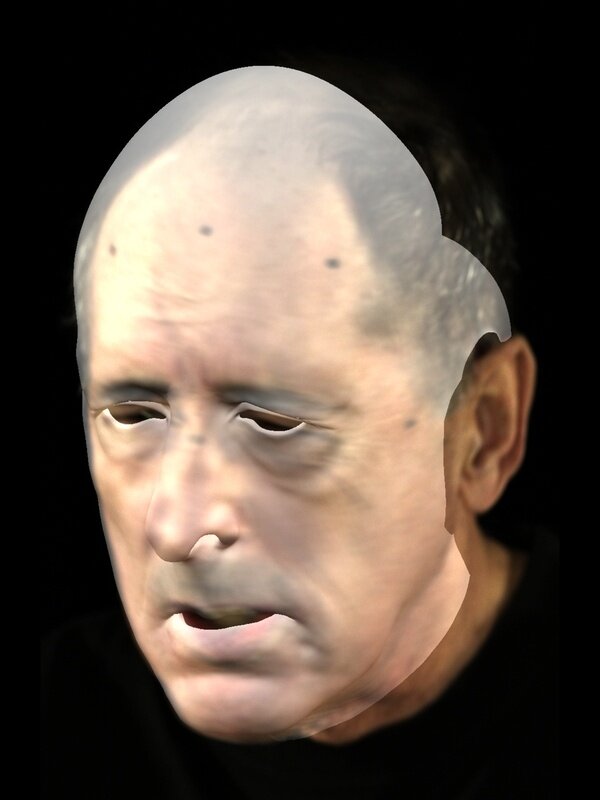}
    \includegraphics[width=\linewidth]{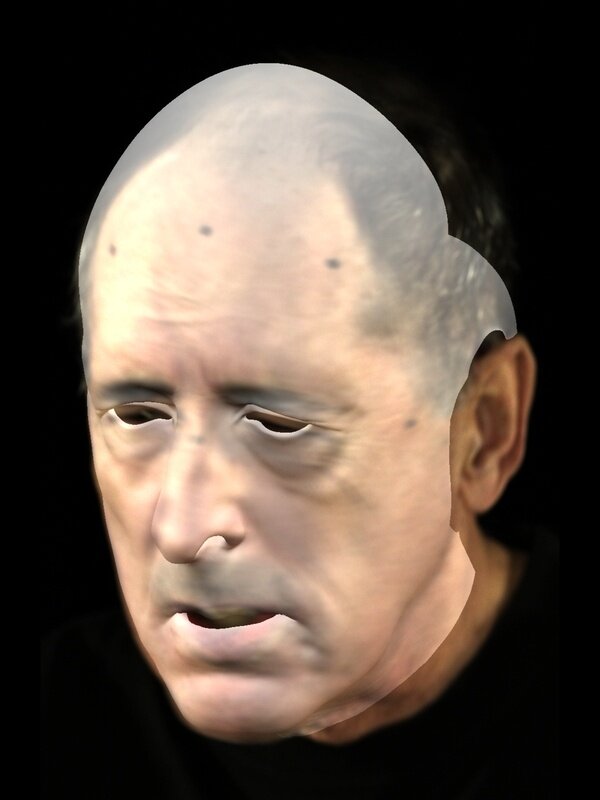}
    \caption*{1150}
\end{subfigure}
\begin{subfigure}[b]{0.15\linewidth}
    \includegraphics[width=\linewidth]{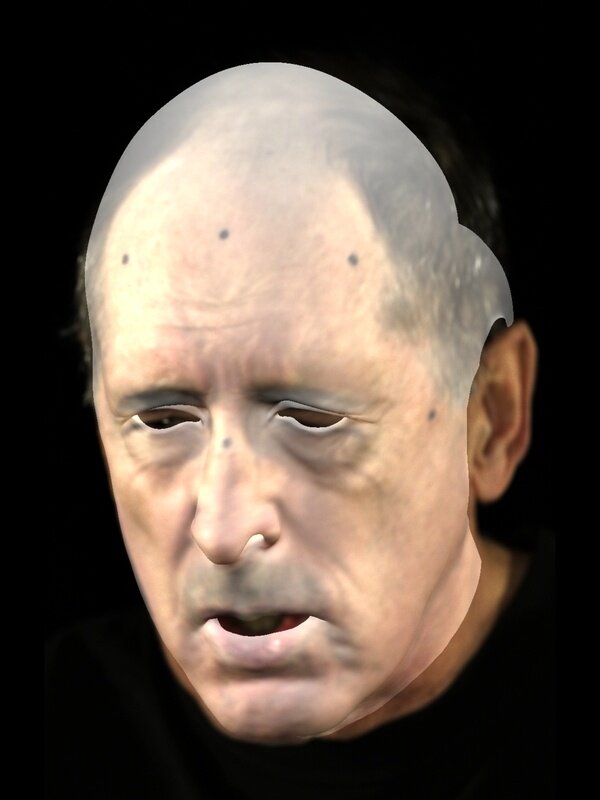}
    \includegraphics[width=\linewidth]{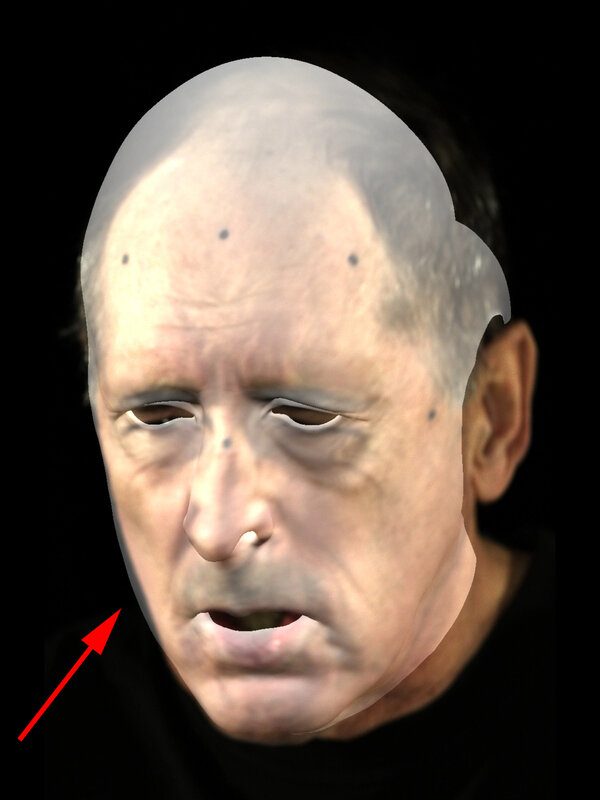}
    \includegraphics[width=\linewidth]{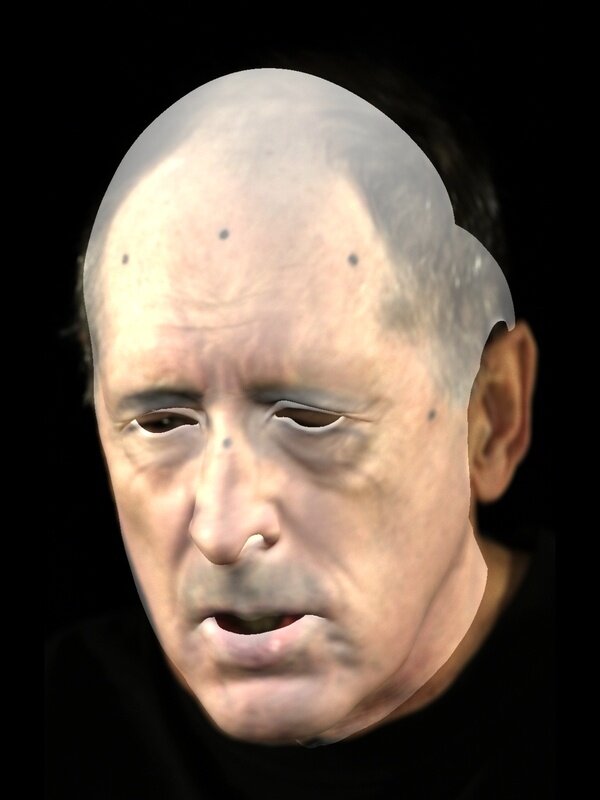}
    \includegraphics[width=\linewidth]{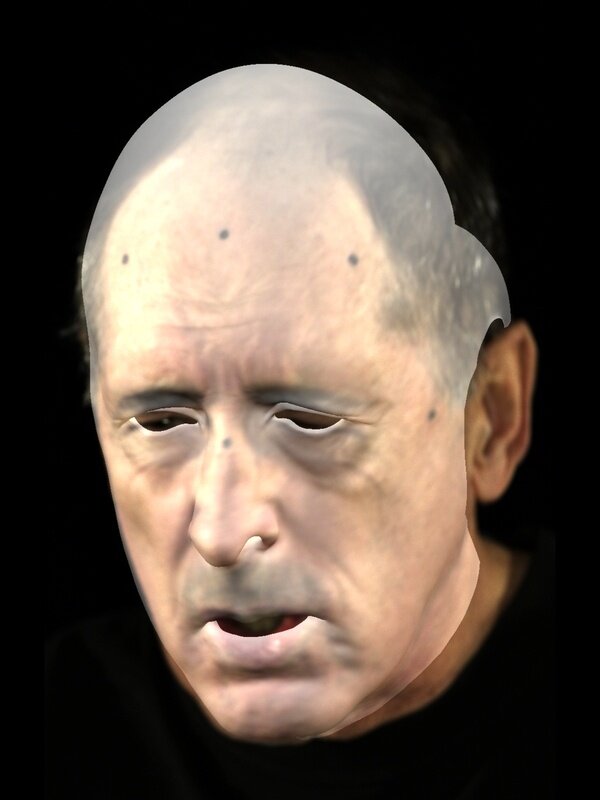}
    \caption*{1151}
\end{subfigure}
\hfill
\caption{A comparison of the geometry produced by the various proposed temporal smoothing methods overlayed on the captured images.
Notice how the techniques involving optical flow information from the captured images better track the nose and the contours of the face.}
\label{fig:smoothing_approaches}
\end{figure*}

Figure \ref{fig:smoothing_approaches} (third row) shows the results obtained by matching the synthetic render's optical flow to the captured image's optical flow (denoted plate flow in Figure \ref{fig:smoothing_approaches}).
Although this generally produces accurate results when looking at each frame in isolation, adjacent frames may still obtain visually disjoint results (see the accompanying video).
Thus, we explore additional temporal smoothing methods.

We first explore temporally smoothing the parameters ($\theta$, $t$, and $w$) by computing a weighted average over a three frame window centered at every frame.
We weigh the current frame more heavily and use the method of \cite{markley2007quaternion} to average the rigid rotation parameters.
While this approach produces temporally smooth parameters, it generally causes the synthetic render to no longer match the captured image.
This inaccuracy is demonstrated in Figure \ref{fig:smoothing_approaches} (top row, denoted as averaging) and is especially apparent around the nose (frames \num{1147} and \num{1148}) and around the lower right cheek (frame \num{1150}).

One could also carry out averaging using an optical flow network.
This can be accomplished by finding the parameters $p_2$ that minimize the difference in optical flow fields between the current frame's synthetic render and the adjacent frames' synthetic renders, \ie $\| N(F_1, F_2) - N(F_2, F_3) \|_2$.
See Figure \ref{fig:smoothing_approaches} (second row, designated self flow).
This aims to minimize the second derivative of the motion of the head in the image plane; however, in practice, we found this method to have little effect on temporal noise while still causing the synthetic render to deviate from the captured image.
These inaccuracies are most noticeable around the right cheek and lips.

\begin{figure*}[p]
\centering
\begin{multicols}{2}
\begin{subfigure}{\linewidth}
  \includegraphics[width=\linewidth]{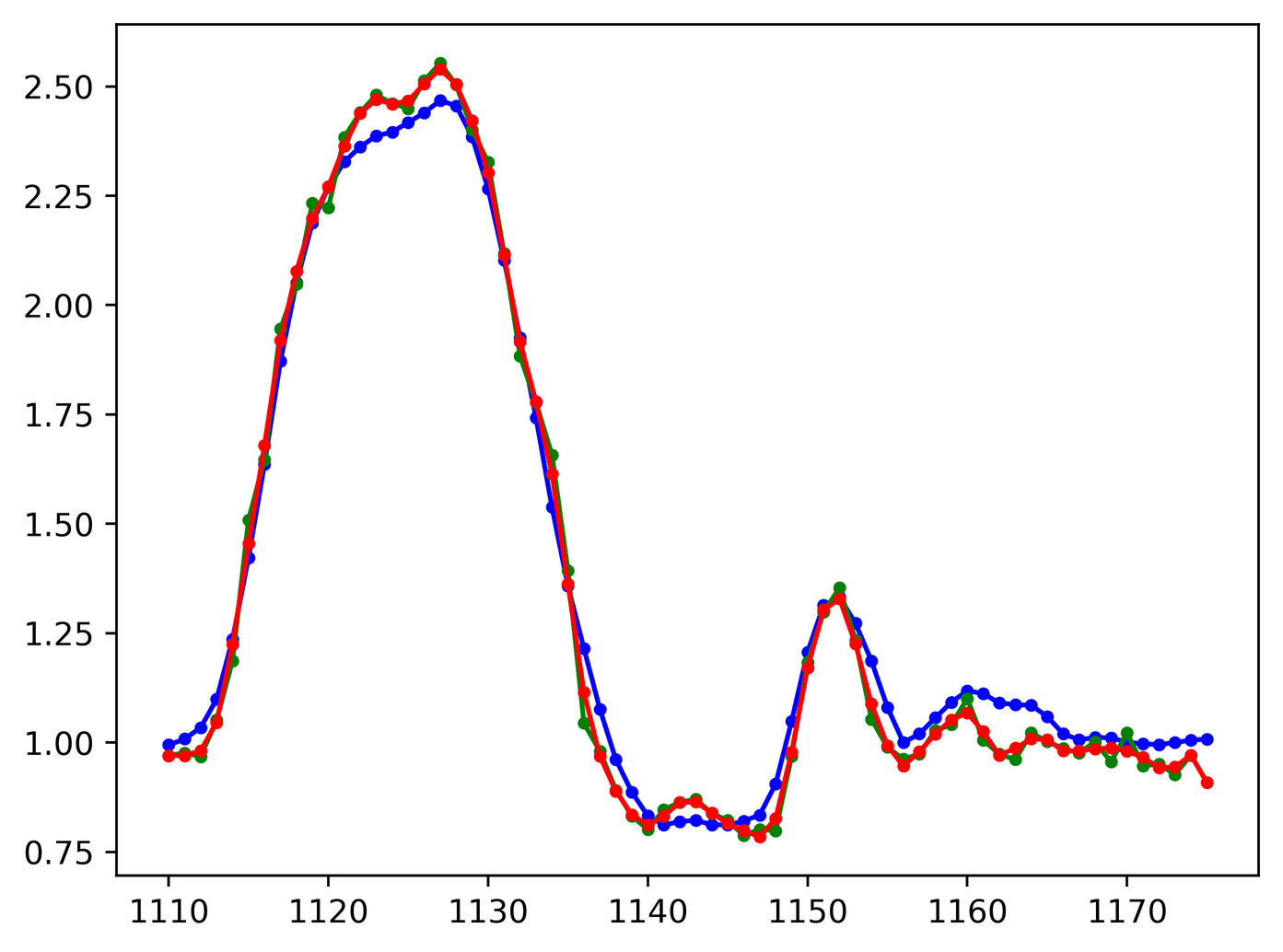}
  \caption{Euler Angle X}
\end{subfigure}
\begin{subfigure}{\linewidth}
  \includegraphics[width=\linewidth]{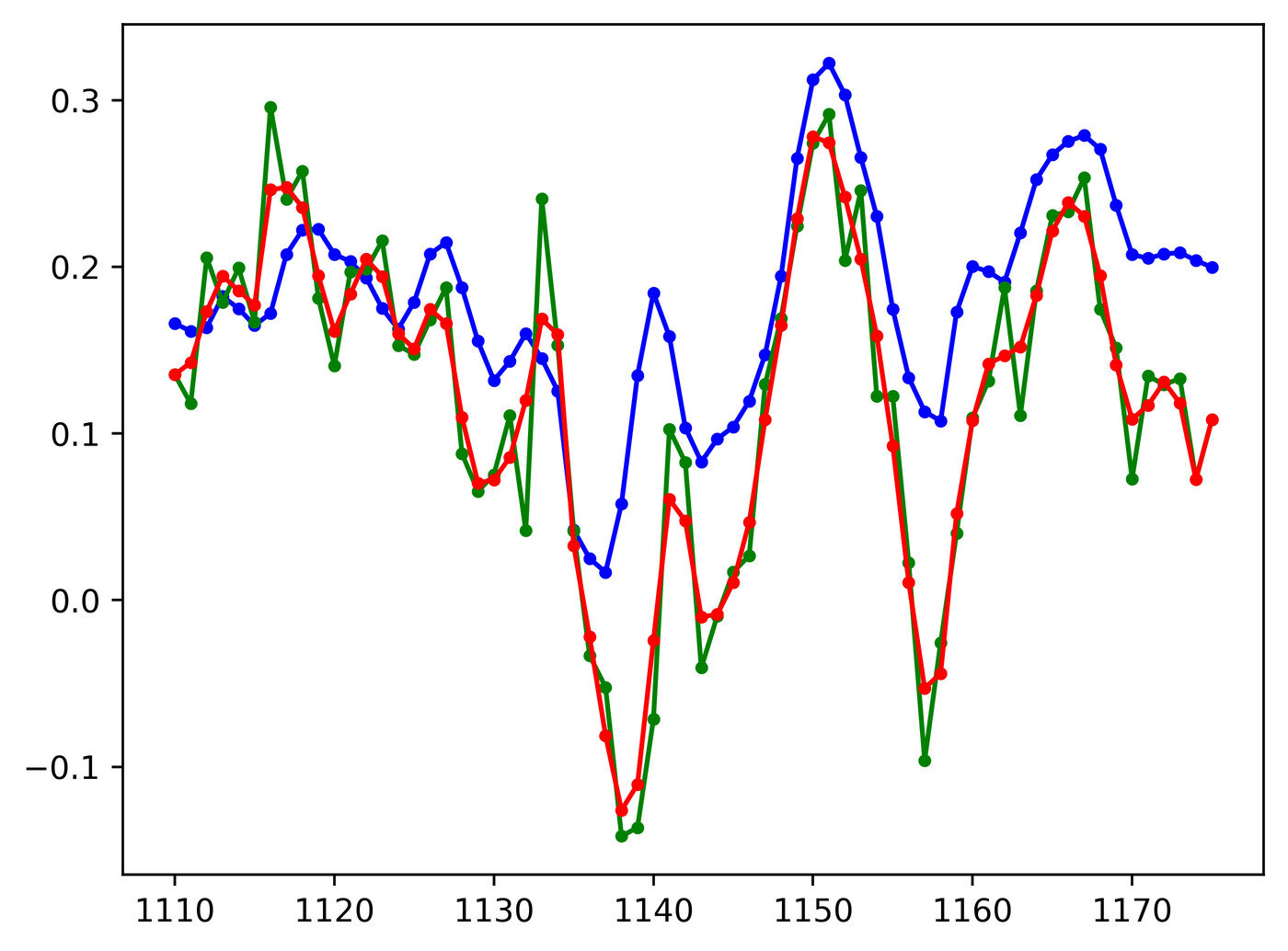}
  \caption{Euler Angle Y}
\end{subfigure}
\begin{subfigure}{\linewidth}
  \includegraphics[width=\linewidth]{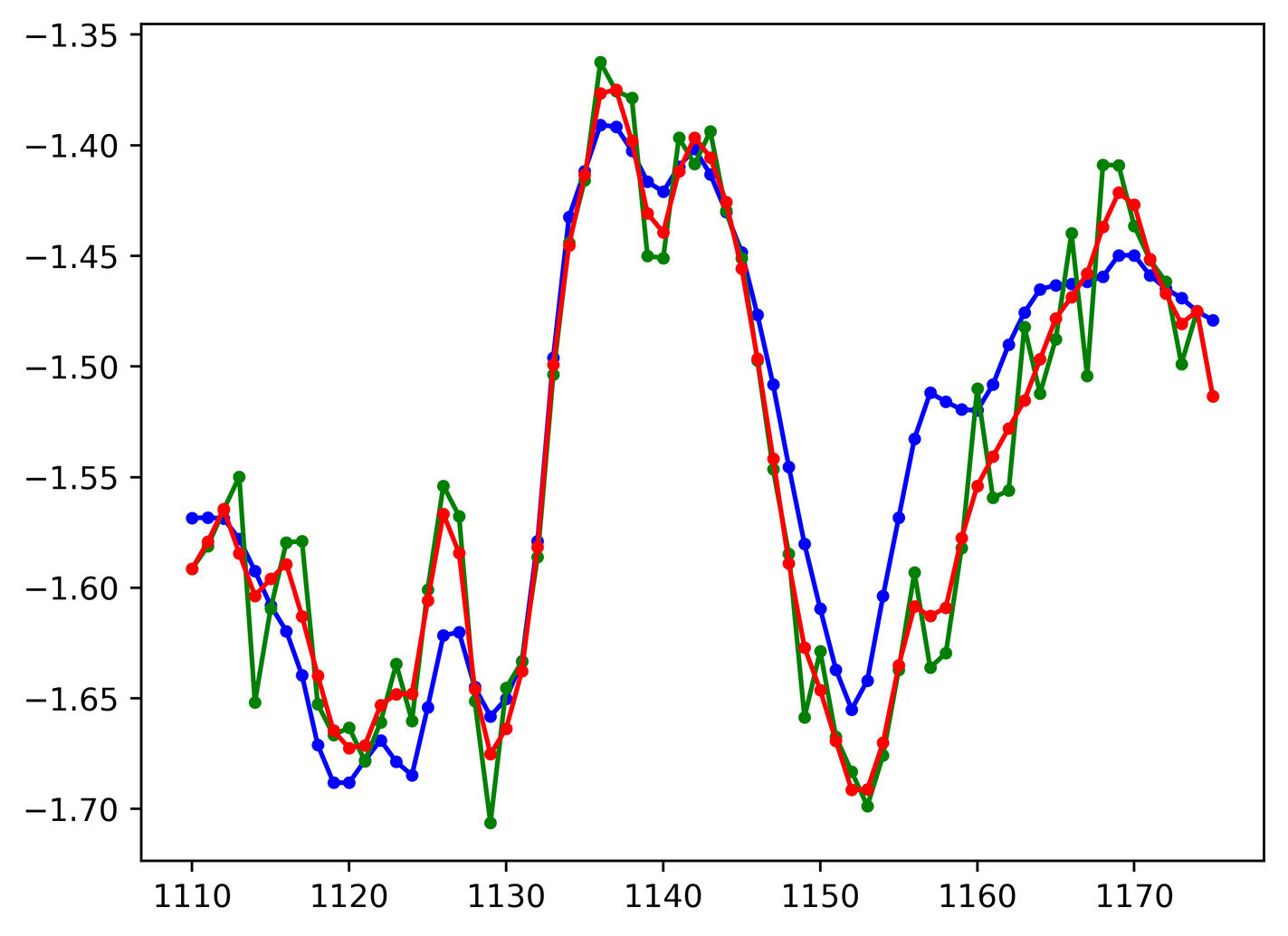}
  \caption{Euler Angle Z}
\end{subfigure}
\begin{subfigure}{\linewidth}
  \includegraphics[width=\linewidth]{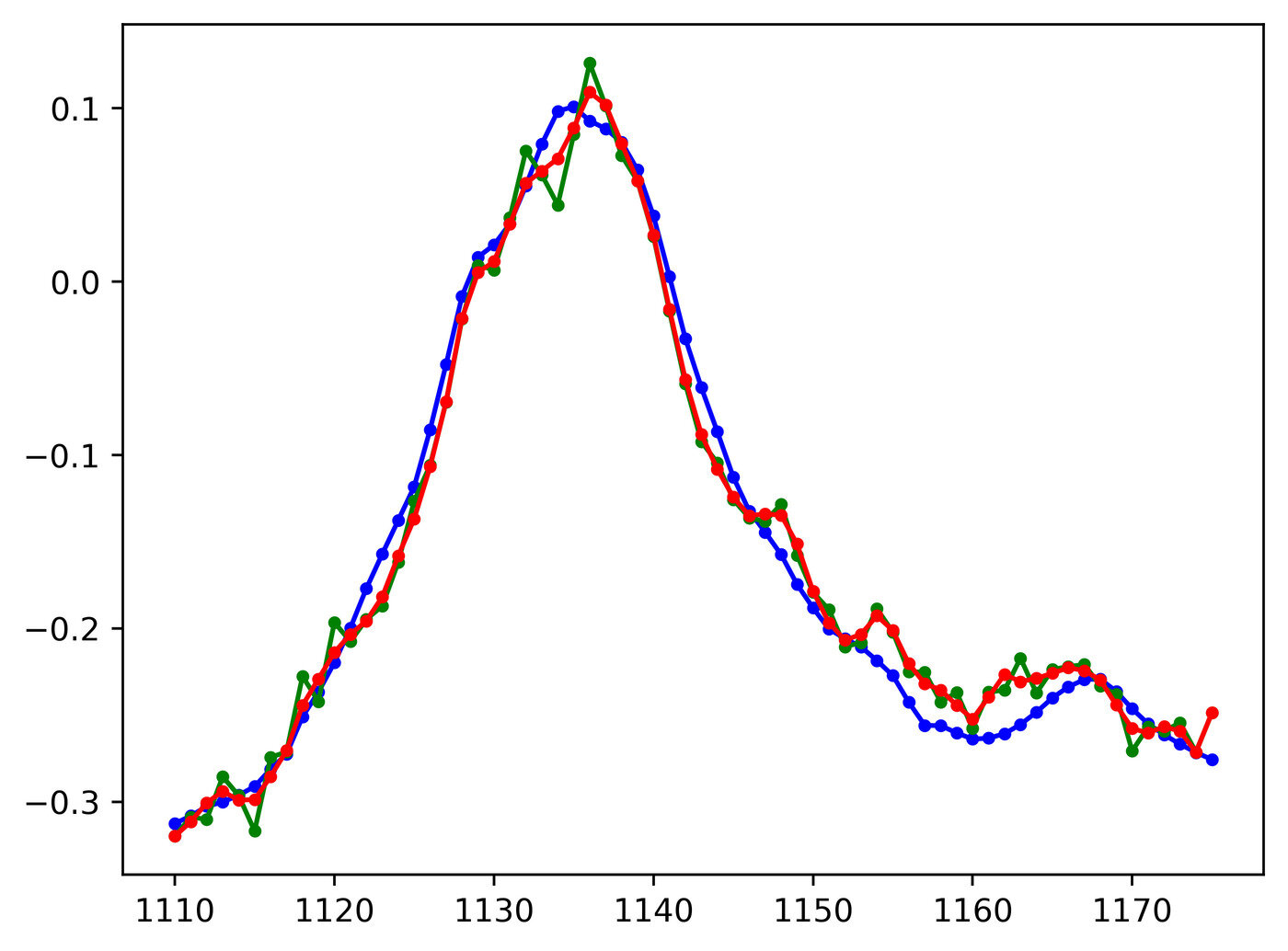}
  \caption{Translation X}
\end{subfigure}
\begin{subfigure}{\linewidth}
  \includegraphics[width=\linewidth]{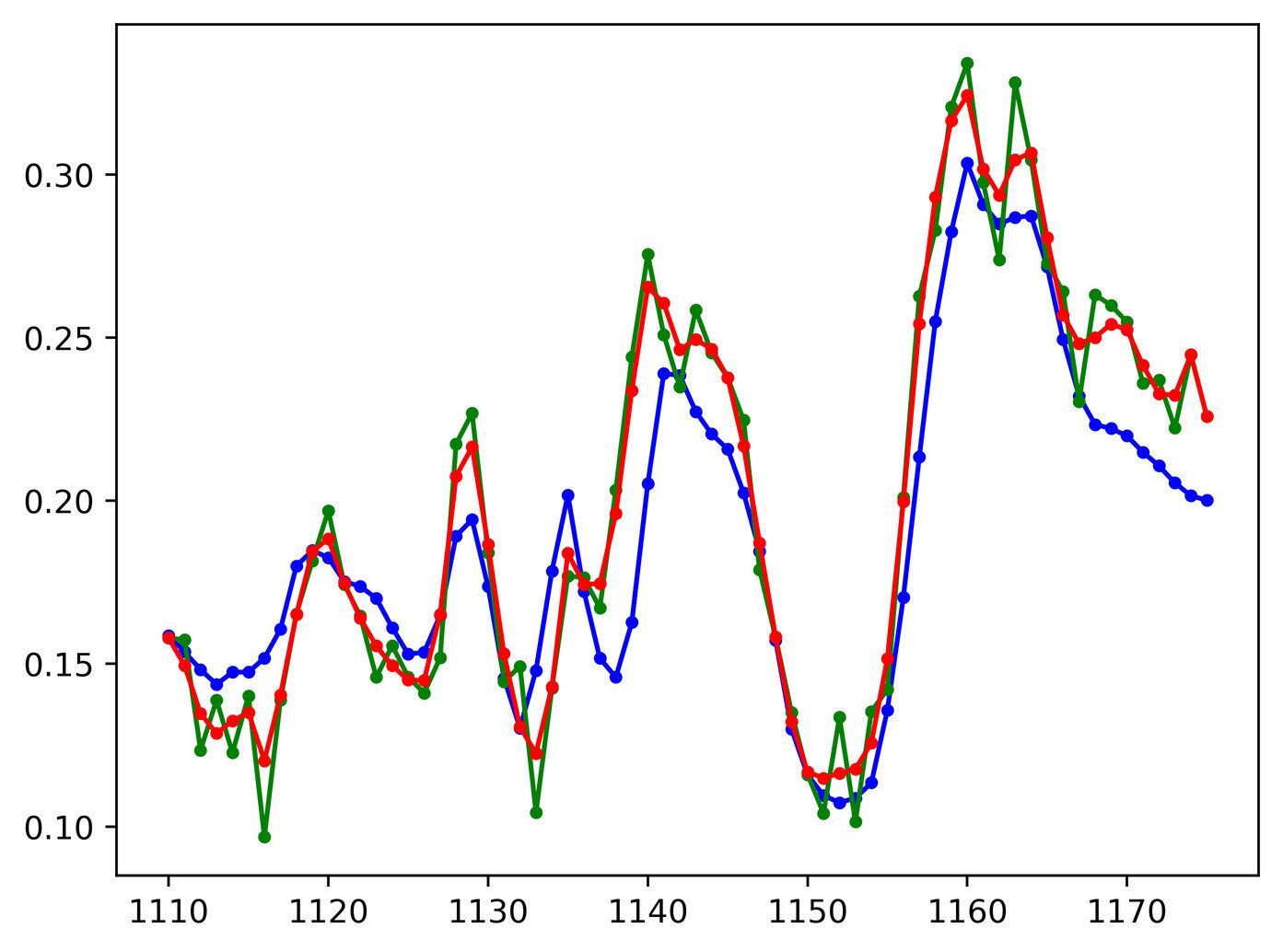}
  \caption{Translation Y}
\end{subfigure}
\begin{subfigure}{\linewidth}	
  \includegraphics[width=\linewidth]{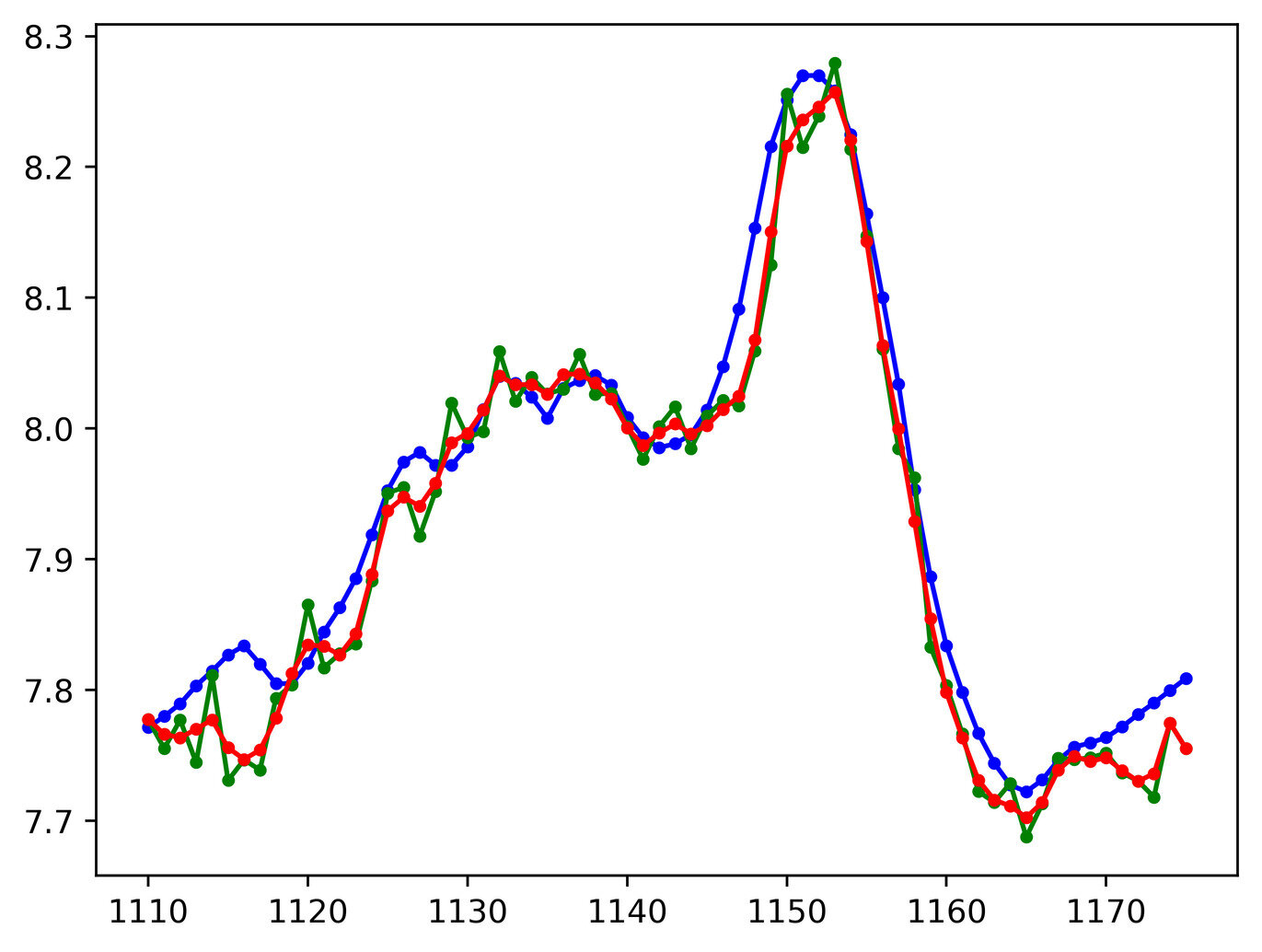}
  \caption{Translation Z}
\end{subfigure}
\end{multicols}
\caption{A comparison of the rigid parameters $\theta$ and $t$ from a manual process (blue), from the initial expression estimation (green), and from the hybrid smoothing approach (red).}
\label{fig:smoothing_rigid_plots}
\end{figure*}

\begin{figure*}[p]
\centering
\includegraphics[width=0.8\linewidth]{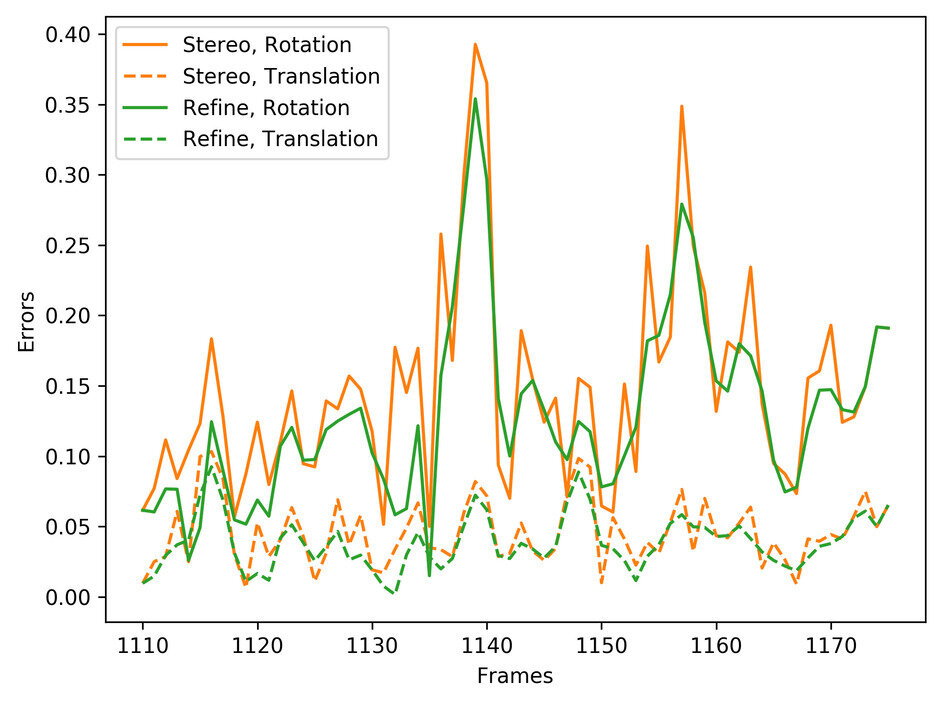}
\caption{Assuming the manual rigid alignment is the ``ground truth,'' we compare the errors for rigid parameters for the stereo results (orange) and the temporally smoothed stereo results (green).
Note that the temporal smoothing does not increase the errors already found in the stereo rigid alignment.}
\label{fig:appendix_temporal_refinement_errors}
\vspace{-5.0mm}
\end{figure*}

\begin{figure*}[p]
\centering
\begin{multicols}{2}
\begin{subfigure}{\linewidth}
  \includegraphics[width=\linewidth]{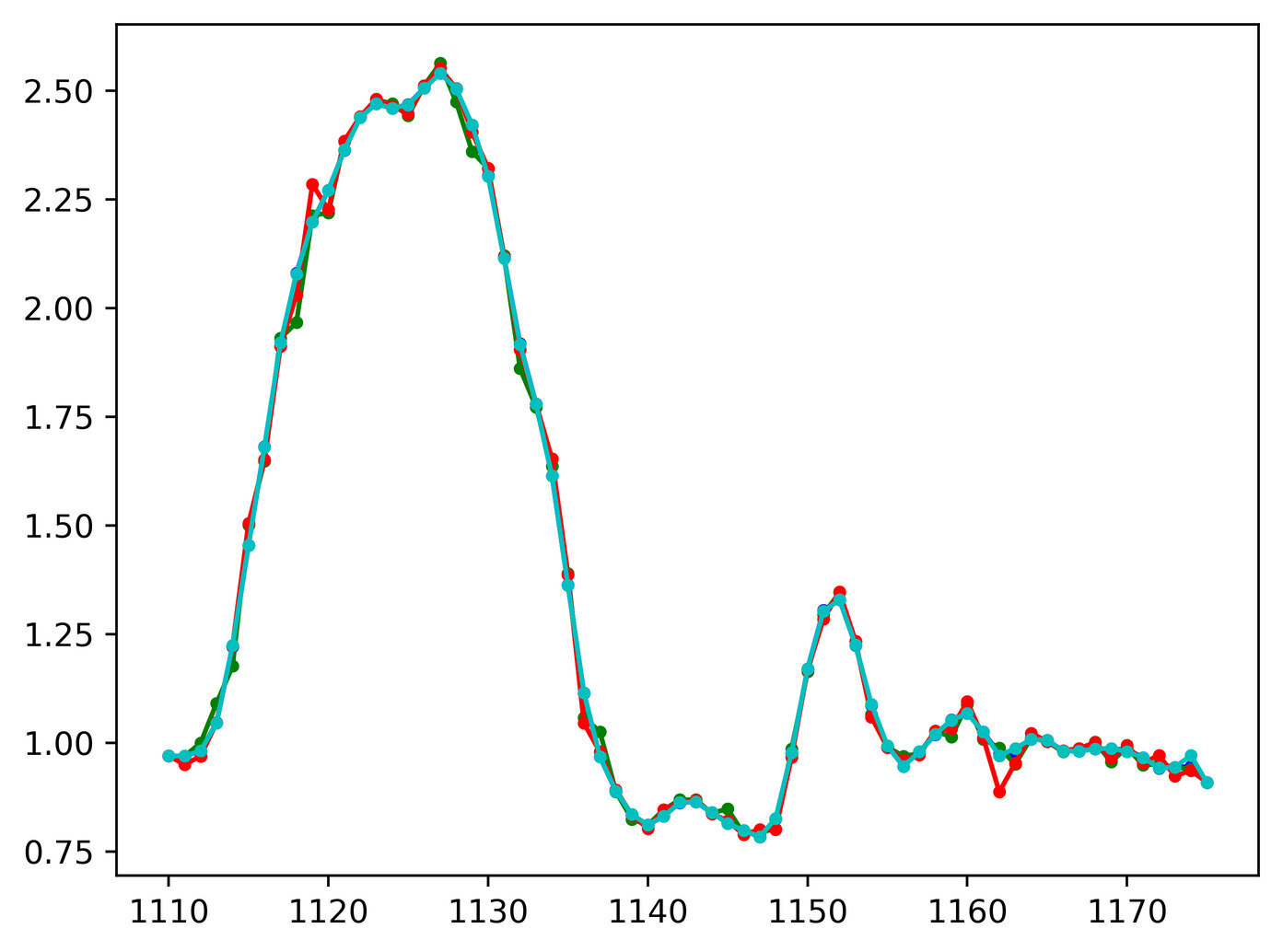}
  \caption{Euler Angle X}
\end{subfigure}
\begin{subfigure}{\linewidth}
  \includegraphics[width=\linewidth]{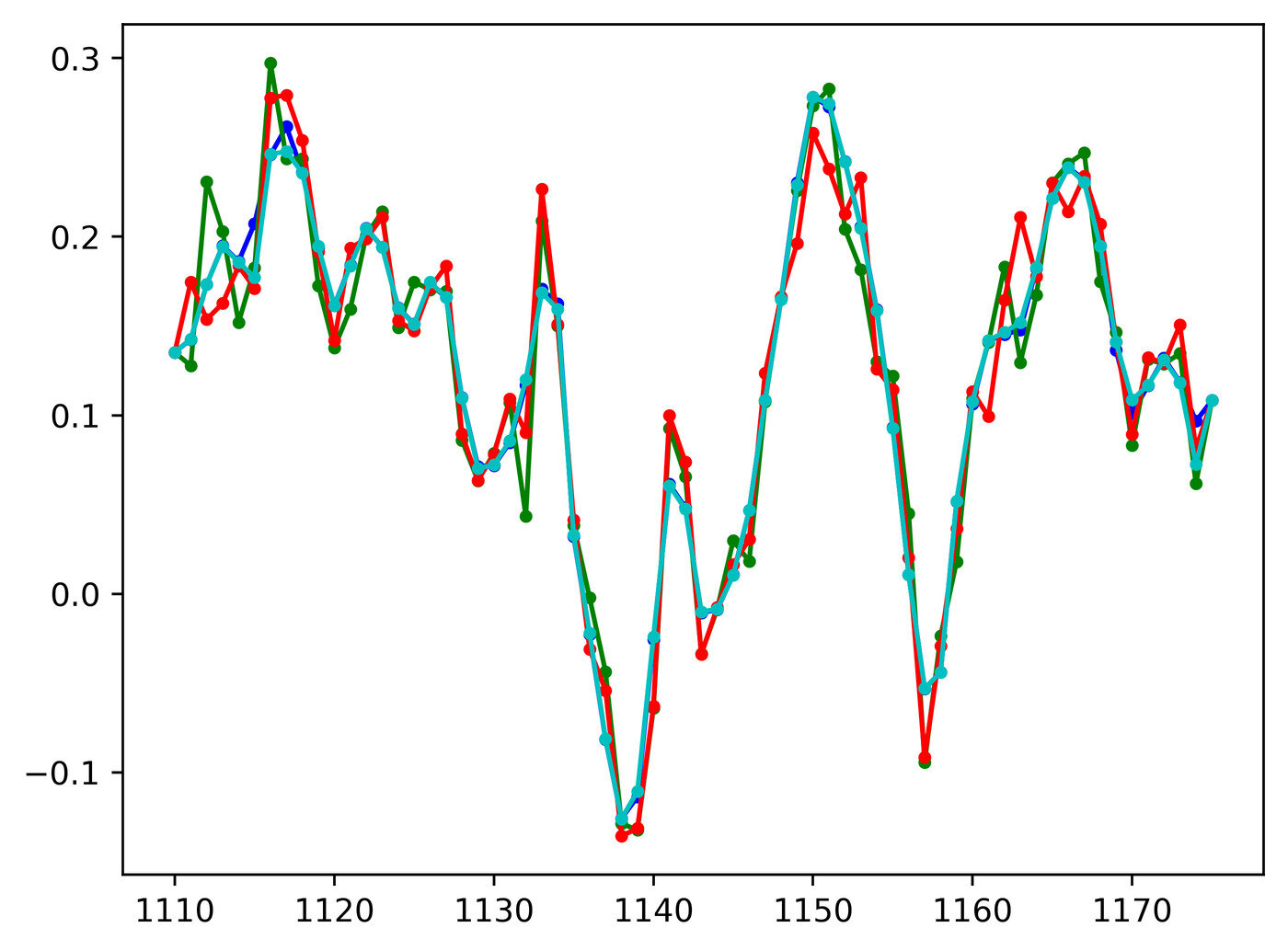}
  \caption{Euler Angle Y}
\end{subfigure}
\begin{subfigure}{\linewidth}
  \includegraphics[width=\linewidth]{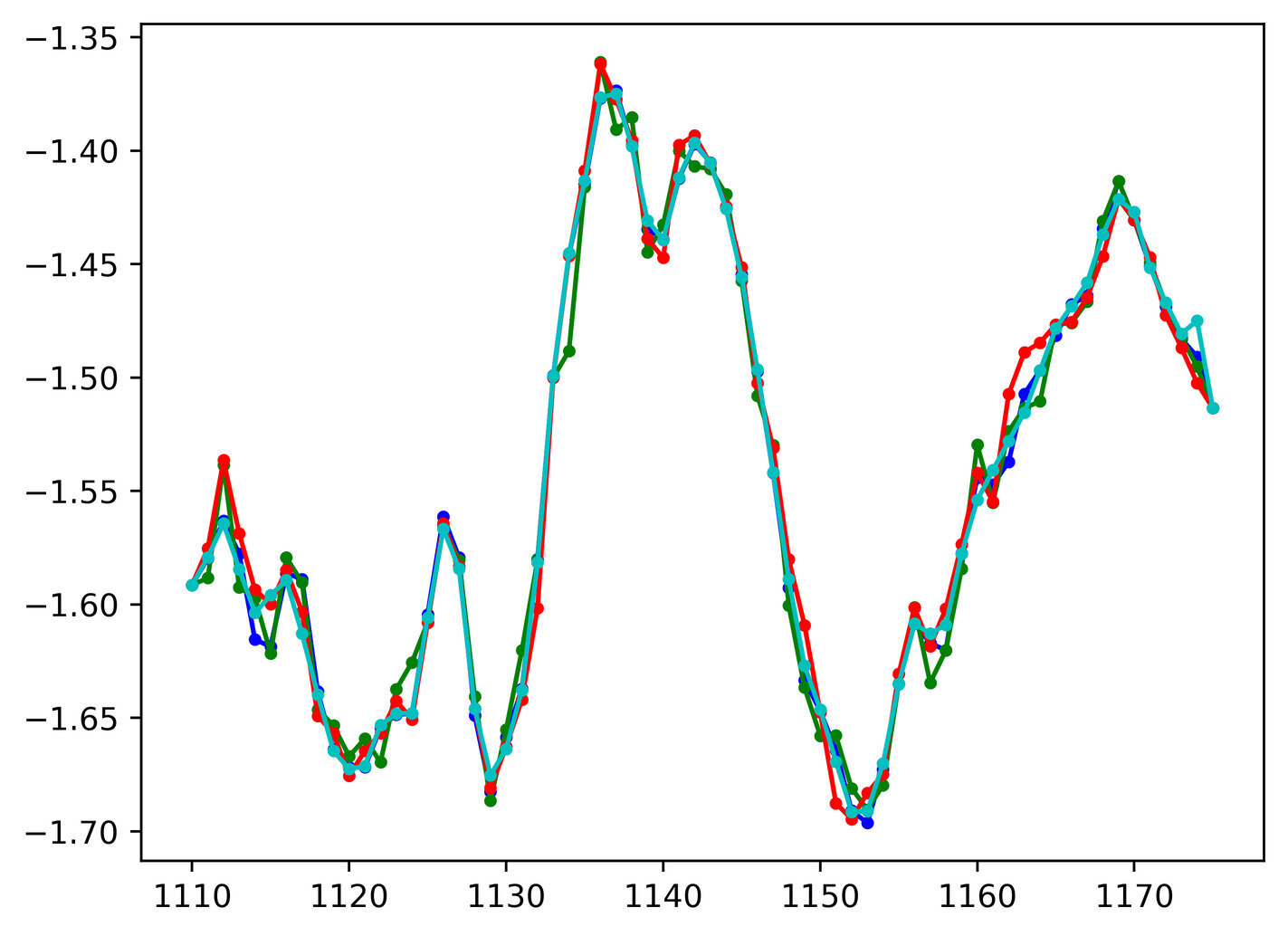}
  \caption{Euler Angle Z}
\end{subfigure}
\begin{subfigure}{\linewidth}
  \includegraphics[width=\linewidth]{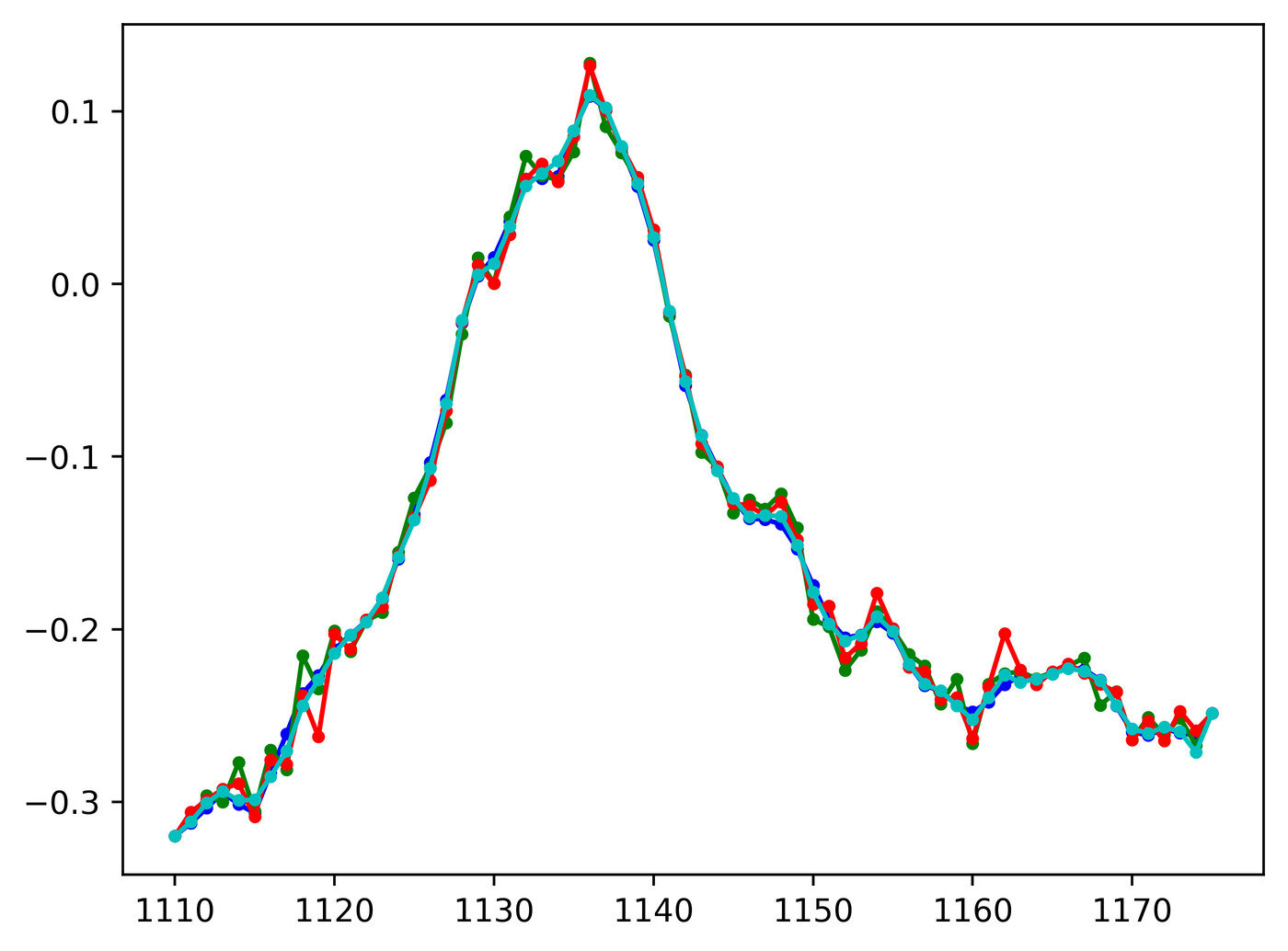}
  \caption{Translation X}
\end{subfigure}
\begin{subfigure}{\linewidth}
  \includegraphics[width=\linewidth]{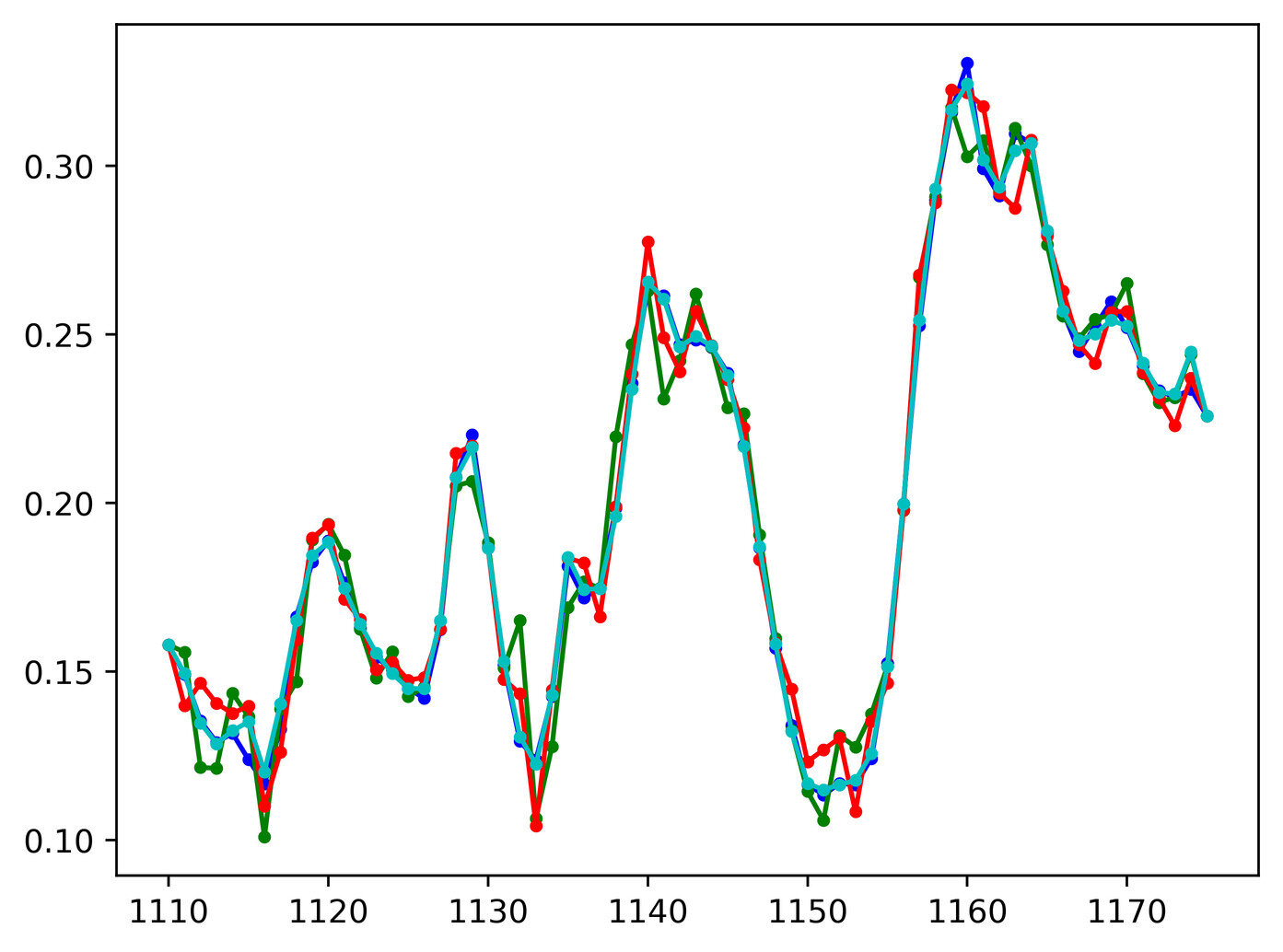}
  \caption{Translation Y}
\end{subfigure}
\begin{subfigure}{\linewidth}	
  \includegraphics[width=\linewidth]{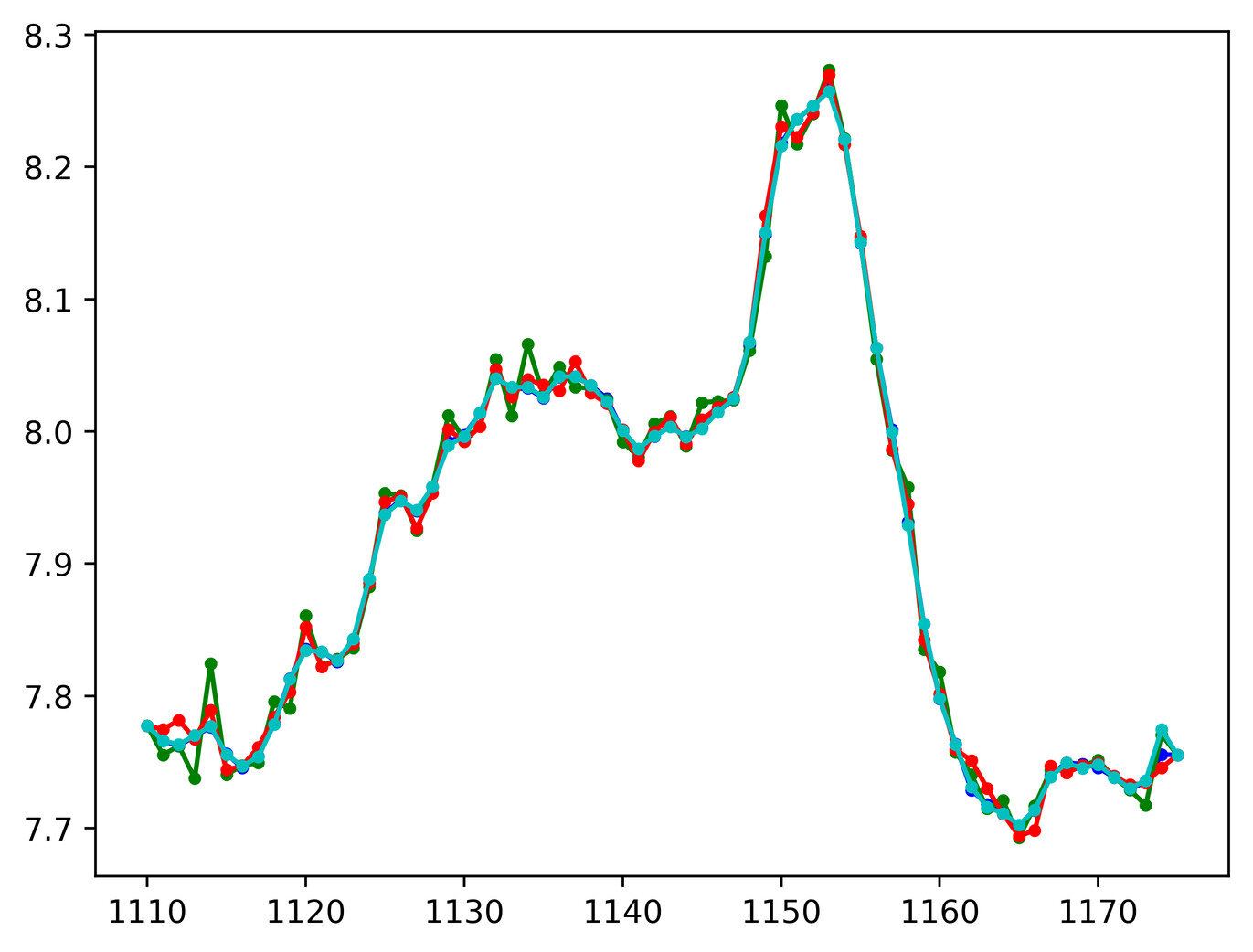}
  \caption{Translation Z}
\end{subfigure}
\end{multicols}
\caption{A comparison of the rigid parameters obtained by averaging (blue), self flow (green), plate flow (red), and the hybrid smoothing approach (cyan).}
\label{fig:smoothing_approaches_rigid_plots}
\end{figure*}

We found the most effective approach to temporal refinment to be a two step process: First, we use averaging to produce temporally consistent parameter values.
Then, starting from those values, we use the optical flow approach to make the synthetic render flow better target that of the plate.
See Figure \ref{fig:smoothing_approaches} (bottom row, denoted hybrid).
This hybrid approach produces temporally consistent results with synthetic renders that still match the captured image.
Figure \ref{fig:smoothing_rigid_plots} shows the rigid parameters before and after using this hybrid approach, along with that obtained manually by a matchmove artist for reference.
Assuming the manual rigid alignment is the ``ground truth,'' Figure \ref{fig:appendix_temporal_refinement_errors} compares how far the rigid parameters are from their manually solved for values both before and after the hybrid smoothing approach.
Figure \ref{fig:smoothing_approaches_rigid_plots} compares all the proposed smoothing methods on this same example.

\subsection{Expression Reestimation}

\begin{figure*}[p]
\centering
\begin{subfigure}[b]{\dimexpr0.20\linewidth+20pt\relax}
    \makebox[20pt]{\raisebox{60pt}{\rotatebox[origin=c]{90}{1125}}}%
    \includegraphics[width=\dimexpr\linewidth-20pt\relax]{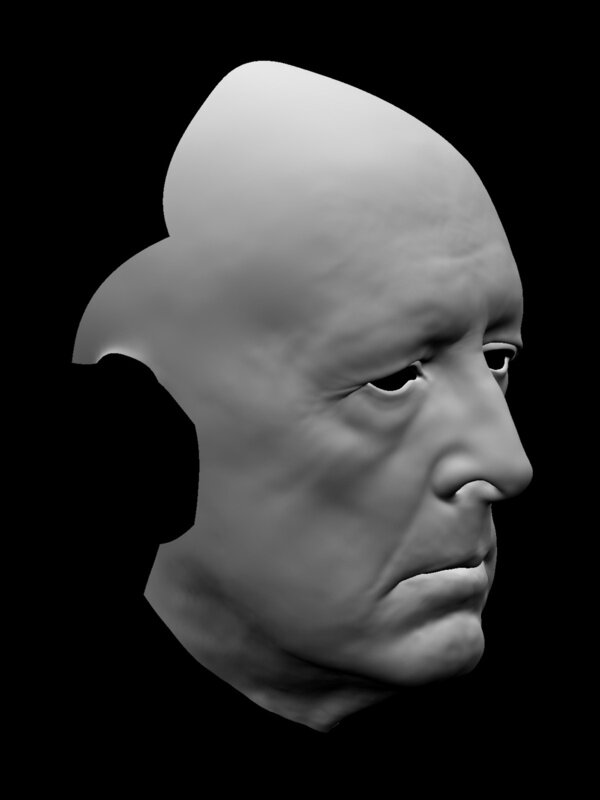}
    \makebox[20pt]{\raisebox{60pt}{\rotatebox[origin=c]{90}{1143}}}%
    \includegraphics[width=\dimexpr\linewidth-20pt\relax]{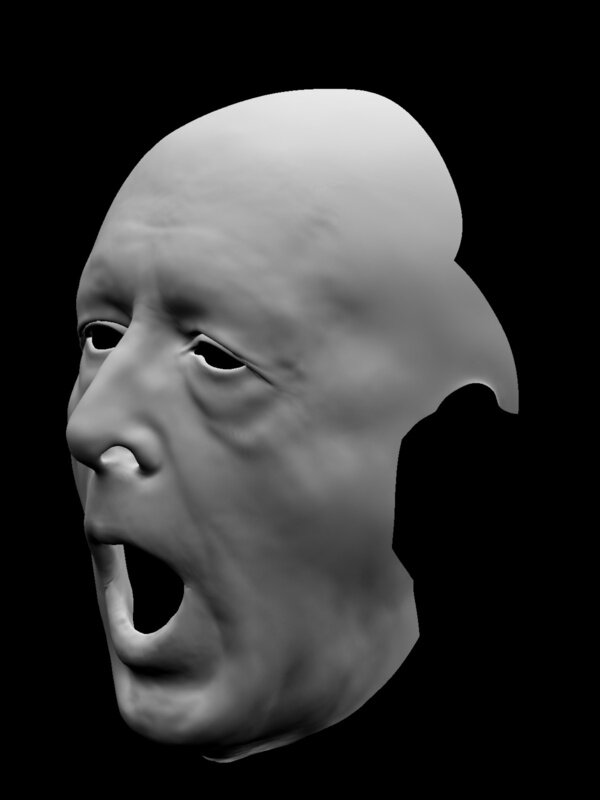}
    \makebox[20pt]{\raisebox{60pt}{\rotatebox[origin=c]{90}{1159}}}%
    \includegraphics[width=\dimexpr\linewidth-20pt\relax]{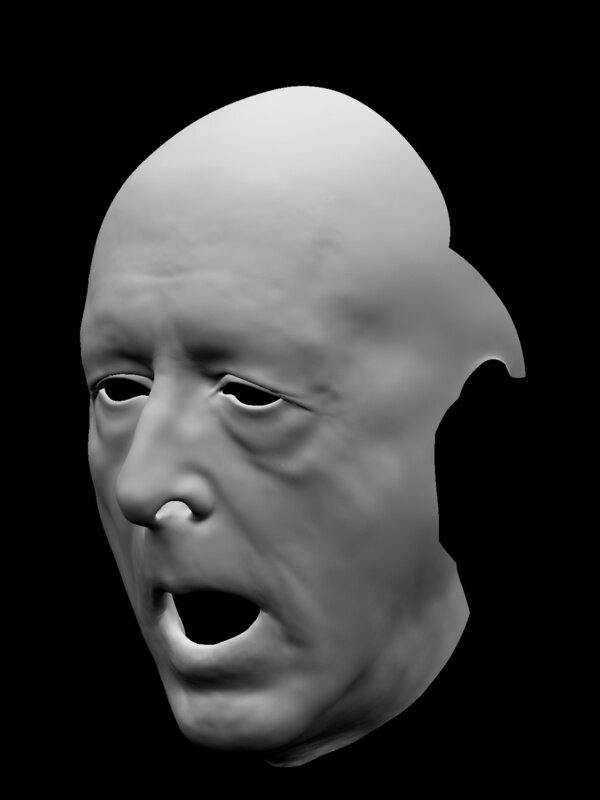}
    \caption{Stereo}
\end{subfigure}
\begin{subfigure}[b]{0.20\linewidth}
    \includegraphics[width=\linewidth]{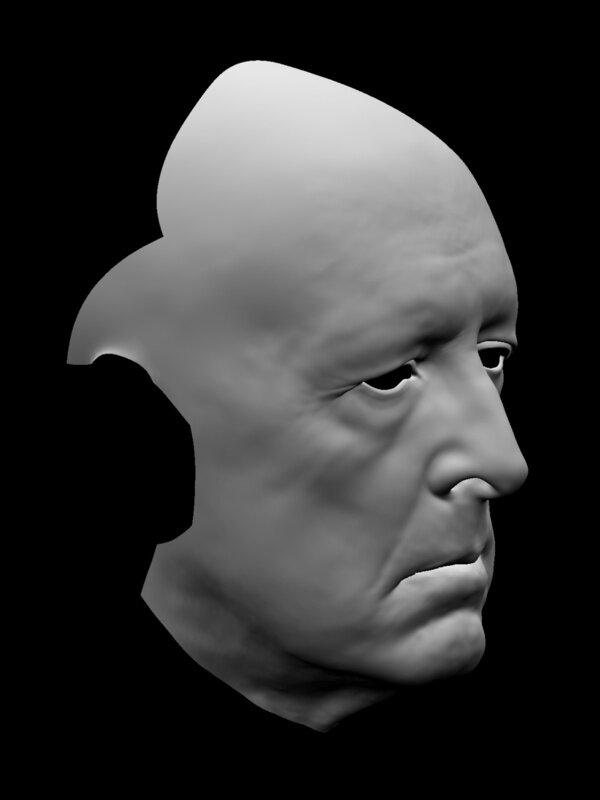}
    \includegraphics[width=\linewidth]{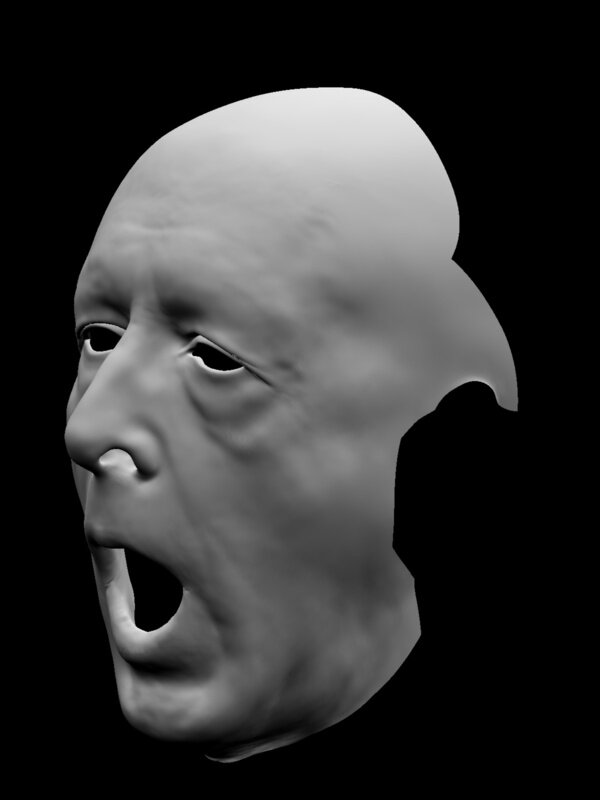}
    \includegraphics[width=\linewidth]{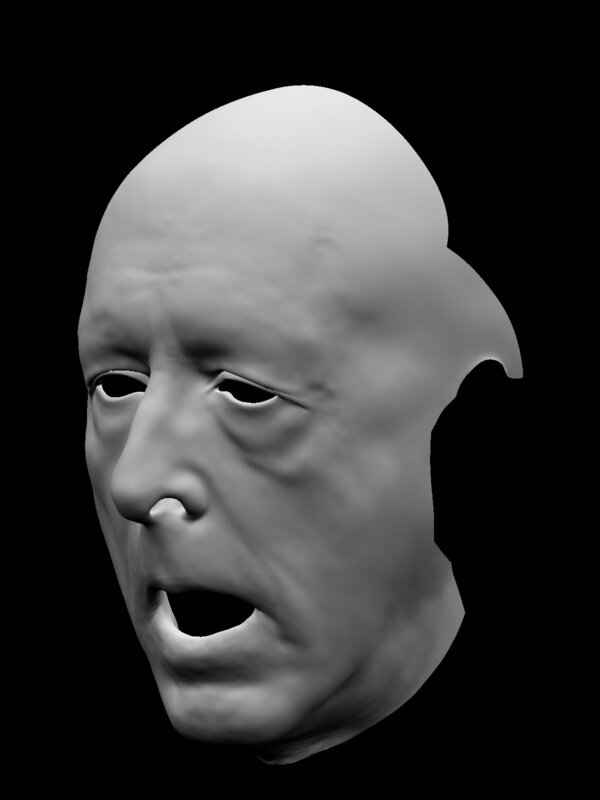}
    \caption{Temporal Smoothing}
\end{subfigure}
\begin{subfigure}[b]{0.20\linewidth}
    \includegraphics[width=\linewidth]{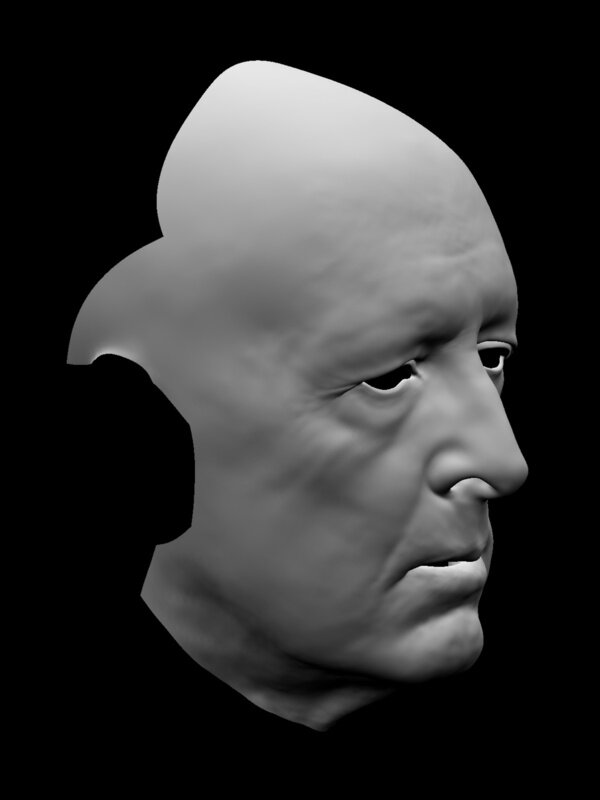}
    \includegraphics[width=\linewidth]{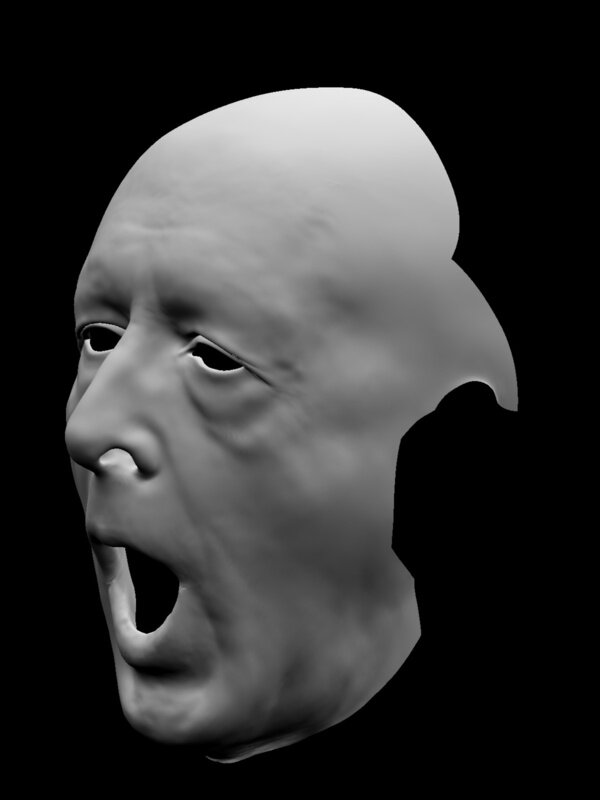}
    \includegraphics[width=\linewidth]{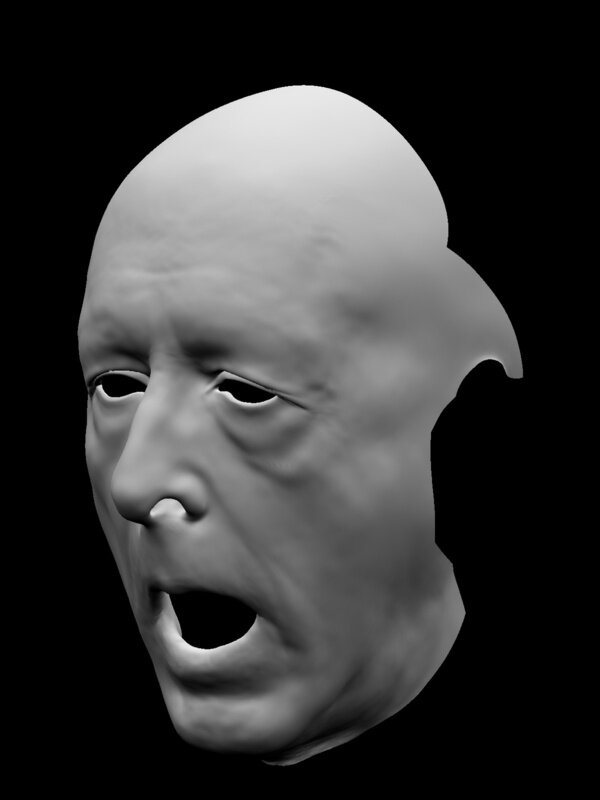}
    \caption{Re-estimation}
\end{subfigure}
\begin{subfigure}[b]{0.20\linewidth}
    \includegraphics[width=\linewidth]{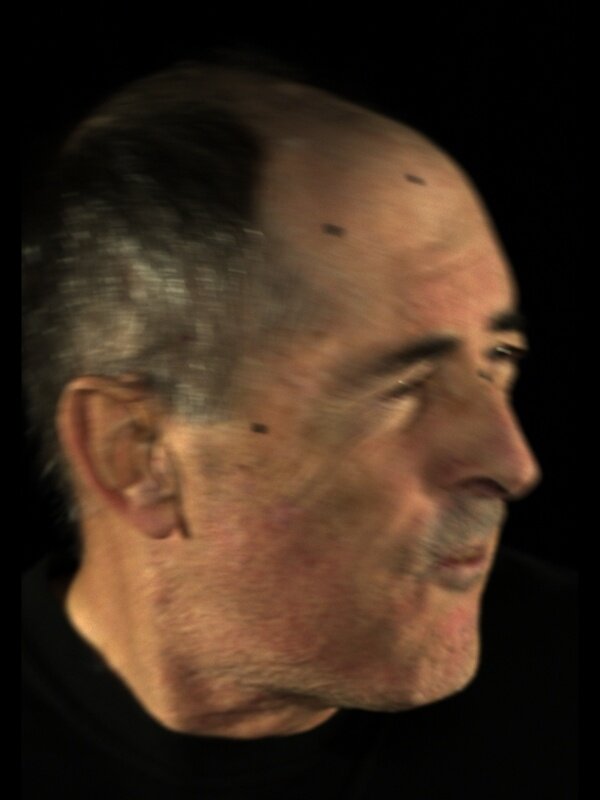}
    \includegraphics[width=\linewidth]{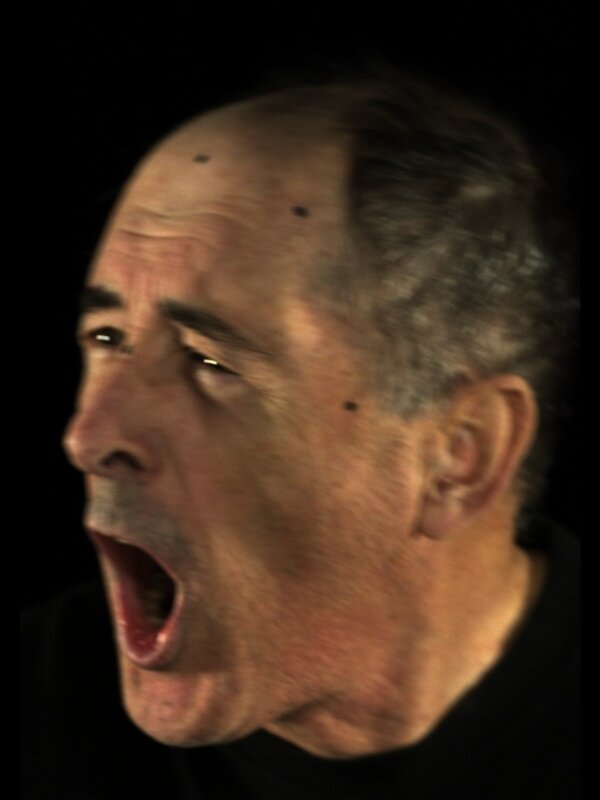}
    \includegraphics[width=\linewidth]{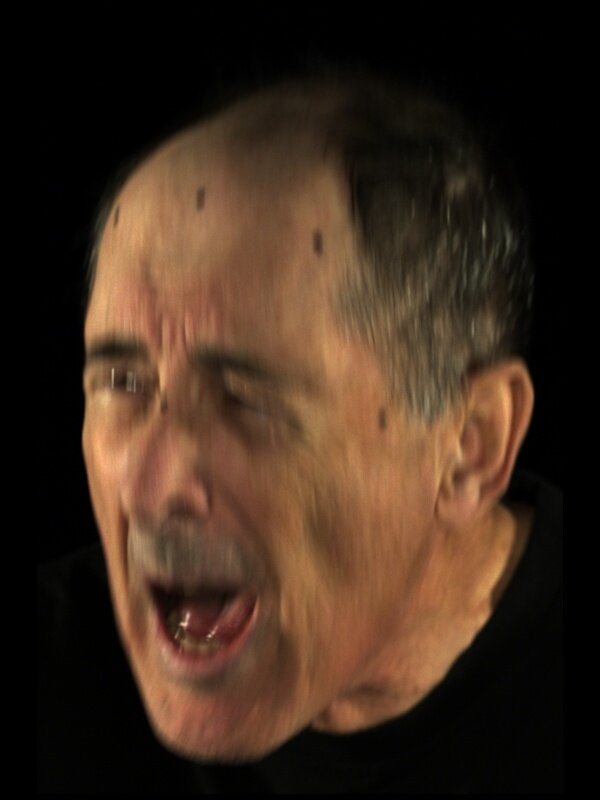}
    \caption{Target}
\end{subfigure}
\hfill
\caption{A comparison of the result of stereo expression estimation, stereo temporal smoothing, and stereo expression re-estimation after temporal smoothing.
}
\label{fig:mouth_after_smoothing}
\end{figure*}

The expression estimation and temporal smoothing steps can be repeated multiple times until convergence to produce more accurate results. 
To demonstrate the potential of this approach, we reestimate the facial expression by solving for the mouth and jaw blendshape parameters (a subset of $w$) while keeping the rigid parameters fixed after temporal smoothing.
As seen in Figure \ref{fig:mouth_after_smoothing}, the resulting facial expression is generally more accurate than the pre-temporal smoothing result.
Furthermore, in the case where temporal smoothing dampens the performance, performing expression re-estimation will once again capture the desired expression (frame \num{1159}).
\clearpage

{\small
\bibliographystyle{ieee}
\bibliography{biblio}
}

\end{document}